\documentclass[lettersize,journal]{IEEEtran}
\usepackage{amsmath,amsfonts}
\usepackage{algorithmic}
\usepackage{algorithm}
\usepackage{array}
\usepackage[caption=false,font=normalsize,labelfont=sf,textfont=sf]{subfig}
\usepackage{textcomp}
\usepackage{stfloats}
\usepackage{url}
\usepackage{verbatim}
\usepackage{graphicx}
\usepackage{colortbl}
\usepackage{color}
\usepackage{cite}
\usepackage{multirow}
\usepackage{bm}
\usepackage[breaklinks=false,bookmarks=false]{hyperref}
\def\x{\bm{\lambda}}
\def\y{\bm{\omega}}
\def\a{\bm{\alpha}}
\def\s{\mathbf{s}}
\def\t{\mathbf{t}}
\def\ms{\mathbf{s}_\mathbf{m}}
\def\mt{\mathbf{t}_\mathbf{m}}
\def\D{\mathcal{D}}
\definecolor{mygray}{gray}{.9}
\definecolor{mycyan}{cmyk}{.3,0,0,0}
\definecolor{update}{rgb}{0,0,0}
\newcommand\twoitems[2]{%
	\item#1%
	\hspace{15pt}%
	\labelitemi
	\hspace{\labelsep}#2
}

\begin{document}

\title{Automated Learning for Deformable Medical Image Registration by Jointly Optimizing Network Architectures and Objective Functions}

\author{Xin~Fan$^{\dag}$,~\IEEEmembership{Senior Member,~IEEE},
        Zi~Li$^\dag$,
        Ziyang~Li,
        Xiaolin~Wang,
        Risheng~Liu,~\IEEEmembership{Member,~IEEE,}\\
        Zhongxuan~Luo,
        and~Hao~Huang
        
\thanks{X. Fan, Z. Li, Z. Li, X. Wang, R. Liu, and Z. Luo are with the DUT-RU International School of Information Science \& Engineering and the Key Laboratory for Ubiquitous Network and Service Software of Liaoning Province, Dalian University of Technology, Dalian 116024, China. Z. Luo is also with the Institute of Artificial Intelligence, Guilin University of Electronic Technology, Guilin 541004, China. (E-mail: xin.fan@ieee.org; alisonbrielee@gmail.com; liziyang1997@mail.dlut.edu.cn;  wxl1009@mail.dlut.edu.cn; rsliu@dlut.edu.cn; zxluo@dlut.edu.cn).
} 
\thanks{H. Huang is with the Department of Radiology, Children’s Hospital of Philadelphia, Philadelphia, PA, United States. H. Huang is also with the Department of Radiology, Perelman School of Medicine, University of Pennsylvania, Philadelphia, PA, United States
(E-mail: huangh6@email.chop.edu).}
\thanks{ $\dag$ Equal contribution. Corresponding author: X. Fan.}}

\markboth{Journal of \LaTeX\ Class Files,~Vol.~14, No.~8, August~2021}%
{Shell \MakeLowercase{\textit{et al.}}: A Sample Article Using IEEEtran.cls for IEEE Journals}

\IEEEpubid{0000--0000/00\$00.00~\copyright~2021 IEEE}

\maketitle

\begin{abstract}
Deformable image registration plays a critical role in various tasks of medical image analysis.
A successful registration algorithm, either derived from conventional energy optimization or deep networks, requires tremendous efforts from computer experts to well design registration energy or to carefully tune network architectures with respect to medical data available for a given registration task/scenario. 
This paper proposes an automated learning registration algorithm (AutoReg) that cooperatively optimizes both architectures and their corresponding training objectives, enabling non-computer experts to conveniently find off-the-shelf registration algorithms for various registration scenarios. 
Specifically, we establish a triple-level framework to embrace the searching for both network architectures and objectives with a cooperating optimization.
Extensive experiments on multiple volumetric datasets and various registration scenarios demonstrate that AutoReg can automatically learn an optimal deep registration network for given volumes and achieve state-of-the-art performance. The automatically learned network also improves computational efficiency over the mainstream UNet architecture from 0.558 to 0.270 seconds for a volume pair on the same configuration.
\end{abstract}

\begin{IEEEkeywords}
Medical image registration, Automatic machine learning, Neural architecture search, Hyperparameter optimization, Convolution neural network.
\end{IEEEkeywords}

\section{Introduction}\label{sec:intro}
\begin{figure*}[t] 
	\centering 
	\includegraphics[width=1\textwidth]{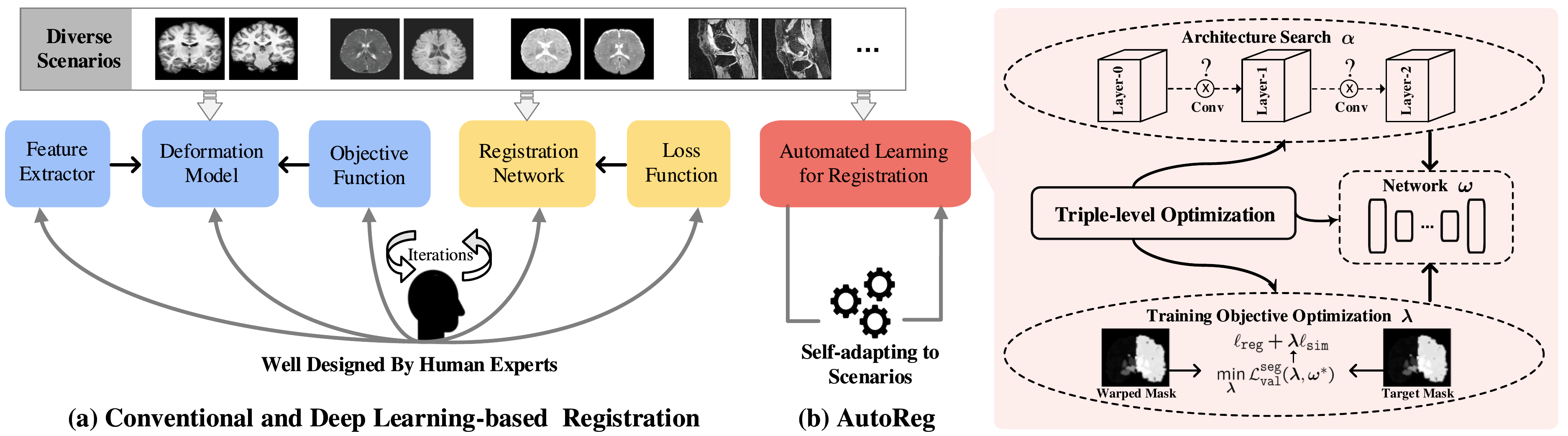}  
	\caption{
		(a) Both conventional and learning-based registration techniques require computer experts to well design core components for different medical image registration tasks, and (b) AutoReg, a user-friendly automatic registration framework, learns off-the-shelf deep registration algorithms for various scenarios by jointly optimizing the network architecture $\a$, hyper-parameters $\x$ in training objective, and network parameters $\y$.
	} 
	\label{fig:piplin}
\end{figure*}

\IEEEPARstart{D}{eformable} image registration (DIR) establishes dense spatial correspondences between different medical image acquisitions~\cite{MaintzV98}. It plays a critical role in various tasks of medical image analysis including anatomical change diagnosis~\cite{jie2018sub}, longitudinal studies~\cite{jocn.2007.19.9.1498} and statistical atlas building~\cite{NIPS2019_8368}. 
Given a source image $\mathbf{s}$ and a target image $\mathbf{t}$ on a spatial domain ${\Omega \in \mathbb{R}^{d}}$, the goal of DIR is to find an optimal non-linear dense transformation or field  $\bm{\varphi} : \Omega \times \mathbb{R} \to \Omega $ that minimizes the energy:
\begin{equation}
\quad\min\limits_{\bm{\varphi}}E_D(\bm{\varphi};f\circ\s,f\circ\t)+ E_R(\bm{\varphi}),
\label{eq:energy}
\end{equation}
where $E_D$ is a data matching term, evaluating the similarity between aligned features $f\circ\s$ and $f\circ\t$ of the source and target while $E_R$ is a regularization term that reflects the nature of the transformation.
\IEEEpubidadjcol
Medical image analysis faces a wide range of registration scenarios involving different facets of developing a registration algorithm~\cite{SotirasDP13}: estimating spatial correspondences relies on distinct features robust to intensity variations from intra/inter subjects and acquisition equipment; the deforming complexity varies with the nature of anatomical organs; the similarity metric for registering images of the same modality is evidently different from that for cross-modality images. Hence, the three major components,~\emph{i.e.}, a feature extractor, a deformation model, and an objective function, have to be well designed in order to construct the energy~\eqref{eq:energy} for a specific scenario, as shown in the left of Fig.~\ref{fig:piplin}(a). This modeling process demands deep understanding of the registration scenario as well as sophisticated mathematical skills. Meanwhile, the energy optimization invokes a computationally intensive process, typically taking minutes or even hours for registration~\cite{SyN,chaudhury2020multilevel,SunNK14}.   

Witnessing the great success of deep learning (DL) in many computer vision tasks, researchers replace manually-crafted feature extractors and parametric deformation models with deep networks so that DIR turns out to be one step prediction with network parameters (weights) learned from image pairs, rendering faster deformation estimation than iterative optimization~\cite{BalakrishnanZSG19,ZhaoDCX19,DalcaBGS19,WangZ20,LiuLZFL20,pan2020octrexpert,zhang2021two,ofverstedt2019fast,cao2018region}. 
Nevertheless, these deep learning based methods highly depend on training examples available for a scenario. One has to re-train the network when applying to a new scenario with significantly different data. In many occasions, tuning network architectures and/or loss functions is also necessary to gain optimal performance on the new scenario, as shown in the right of Fig.~\ref{fig:piplin}(a). For example, a network trained for unimodal registration tasks cannot accurately align images of different modalities. Recent studies~\cite{HoopesHFGD21,9551747} present new training strategies that automatically find optimal hyper-parameters for loss functions. Unfortunately, algorithm developers still have to adjust network architectures for feature extraction and/or deformation by exploring numerous possibilities. This manual train-trial process is extremely time consuming since one single training stage may take hours or even days.  

A successful DIR algorithm, either derived from conventional energy optimization or deep networks, requires tremendous efforts from computer experts to well design registration energy or to carefully tune network architectures. Therefore, existing paradigms prohibit medical/clinical users from exploring registration adaptive to their scenarios and available data. A new user-friendly automatic registration paradigm is desired to enable medical/clinical users even without algorithm-developing expertise to conveniently find off-the-shelf registration algorithms for various scenarios.  

\subsection{Our Contributions}
We propose a triple-level optimization framework that jointly learns the weights, architecture, and loss function of a deep network for DIR, as shown in Fig.~\ref{fig:piplin}(b). Specifically, besides one level of optimization to learn network weights, we formulate one level to search optimal architectures for \emph{feature} and \emph{deformation} networks and another level to discover proper hyper-parameters for \emph{objective functions}. Consequently, our framework enables automatically learning nearly all major components of a registration algorithm tailored to a given scenario, and thus frees both medical researchers and computer engineers from enormous efforts to sophisticated modeling and tedious parameter tuning. Moreover, the auto-learned registration also runs extremely fast inheriting from common deep-based algorithms.  
We summarize our contributions below:
\begin{itemize}
	\item{
	We devise an automated learning registration algorithm (AutoReg) to cooperatively optimize the architecture and loss function of a deep registration network. The auto-search strategy can be combined with the existing registration methods to achieve automatic search. To our best knowledge, this is the first piece of work that achieves Automated Machine Learning (AutoML)\footnote{AutoML automates the time-consuming, iterative tasks of ML development, making ML available for non-ML experts.} for medical image registration.}
	\item{
	To efficiently discover the architecture adaptive to the registration scenario of interest, we construct a search space with an adaptive feature cell and a task-aware deformation cell, which consist of various convolution operators practically effective to feature learning and deformation estimation, respectively.}
	\item{ 
    We introduce the similarity from two aspects of the intensity distribution and structural preservation into a unified loss function and further optimize the hyper-parameters weighing the similarity and regularization terms in order to deploy the scenario-oriented training loss.}
	\item{
	We convert the triple-level optimization into hierarchical bilevel sub-problems and then iteratively apply efficient gradient-based algorithms to the complicated optimization that couples variables of different natures. This auto-search mechanism with cooperating optimization drastically alleviates time expenses compared with the manual train-trial process.}
\end{itemize}
We validate the AutoReg algorithm on several typical MRI registration scenarios ranging from mono-modality (registering brain T1 images to an averaged T1 template, brain T1 to T1 images, brain T2 to T2 images, knee T1 to T1 images, and lung CT inspiration to expiration images) to multi-modality (registering brain T1 to T2 images\textcolor{update}{, brain T2 to T1 images}). Despite of significant data variations occurring in these scenarios, AutoReg can learn an efficient algorithm that outputs more accurate deformation fields than the state-of-the-art algorithms re-trained by the data from respective scenarios.

This paper is organized as follows: we first describe related works in Section~\ref{sec:related} and introduce our automated learning for registration in Section~\ref{sec:method}. Next, we demonstrate ablation studies and experimental comparisons in Section~\ref{sec:experiment}. Section~\ref{sec:conclusion} concludes this paper.

\section{Related Works}\label{sec:related}

This section briefly reviews recent deep algorithms for DIR and AutoML techniques for computer vision tasks other than image registration.

\subsection{Deep Learning-based Deformable Image Registration}

Deep learning based algorithms have gained impressive progress in DIR by taking advantage of powerful representation ability of convolutional neural networks. Inspired by the spatial-transformer work~\cite{JaderbergSZK15}, DL-based approaches employ unsupervised feature learning in the training process and apply one-step inference rather than costly iterative optimization in conventional registration to generate deformation fields~\cite{BalakrishnanZSG19,ZhaoDCX19}.
Further studies estimate the velocity or momentum fields to impose diffeomorphism to the final deformations~\cite{DalcaBGS19,WangZ20}.
Alternatively, Niethammer~\emph{et al.} embed a spatially-varying regularizer into a feature network~\cite{NiethammerKV19}. 
In our previous work, we constitute an unsupervised deep network to simultaneously learn deformation maps and features~\cite{LiuLZFL20}. 

Recent attempts improve the generalization ability of deep registration adapting to a wider range of scenarios.
Hoopes~\emph{et al.} introduce a hyper-parameter and modulate a registration hypernetwork in order to produce an optimal deformation field given the hyper-parameter value~\cite{HoopesHFGD21}.
The work of~\cite{9551747} proposes a bilevel training strategy that learns optimal hyper-parameters of training loss functions for deep networks of DIR.
Nevertheless, existing methods can learn a subset of major components of a deep registration network from given data, but manually designing and/or tuning other components still requires tremendous human experience and efforts.

\subsection{AutoML for Computer Vision Tasks}
We focus on Neural Architecture Search (NAS) that automates designing architectures of deep neural networks for computer vision tasks. NAS has drawn increasing attention owing to its capability of relieving intensive human labors meanwhile outputting considerable performance.
In pioneering works, aiming at automating the design of network, researchers  develop either reinforcement learning~\cite{ZophVSL18}, evolutionary algorithms~\cite{RealAHL19} or bayesian optimization~\cite{kandasamy2018neural,dai2019chamnet} to find the solution to optimal network architectures for specific tasks.
Unfortunately, these optimization strategies are computationally demanding.
Liu~\emph{et al.} propose the continuous relaxation of the discrete search space and take differentiable optimization techniques in order to efficiently learn deep network architectures for image classification and language modeling tasks~\cite{LiuSY19}. Zhou~\emph{et al.} propose a Bayesian approach to optimize
one-hot Neural Architecture Search~\cite{zhou2019bayesnas}.
These differentiable NAS strategies can significantly reduce the search time from months/days to even hours, making NAS viable for common users~\cite{DongY19,liu2020block}.

These NAS techniques have been applied to various types of deep networks for many CV tasks including low-level vision, image segmentation, and object detection ~\cite{GhiasiLL19,PengBFFK20,Liu0Z0L21,liu2021adaptive,yang2022video,tang2021autopedestrian}.
Recently, researchers employ NAS to search UNet-like architectures for medical image segmentation~\cite{YuYRBZYX20,GuoJZHHCX020,He0RZX21}.
Automated designing networks for DIR desires searching heterogeneous networks for both feature and deformation learning. This study develops such an algorithm along with hyper-parameter optimization, which fully automates learning optimal DIR algorithms adaptive to various scenarios.

\section{Automated Learning for Deformable Image Registration}\label{sec:method}

We construct a triple-level algorithmic framework to jointly optimize the network architecture and training objective so that we can automate the process of designing three major components for DIR,~\emph{i.e.}, feature extraction, deformation model, and objective function, given a set of data for a scenario. This section begins with the optimization formulation, then gives the two levels of optimization for architecture and objective, and ends with the optimization/learning process.

\subsection{Problem Formulation}\label{sec:methodFormulation}
In a general deep learning-based framework for DIR, the spatial correspondence map $\bm{\varphi}$ between the source $\s$ and targe $\t$ is represented by a deep network $\Psi(\y;\s,\t)$, where $\y$ denotes learnable network parameters of $\Psi$. Taking the inputs $\s$ and $\t$, the prediction or inference of $\Psi$ upon the learned $\y$ yields the deformable map~\cite{BalakrishnanZSG19}.

Herein, we introduce an architecture parameter $\a$ that represents a series of covolutional operators constituting the network. Meanwhile, the training objective for $\Psi$ contains hyper-parameters $\x$ that weigh the objective terms for similarity and regularization.
We formulate jointly learning $\a$, $\x$ and $\y$ as triple-level optimization:
\begin{equation}
\begin{array}{l}
\qquad  \quad \mathop{ \min}\limits_{ \x } \ \mathcal{L}^{\mathtt{seg}}_{\mathtt{val}} ( \x, \a^*, \y^*; \s, \t),  \\
s.t. \left\{
\begin{array}{lr}
\a^*(\x) = \mathop{\arg \min} \limits_{\a} \ \mathcal{L}^{\mathtt{reg}}_{\mathtt{val}}( \a, \y^*(\a);\x, \s, \t), \\ s.t. \ \y^*(\a) = \mathop{\arg \min} \limits_{\y} \ \mathcal{L}^{\mathtt{reg}}_{\mathtt{tr}}( \y ; \a, \x, \s, \t).
\end{array}
\right.
\end{array}
\label{eq:triplelevel}
\end{equation}

The first level of optimization trains the network parameters $\y$ using a training data set for a scenario given a network architecture and a loss function $\mathcal{L}^{\mathtt{reg}}_{\mathtt{tr}}$ evaluating similarity between the target and deformed source. The second and third levels learn $\a$ and $\x$ from a validation set for the same scenario, respectively. The training losses of $\y$ and $\a$ share the same definition $\mathcal{L}^{\mathtt{reg}}$ as the learned network $\Psi_{\a}(\y)$ given by $\y$ and $\a$ determines the deformation map for registration. We define the loss function $\mathcal{L}^{\mathtt{seg}}_{\mathtt{val}}$ for learning $\x$ by evaluating the overlapped anatomical regions between the target and deformed source. This definition is inspired by the use of Dice scores when manually tuning hyper-parameters $\x$. Our formulation embraces learning the three major components for DIR as $\Psi_{\a}(\y)$ implies feature extraction and deformation models while $\x$ adjusts objective functions, shown in Fig.~\ref{fig:piplin}(b).

\subsection{Architecture Search}\label{sec:methodArchitecture}
\begin{figure}[!t] 
	\centering 
	\includegraphics[width=0.48\textwidth]{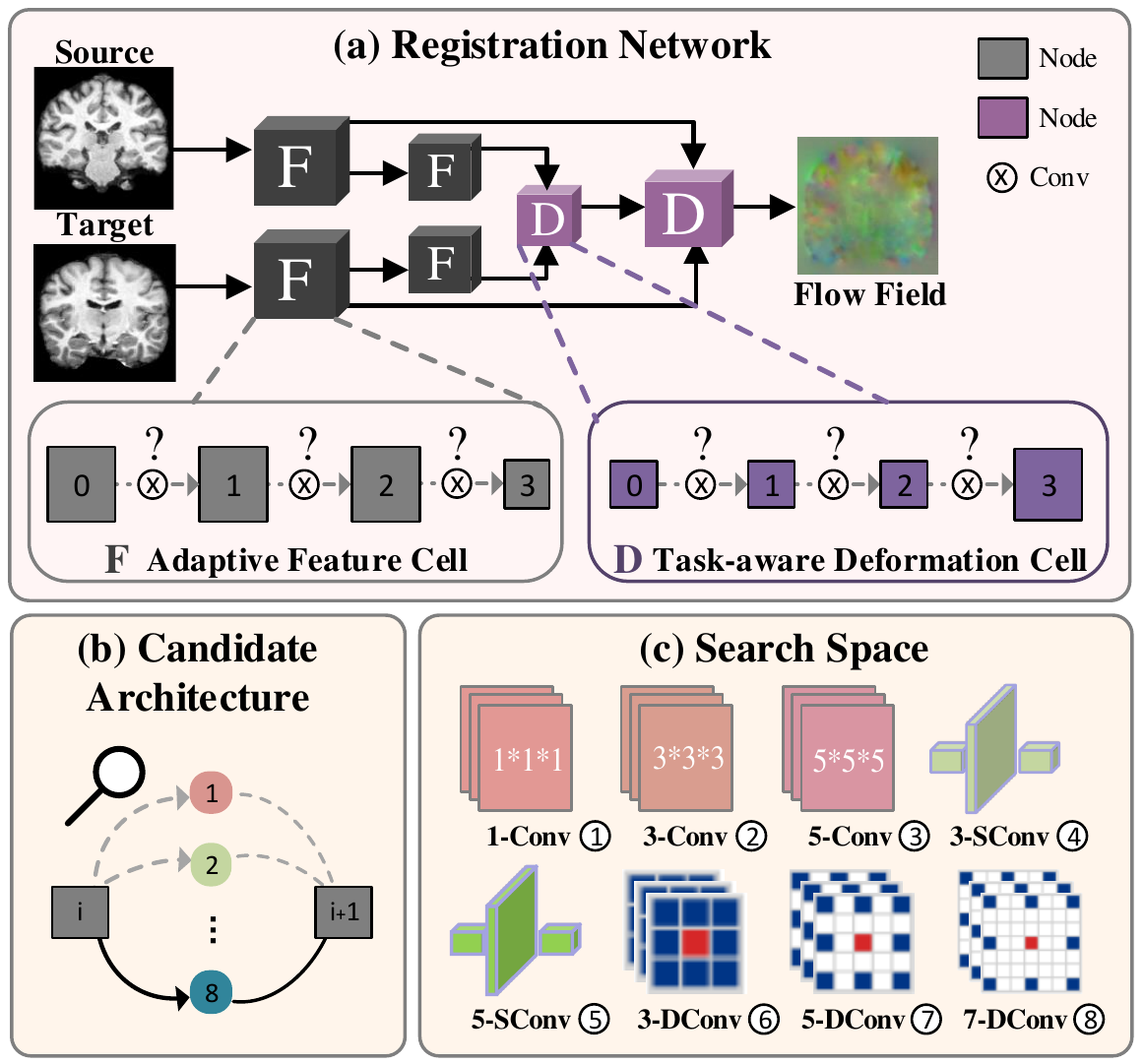}  
	\caption{ 
		(a) The network backbone consisting of cells for feature extraction $F$  and deformation generation $D$ at multiple scales, 
		(b) the structure for searching cells, and (c) eight candidate operations.
	} 
	\label{fig:piplin-nas}
\end{figure}

Both feature extraction and deformation generation take a three-layer convolutional network, running at two scales, shown as the dark and purple blocks in the top of Fig.~\ref{fig:piplin-nas}(a). We take this multi-scale architecture as the backbone because extensive studies on brain MRI registration validate its superior performance~\cite{LiuLZFL20,9551747}. However, image intensities and/or organ geometries significantly vary with different registration scenarios. Computer engineers have to devote great efforts to trial various types and kernel sizes of convolutional operators applied between layers, that are central building blocks for neural networks, in order to accommodating significant data variations. This study strives to automatically discover appropriate operations, and hence free users from tedious hand-crafted trial-tests.

Specifically, we construct an Adaptive Feature Cell and a Task-aware Deformation Cell at each scale to represent candidate architectures for feature learning and deformation generation, respectively, shown in the bottom of Fig.~\ref{fig:piplin-nas}(a). These two types of cells share a common structure,~\emph{i.e.}, a directed acyclic graph (DAG) with four nodes, whose edges are candidate convolutional operations. The first node is the network input while the other three represent the outputs by applying the edge operator to the previous node. Fig.~\ref{fig:piplin-nas}(b) illustrates the structure of two consecutive nodes. The cells differ in the first and/or last nodes since their input and output are different as discussed below.  

\noindent\textbf{Adaptive Feature Cell:} We take the input source or target image as the first node while down-sample the output of the last convolution by taking the stride of 2 as the last node. We learn one common cell for both source and target so that they share the same network architecture but have different network parameters given the architecture. 

\noindent\textbf{Task-aware Deformation Cell:} We concatenate the outputs of the feature learning network for the source and target images as the first node. Meanwhile, we up-sample the output at the last convolution by a resizing operation as the last node. To obtain a diffeomorphism deformation, we parameterize the deformation field with a stationary velocity field, and thus generate the final field under the government of ordinary differential equations~\cite{Ashburner07}.

As shown in Fig.~\ref{fig:piplin-nas}(b), we set eight convoutional candidates as the edge connecting two nodes including three types of convolutions,~\emph{i.e.}, regular, separable, and dilated ones, with two or three kernel sizes. In practice, a computer engineer typically attempts to numerate the combinations of these operations as much as possible in order to gain improvements for a new registration scenario. Fig.~\ref{fig:piplin-nas}(c) visualizes the eight operations listed below.
\begin{itemize}
	\twoitems{1$\times$1$\times$1 Conv (1-Conv) }{3$\times$3$\times$3 Conv (3-Conv) }
	\item{ 5$\times$5$\times$5 Conv (5-Conv)}
	\item{ 3$\times$3$\times$3 Separable Conv (3-SConv)}
	\item{ 5$\times$5$\times$5 Separable Conv (5-SConv)}
	\item{ 3$\times$3$\times$3 Dilation Conv (3-DConv)}
	\item{ 5$\times$5$\times$5 Dilation Conv (5-DConv)}
	\item{ 7$\times$7$\times$7 Dilation Conv (7-DConv)}
\end{itemize} 
Therefore, the architecture search becomes finding an optimal path composed of six convolutional operations~\footnote{All these convolutions are followed by a Leaky ReLU function.} traversing the two cells at every scale. Our search space spans $8^{(3+3)}$ at one scale and $8^{12}$ possible candidates in total (two scales).

We follow the idea of continuous relaxation in~\cite{LiuSY19} to convert the time consuming discrete search over a huge space into a continuous optimization problem where efficient gradient-based algorithms are applicable.   
Let $o(\cdot)$ denotes one operation of the candidate set $\mathcal{O}$ to be applied to the node $x^{(i)}$. We parameterize the weights of candidate operations for a pair of nodes $x^{(i)}$ and $x^{(i+1)}$ as a column vector $\alpha_o^{(i, i+1)}$ of the dimension $|\mathcal{O}|$ (eight for our setting). We relax the discrete choice of a particular operation to a softmax over all possible operations:
\begin{equation}
\bar{o}^{(i, i+1)}(x^{(i)}) = \sum_{o \in \mathcal{O} } \frac{ \mathrm{exp} (\alpha^{(i,i+1)}_o)}{ \sum_{o^{\prime} \in \mathcal{O} } \mathrm{exp} (\alpha^{(i,i+1)}_{o^{\prime}}) }  o(x^{(i)}) .
\end{equation}
Thereby, searching network architecture reduces to computing a set of continuous variables $\a = \{\a_F,\a_D\}=\{ \alpha_F^{(i, i+1)},\alpha_D^{(i, i+1)} \}_{i=0,1,2}$, and finally taking each edge with the most likely operation, $o^{(i, i+1)} = \arg\max_{o \in \mathcal{O}} \alpha^{(i,i+1)}_o $. 
\textcolor{update}{Namely, at the end of the architecture search, the optimal architecture can be obtained by replacing each mixed operation with the most likely operation by $\arg\max$ function.}
We will detail the gradient-based optimization of $\a$ in Sec.~\ref{sec:methodOptimization}\textcolor{update}{, where the $\arg\max$ function is used at the end of Stage 1 and Stage 2 of Algorithm~\ref{alg:optimization} to induce the optimal architecture $\a_F$ and $\a_D$, respectively}.

\subsection{Training Objective Optimization}\label{sec:methodObjective}

Training objectives for deep networks determine both training convergence and prediction accuracy. Our triple-level formulation enables bringing additional cues for different levels of optimization so that it involves two types of objective losses, $\mathcal{L}^{\mathtt{reg}}$, $\mathcal{L}^{\mathtt{seg}}$, for training the network architectures $\a$ and parameters $\y$, and the hyper-parameters $\x$, respectively.

%
%
We define the similarity term in the objective $\mathcal{L}^{\mathtt{reg}}$ from two aspects,~\emph{i.e.}, intensity distribution and structural consistency, in order to accommodate a wide range of scenarios including mono-modality and cross-modality registrations. The local Normalized Cross-correlation Coefficient (NCC) evaluates the similarity of intensity distributions between an image pair from the same modality. Nevertheless, NCC can hardly characterize the similarity when evident intensity discrepancies exist especially for images from different modalities. Hence, we introduce 
the Modality Independent Neighborhood Descriptor (MIND)~\cite{HeinrichJBMGBS12} into similarity evaluation. MIND extracts various types of geometrical structures,~\emph{e.g.}, corners and edges, in a local neighborhood of an image, and can thus evaluate the structural consistence between images across modalities. 
%
We combine NCC and MIND into one similarity metric by a trade-off parameter $\lambda_1$:
\begin{align}
\ell_{\mathtt{sim}} =   \lambda_{\mathtt{1}} \ell_{\mathtt{NCC}} + (1- \lambda_{\mathtt{1}}) \ell_{\mathtt{MIND}}.
\end{align}
We also incorporate a diffusion regularizer on spatial gradients of deformation fields~\cite{BalakrishnanZSG19} together with multi-scale similarities~\cite{DosovitskiyFIHH15} into the registration loss $\mathcal{L}^{\mathtt{reg}}$:
%
\begin{equation}
\mathcal{L}^{\mathtt{reg}}  =   \ell_{\mathtt{sim}}^{1}  +   \lambda_{2} \ell_{\mathtt{smooth}} + \lambda_{3} \ell_{\mathtt{sim}}^{1/2} + \lambda_{4} \ell_{\mathtt{sim}}^{1/4}  , \\ 
\label{eq:regloss}
\end{equation}
where $\ell_{\mathtt{sim}}^{1}$, $\ell_{\mathtt{sim}}^{1/2}$, and $\ell_{\mathtt{sim}}^{1/4}$ denote similarity loss functions at full, half, and quarter resolutions, respectively, and $\ell_{\mathtt{smooth}}$ represents the regularization loss. We employ the diffusion regularizer on spatial gradients of the registration fields. We assemble all these scalar hyper-parameters into one vector $\x := \{\lambda_{1}, \lambda_{2}, \lambda_{3}, \lambda_{4}\}$ and optimize $\x$ with respect to a given validate set for a scenario.

A human expert may visually inspect the coincidence of aligned regions with anatomical parcellation when tuning the hyper-parameters $\x$. From this respect, we leverage an auxiliary segmentation map, which assigns each voxel to an anatomical structure, to define the loss $\mathcal{L}^{\mathtt{seg}}$ for the top level of optimization in \eqref{eq:triplelevel} that automatically learns a scenario-specific $\x$ as shown in Fig.~\ref{fig:piplin}(b).
The regions in the warped source by an accurate deformation field $\s \circ \bm{\varphi}$ would be perfectly coincident with those in the target $\t$ from the same anatomical structure. We quantify the volume overlapping for all structures using the Dice score~\cite{Dice} and define the loss as: 
\begin{equation}
\label{eq:segloss}
\mathcal{L}^{\mathtt{seg}} = \ell_{\mathtt{dice}}(\ms \circ \bm{\varphi}, \mt),
\end{equation}
where $\ms$ and $\mt$ are the segmentation maps for $\s$ and $\t$, respectively. Similar to~\cite{BalakrishnanZSG19}, we convert the segmentation masks to the one-hot format and spatially transform them using linear interpolation when computing the loss.

\begin{algorithm}[!t]
	\caption{Procedure for solving the model in Eq.~(\ref{eq:triplelevel})}
	\label{alg:optimization}
	\begin{algorithmic}[1]
		\REQUIRE 
		The search space,  $\D_{\mathtt{tr}}$ and $\D_{\mathtt{val}}$.
		\ENSURE The searched architecture and hyper-parameter. 
		\STATE \textcolor{update}{Initialize model weight $\y = \lbrace\y_{SF},\y_{D}\rbrace. $}\\
		\%\% Stage 1: Search for  $\a_{F}$ with fixed $\x$.
		\WHILE{not converged}
		\STATE $ \y^{\prime} = \y - \frac{\mathrm{d}\mathcal{L}^{\mathtt{reg}}_{\mathtt{tr}}(\y, \a_{F})}{\mathrm{d}{\y}}$, $\y^{\pm}=\y\pm\epsilon\frac{\mathrm{d} \mathcal{L}^{\mathtt{reg}}_{\mathtt{val}} (\y^{\prime}, \a_{F})}{\mathrm{d}\y^{\prime}}$
		\STATE $  V_{\a_{F}}  = - \frac{1}{2\epsilon} (\frac{\partial \mathcal{L}^{\mathtt{reg}}_{\mathtt{tr}} (\y^{+},\a_{F} )}{\partial \a_{F}} -\frac{\partial \mathcal{L}^{\mathtt{reg}}_{\mathtt{tr}} (\y^{-},\a_{F})}{\partial \a_{F}})  $
		\STATE $\a_{F} \leftarrow \a_{F} - \frac{\partial \mathcal{L}^{\mathtt{reg}}_{\mathtt{val}} (\y^{\prime}, \a_{F})}{\partial \a_{F}} - V_{\a_{F}} $
		\STATE $\y \leftarrow \y - \frac{\mathrm{d}\mathcal{L}^{\mathtt{reg}}_{\mathtt{tr}}(\y, \a_{F})}{\mathrm{d}{\y}}   $
		\ENDWHILE \\
		\%\% Stage 2: Search for  $\a_{D}$ with fixed $\x$.
		\STATE \textcolor{update}{Initialize model weight $\y = \lbrace\y_F,\y_{SD}\rbrace $. } \\
		\WHILE{not converged}
		\STATE $ \y^{\prime} = \y - \frac{\mathrm{d}\mathcal{L}^{\mathtt{reg}}_{\mathtt{tr}}(\y, \a_{D})}{\mathrm{d}{\y}}$, $\y^{\pm}=\y\pm\epsilon\frac{\mathrm{d} \mathcal{L}^{\mathtt{reg}}_{\mathtt{val}} (\y^{\prime}, \a_{D})}{\mathrm{d}\y^{\prime}}$
		\STATE $  V_{\a_{D}} = - \frac{1}{2\epsilon} (\frac{\partial\mathcal{L}^{\mathtt{reg}}_{\mathtt{tr}} (\y^{+},\a_{D} )}{\partial\a_{D}} -\frac{\partial\mathcal{L}^{\mathtt{reg}}_{\mathtt{tr}} (\y^{-},\a_{D})}{\partial \a_{D}})  $		
		\STATE $\a_{D} \leftarrow \a_{D} - \frac{\partial \mathcal{L}^{\mathtt{reg}}_{\mathtt{val}} (\y^{\prime}, \a_{D})}{\partial \a_{D}}  - V_{\a_{D}}  $
		\STATE $\y \leftarrow \y - \frac{\mathrm{d}\mathcal{L}^{\mathtt{reg}}_{\mathtt{tr}}(\y, \a_{D})}{\mathrm{d}{\y}} $
		\ENDWHILE \\
		\%\% Stage 3: Search for  $\x$ with optimal $\a$.
		\STATE \textcolor{update}{Initialize model weight $\y = \lbrace\y_F,\y_D\rbrace $. Warm-start.} \\
		\WHILE{not converged}
		\STATE $ \y^{\prime} = \y - \frac{\mathrm{d}\mathcal{L}^{\mathtt{reg}}_{\mathtt{tr}}(\y, {\x})}{\mathrm{d}{\y}}$, $\y^{\pm}=\y\pm\epsilon\frac{\mathrm{d} \mathcal{L}^{\mathtt{seg}}_{\mathtt{val}} (\y^{\prime}, {\x})}{\mathrm{d}\y^{\prime}}$
		\STATE $ V_{\x} = - \frac{1}{2\epsilon} (\frac{\partial \mathcal{L}^{\mathtt{reg}}_{\mathtt{tr}} (\y^{+},\x)}{\partial {\x}} - \frac{\partial\mathcal{L}^{\mathtt{reg}}_{\mathtt{tr}} (\y^{-},\x)}{\partial{\x}})  $
		\STATE $\x \leftarrow \x - \frac{\partial\mathcal{L}^{\mathtt{seg}}_{\mathtt{val}} (\y^{\prime}, \x)}{\partial{\x}}  - V_{\x} $
		\STATE $\y \leftarrow \y - \frac{\mathrm{d}\mathcal{L}^{\mathtt{reg}}_{\mathtt{tr}}(\y, \x)}{\mathrm{d}{\y}}  $
		\ENDWHILE
		\RETURN  Architecture and hyper-parameter  $\a_{F}^*, \a_{D}^*, \x^*$.  
	\end{algorithmic}
\end{algorithm}

\subsection{Optimization Procedure}\label{sec:methodOptimization}
It is challenging to solve the triple-level optimization in Eq.~(\ref{eq:triplelevel}) owing to coupling several large-scale optimization problems of different natures. We convert the triple-level model into hierarchical bilevel optimization by fixing the other variables when optimizing one. We first tackle the bi-level optimization of the architecture $\a_{F}, \a_{D}$ and network parameters $\y$ given the hyper-parameter $\x$ and training objective:
\begin{equation}
\begin{array}{l}
\begin{array}{lr}
\mathop{ \min}\limits_{\a_{F}, \a_{D}} \ \mathcal{L}^{\mathtt{reg}}_{\mathtt{val}} ( \a_{F}, \a_{D}, \y^*) ,  \\
\ s.t.  \ \y^* = \mathop{ \arg \min}\limits_{ \y} \ \mathcal{L}^{\mathtt{reg}}_{\mathtt{tr}} ( \y, \a_{F}, \a_{D}), 
\end{array}
\end{array}
\label{eq:M}
\end{equation}
where $\mathcal{L}^{\mathtt{reg}}_{\mathtt{val}}$ and $\mathcal{L}^{\mathtt{reg}}_{\mathtt{tr}}$ are the registration losses~\eqref{eq:regloss} evaluated on the validation and training sets for a scenario, respectively. 
Subsequently, we fix the network architecture $\a$ and hence resolve the bi-level optimization of the hyper-parameter $\x$ and network parameters $\y$:
\begin{equation}
\min\limits_{\x} \ \mathcal{L}^{\mathtt{seg}}_{\mathtt{val}}(\x,\y^*), \ s.t. \ \y^* = \arg\min\limits_{\y} \mathcal{L}^{\mathtt{reg}}_{\mathtt{tr}} ( \y, \x), 
\label{eq:lamda_op}
\end{equation}
where we evaluate the segmentatin loss~\eqref{eq:segloss} on the validation set for the scenario of interest.


Algorithm~\ref{alg:optimization} summarizes the detailed optimization procedure where we adopt the gradient-based strategy for both bi-level optimization problems~\eqref{eq:M} and~\eqref{eq:lamda_op}.
\textcolor{update}{In Stage 1, we search for $\alpha_F$ with fixed $\alpha_D$ which corresponds to a super-network of feature extractor and a fixed deformation model whose network parameters are $\y=\lbrace\y_{SF}, \y_D\rbrace$. The super-network contains all the candidate operations. In Stage 2, we search for $\alpha_D$ with optimal $\alpha_F$ which corresponds to a fixed feature extractor and a super-network of deformation model whose network parameters are $\y=\lbrace\y_F, \y_{SD}\rbrace$. In Stage 3, we search for $\x$ and we warm-start the search by updating \bm{$w$} from scratch with the optimal $\a$. Step 1 and Step 8 in Algorithm~\ref{alg:optimization} means randomly initializing model weight $\y$. Our optimization procedure includes a three-stage
optimization, where the optimization of the current stage relies on the optimal solution
of the previous stage.} We calculate the gradients of $\a_{F}$, $\a_{D}$ and $\x$ via one-step first-order approximation~\cite{LiuSY19,9551747} for the steps 3-5, 10-12 and 17-19 in Alg.~\ref{alg:optimization}~\footnote{We omit the derivations of the gradients due to page limit, and our codes are available at https://github.com/Alison-brie/AutoReg.}, and the parameter $\epsilon$ is set to be a small scalar value equal to the learning rate.

\section{Experimental Results and Analysis}\label{sec:experiment}
This section first provides detailed experimental settings and then presents ablation studies that validate the effectiveness of individual modules of the proposed method. Finally, we compare popular conventional and recent deep-based approaches on several data sets covering various scenarios. 
%

\subsection{Training Details}\label{sec:methodTraining}
Our automated registration contains two stages,~\emph{i.e.}, train-search and train-evaluation. In the first train-search stage, we search for network architecture according to Stage 1 and Stage 2 in Algorithm 1. We search for training objectives with respect to a validation dataset given a specific scenario according to stage 3 in Algorithm 1.
We initialize all the operation variables $\a$ to zeros indicating equal possibilities over all operation candidates in the beginning. 
Then, we fix all $\x$ to search for $\a_{F}$ according to steps 2-7. Next, we randomly initialize the weight and the search for $\a_{D}$ is according to steps 9-14. Finally, we warm-start the model weight and search for $\x$ with searched $\a_{F}$ and $\a_{D}$.
We use Adam~\cite{PaszkeGMLBCKLGA19} as the optimizer with the learning rates $1\times10^{-4}$ and $4\times10^{-3}$ for $\a$ and $\x$, respectively. 
The searching procedure for $\a$ and $\x$ stops when they remain stable in 10 consecutive epochs.
%
Subsequently, we update the network weights $\y$ given the searched $\a$  and $\x$ for 15 epochs, and then train all the parameters for additional 30 epochs.
In the train-evaluation stage, we train the weights $\y$ of the registration network given in the train-search stage for 200 epochs with the batch size set to 1 on the training set of the scenario. The learning rate is set to $1\times10^{-4}$ for the Adam optimizer. The network predicts a half-resolution deformation field and we up-sample it via linear interpolation, yielding the final full resolution deformation field for evaluation.
All the experiments were performed in Pytorch on 3.20GHz Intel(R) i7-8700 CPU with 32GB RAM and a NVIDIA TITAN XP GPU. 

\subsection{Data Preparation and Evaluation Metrics}
We applied AutoReg to a wide range of MRI registration scenarios from mono-modality (registering brain T1 images to an averaged T1 template, brain T1 to T1 images, brain T2 to T2 images, knee T1 to T1 images, and lung CT inspiration to expiration images) to multi-modality (registering brain T1 to T2 images\textcolor{update}{, brain T2 to T1 images}).

{\bf{Brain MR Image-to-Atlas registration.}} For image-to-atlas on T1 weighted brain MRI registration task, we use 528 scans from: ADNI~\cite{MUELLER200555}, ABIDE~\cite{Martino15}, PPMI~\cite{MAREK2011629} and OASIS~\cite{jocn.2007.19.9.1498}. We adopt the atlas in~\cite{BalakrishnanZSG19}. We divide our data into 377, 21 and 130 volumes for training, validation and testing. 
Standard pre-processing operations, \emph{e.g.,} motion correction, NU intensity correction, normalization, skull
stripping, with FreeSurfer~\cite{Fischl12} and affine normalization with FSL~\cite{WoolrichJPCMBBJS09} are conducted.
We also segment the testing data with FreeSurfer, resulting in 29 anatomical structures in each volume. All images are cropped to size of $160 \times 192 \times 224$ with 1 mm isotropic resolution.
For evaluation, all test MRI scans are anatomically segmented with Freesurfer to extract 30 anatomical structures.

{\bf{Brain T1-to-T1 registration.}} For the image-to-image case, the target images are randomly selected from PPMI~\cite{MAREK2011629} dataset. We divide our data into 65, 10 and 59 volumes for training, validation and testing. All images are cropped to size of $160 \times 192 \times 224$ with 1 mm isotropic resolution.

{\bf{Knee T1-to-T1 registration.}} We employ knee MRIs from the Osteoarthritis Initiative~\footnote{https://nda.nih.gov/oai/} with corresponding segmentations of femur and tibia as well as femoral and tibial cartilage~\cite{AmbellanTEZ19}. 
We divide images into 377, 21 and 130 volumes for training, validation and testing.  All images are resampled to isotropic spacing of 1mm, in size of $160 \times 160 \times 160$.

{\bf{Brain T2-to-T1 registration.}} For multi-modal registration, we utilize 135 cases from BraTS18~\footnote{https://www.med.upenn.edu/sbia/brats2018.html} and ISeg19~\footnote{https://iseg2019.web.unc.edu/} datasets, and each case includes two image modalities: T1 and T2 brain images with the size of $160 \times 160 \times 160$. Among them, 10 cases have segmentation ground truth. The data set is split into 115, 10 and 10 for train, validation and test.  As most T1 and T2 images are already aligned, we randomly choose one T1 scan as the atlas and try to register T2 scans to it.

\textcolor{update}{{\bf{Brain T1-to-T2 registration.}} We randomly choose one T2 scan as the atlas and try to register T1 scans to T2. The data set is split into 115, 10 and 10 for train, validation and test. All T1 and T2 data are from the same dataset used in the Brain T2-to-T1 registration experiment.}

{\bf{Brain T2-to-T2 registration.}} For the T2 weighted image-to-image case, the target images are randomly selected from T2 brain images in the above multi-modal datasets, resulting in 122 training pairs, 90 validation pairs and 59 test pairs. All T2 data are from the same dataset used in the Brain T2-to-T1 registration experiment.

{\bf{Lung CT inspiration-expiration registration.}} We download CT lung images from learn2reg challenge~\footnote{ http://doi.org/10.5281/zenodo.3835682}, including 20 training and 10 test images in size of $192 \times 192 \times 208$ with segmentation annotations. We take expiration data as target images and inspiration as source images.

\begin{table}[!t]
	\footnotesize
	\centering
	\caption{ Quantitative comparisons in terms of running time and memory usage on brain MR datasets of size $ 160 \times 192 \times 224$. }
	\begin{tabular}{|m{2.8 cm}<{\centering}| m{1.5cm}<{\centering}|m{3.6cm}<{\centering}|}
		\hline
		- &  Training    & Auto-search + Training      \\
		\hline
		Runtime (Hour) & 23    & 71    \\ 
            \hline
           Memory Usage (\#MB) & 6041   & 11143   \\ 
		\hline
	\end{tabular}
	\label{tab:Computationaltime}
\end{table}

We adopt the average Dice score~\cite{Dice} across a representative set of structures over all testing pairs as the evaluation metric. To evaluate the smoothness of the deformation field, we compute the Jacobian matrix and count all the folds~\cite{Ashburner07}.
Furthermore, NCC, as an auxiliary metric, is employed to verify alignment performance.
Both higher Dice scores and NCC values indicate more accurate registration, a lower variance reveals our method is more stable and robust.

\subsection{Ablation Study}

We first assess how the computational cost of our search compares to the standard approach.   Then, we investigate how our auto-search works across diverse registration tasks.
Furthermore, we demonstrate the effect that this strategy can have on existing registration networks.


\subsubsection{Computational cost} 
As shown in Table.~\ref{tab:Computationaltime}, 
we first list the running time and memory usage of traditional training and training with auto-search. It takes about 3 days for the proposed method to get a scenario-oriented registration model. We trade memory usage for better performance as the memory usage during the training search phase is twice that of traditional training.
Traditional training baseline typically involves training many separate models with various hyper-parameter configurations (architecture designs and hyper-parameters in loss function). And its computational cost mainly depends on the number of combinations of hyper-parameter configurations, while the number is usually set much larger than 10, which is computationally prohibitive. 
For models with many hyper-parameters, the value would be even more significant. Whereas the proposed automated searching strategy could drastically alleviate computation burdens, highly efficient than manual search by orders of magnitude.

\begin{table*}[!t]
	\centering
	\caption{ Ablation analysis of automatic learning across different kinds of registration scenarios in terms of Dice scores and NCC. The best result is in red whereas the second best one is in blue.}
	\footnotesize
	\begin{tabular}{|m{1.8cm}<{\centering}| m{.8cm}<{\centering}| m{1.8cm}<{\centering}| m{1.8cm}<{\centering}| m{1.8cm}<{\centering} |  m{1.8cm}<{\centering} |
	m{1.8cm}<{\centering} |
	m{1.8cm}<{\centering} |}
		\hline
		\multicolumn{2}{|c| }{$\y$ }       &  $\times$    & $\checkmark$   & $\checkmark$  &   $\checkmark$ &  $\checkmark$ &  $\checkmark$\\
		\multicolumn{2}{|c| }{$\a_{F}$}       &  $\times$   & $\times$  & $\checkmark$  &   $\times$ & $\times$ &  $\checkmark$\\
		\multicolumn{2}{|c| }{$\a_{D}$}       &  $\times$   & $\times$  & $\times$  &    $\checkmark$ &  $\times$ &  $\checkmark$\\
		\multicolumn{2}{|c| }{$\x$}       &  $\times$   & $\times$  & $\times$  &    $\times$ &  $\checkmark$  &  $\checkmark$ \\
		
		\hline
		\multirow{2}*{ADNI} & Dice  &  0.745 $\pm$ 0.027 & 0.754 $\pm$ 0.024  &  \multicolumn{1}{>{\columncolor{mycyan}}c}{\textcolor{blue}{0.763 $\pm$ 0.020}} &  0.758 $\pm$ 0.024 &  0.757 $\pm$ 0.021   &  \textcolor{red}{0.774 $\pm$ 0.022} \\
		~ & NCC    & 0.217 $\pm$ 0.007 & 0.221 $\pm$ 0.007 & \multicolumn{1}{>{\columncolor{mycyan}}c}{\textcolor{blue}{0.234 $\pm$ 0.006}}  &  0.225 $\pm$ 0.006  & 0.229 $\pm$ 0.006  & \textcolor{red}{0.258 $\pm$ 0.005} \\  \hline
		\multirow{2}*{Brain T1-to-T1} & Dice  & 0.706 $\pm$ 0.038 & 0.750 $\pm$ 0.022  & \multicolumn{1}{>{\columncolor{mycyan}}c}{\textcolor{blue}{0.764 $\pm$ 0.024}}  &  0.761 $\pm$ 0.022 &  0.750 $\pm$ 0.027 &  \textcolor{red}{0.778 $\pm$ 0.023} \\
		~ & NCC    & 0.186 $\pm$ 0.011  & 0.233 $\pm$ 0.009 & \multicolumn{1}{>{\columncolor{mycyan}}c}{0.243 $\pm$ 0.011}  & 0.233 $\pm$ 0.011   &  \textcolor{blue}{0.251 $\pm$ 0.011}   & \textcolor{red}{0.258 $\pm$ 0.010} \\  \hline
		\multirow{2}*{Brain T2-to-T2} & Dice  & 0.438 $\pm$ 0.110  & 0.636 $\pm$ 0.008 & \multicolumn{1}{>{\columncolor{mycyan}}c}{\textcolor{red}{0.646 $\pm$ 0.010}}  & 0.640 $\pm$ 0.010 &  \textcolor{blue}{0.643 $\pm$ 0.010}  & \textcolor{red}{0.646 $\pm$ 0.010}\\
		~ & NCC    & 0.033 $\pm$ 0.002  & 0.132 $\pm$ 0.002 &  \multicolumn{1}{>{\columncolor{mycyan}}c}{\textcolor{blue}{0.138 $\pm$ 0.002}}   &  0.132 $\pm$ 0.002   &  \textcolor{blue}{0.138 $\pm$ 0.002}   & \textcolor{red}{0.143 $\pm$ 0.002} \\  \hline
		\multirow{2}*{Knee T1-to-T1} & Dice  & 0.393$\pm$ 0.107  & 0.588 $\pm$ 0.157  &  \textcolor{red}{0.616 $\pm$ 0.150}   &  \multicolumn{1}{>{\columncolor{mycyan}}c}{\textcolor{blue}{0.614 $\pm$ 0.119} }&  0.606 $\pm$ 0.154  &  \textcolor{red}{0.616 $\pm$ 0.150} \\
		~ & NCC    & 0.140 $\pm$ 0.012  & 0.234 $\pm$ 0.032 & 0.202 $\pm$ 0.023   & \multicolumn{1}{>{\columncolor{mycyan}}c}{\textcolor{blue}{0.242 $\pm$ 0.030}}  & 0.239 $\pm$ 0.030  & \textcolor{red}{0.267 $\pm$ 0.023} \\  \hline
		\multirow{2}*{Brain T2-to-T1} & Dice   & 0.407 $\pm$ 0.005 & 0.599 $\pm$ 0.007  & 0.599 $\pm$ 0.006  & 0.598 $\pm$ 0.008 &  \multicolumn{1}{>{\columncolor{mycyan}}c}{\textcolor{blue}{0.621 $\pm$ 0.004} } & \textcolor{red}{0.622 $\pm$ 0.007} \\
		~ & NCC    &  0.033 $\pm$ 0.001  & 0.107 $\pm$ 0.002  & 0.108 $\pm$ 0.002 & 0.107 $\pm$ 0.002 &  \multicolumn{1}{>{\columncolor{mycyan}}c}{\textcolor{red}{0.127 $\pm$ 0.002}}  & \textcolor{blue}{0.108 $\pm$ 0.001}  \\  \hline
	\end{tabular}
	\label{tab:ablation-tasks}
\end{table*}

\begin{figure*}[!htp]
	\footnotesize
    \centering
		\begin{tabular}{c@{\extracolsep{0.5em}}c@{\extracolsep{0.5em}}c@{\extracolsep{0.5em}}c}
			\includegraphics[width=0.222\linewidth]{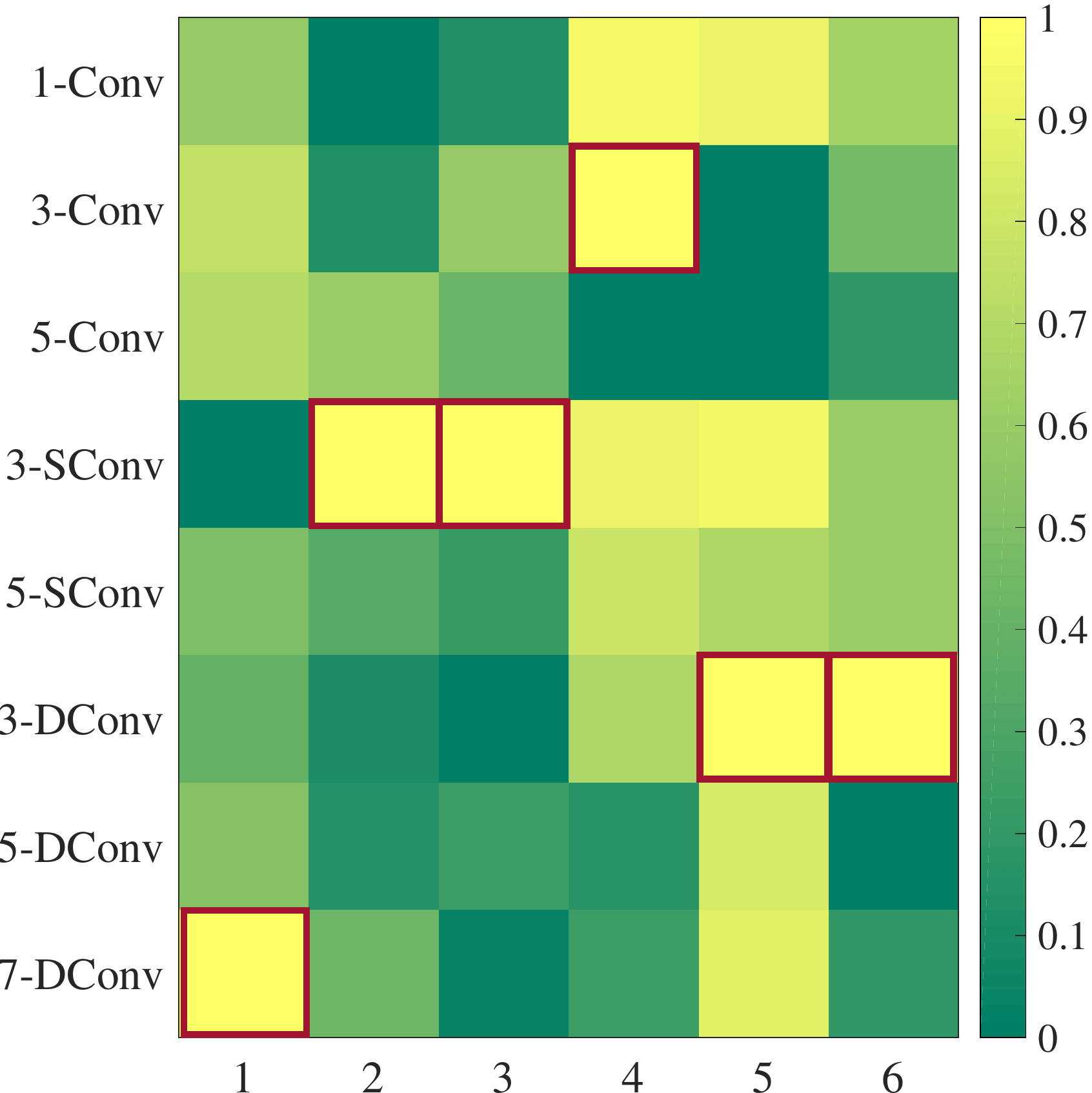}&
			\includegraphics[width=0.222\linewidth]{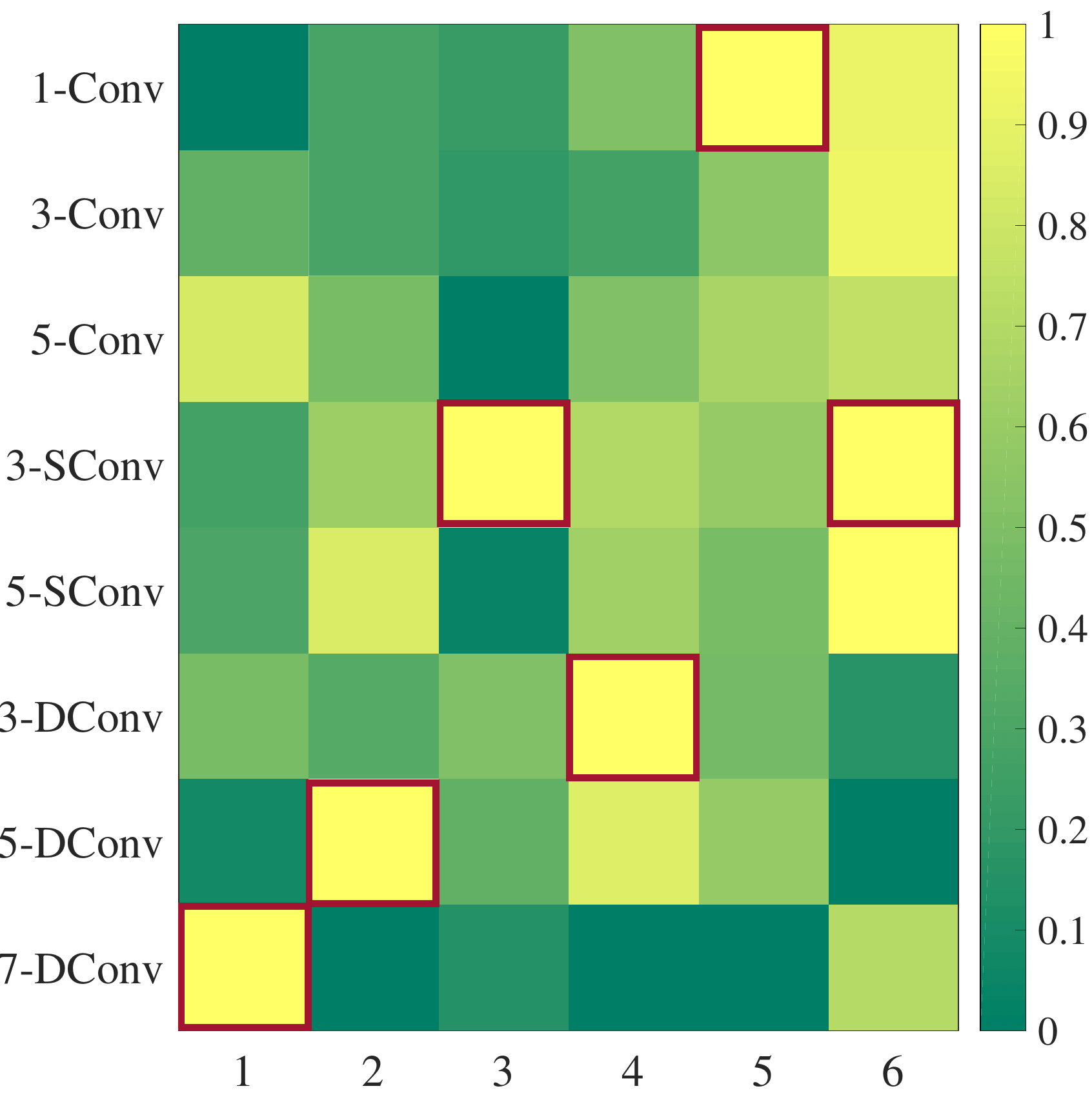}&
			\includegraphics[width=0.222\linewidth]{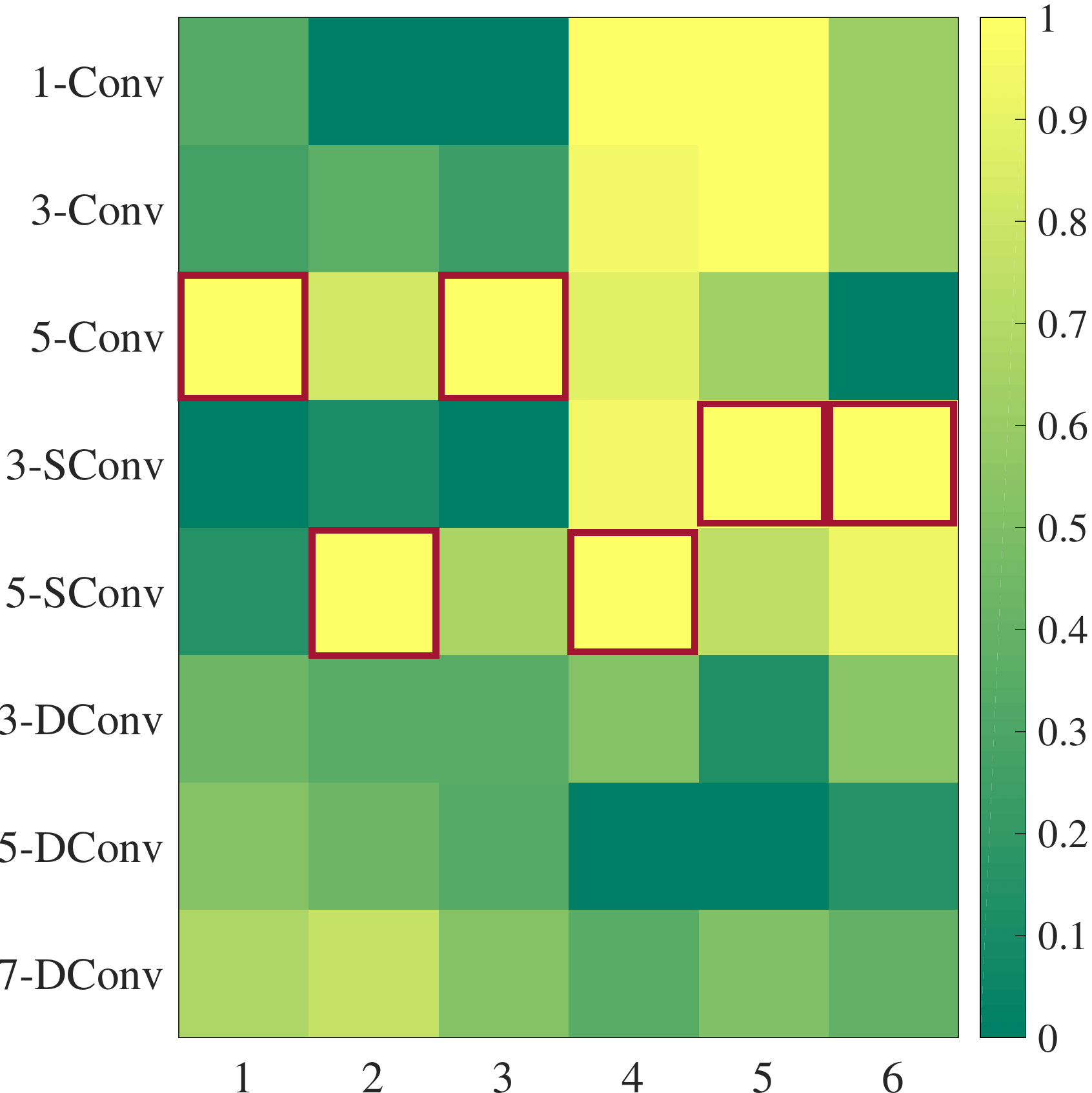}&
			\includegraphics[width=0.222\linewidth]{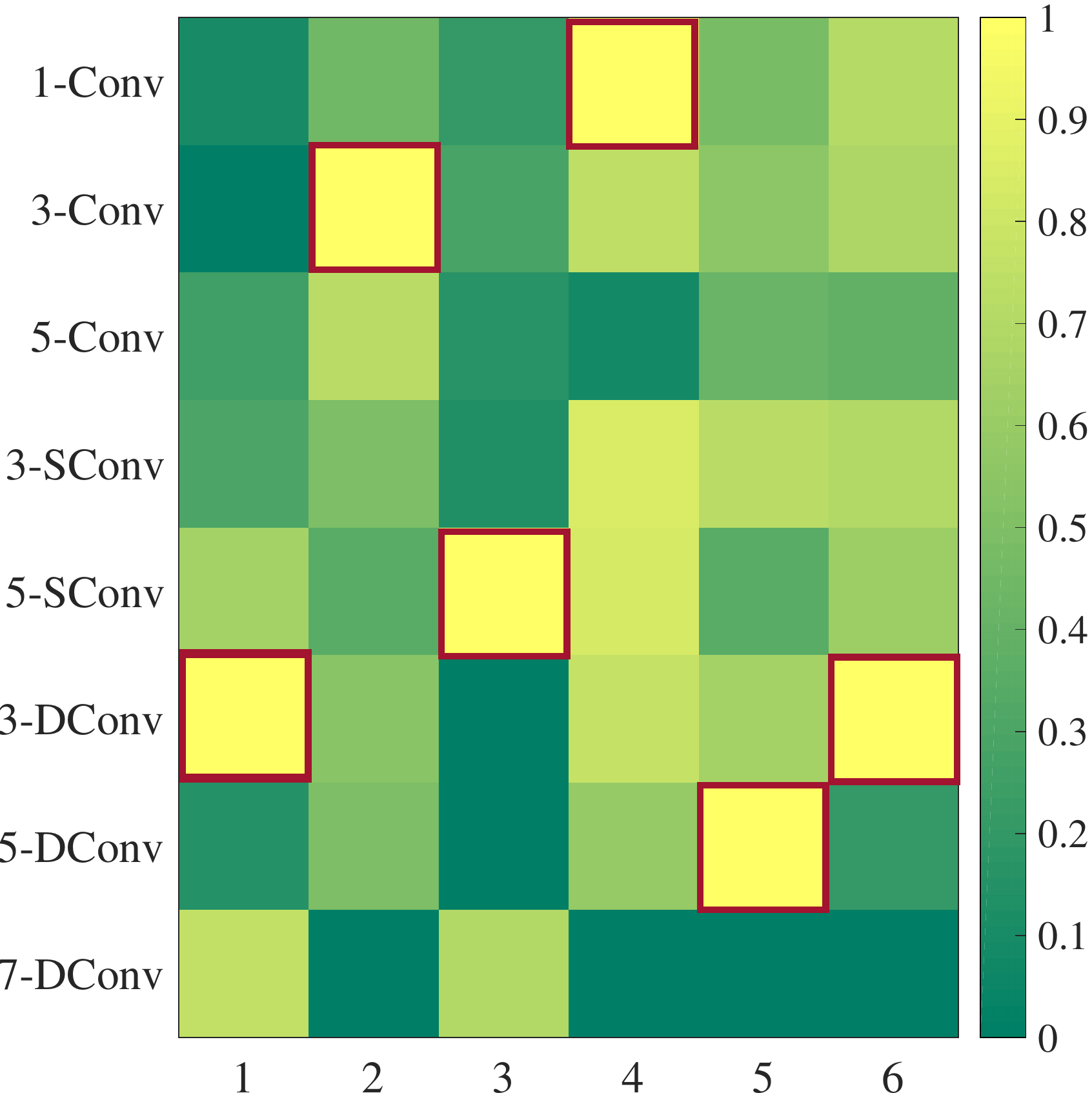}
			\\
			\includegraphics[width=0.222\linewidth]{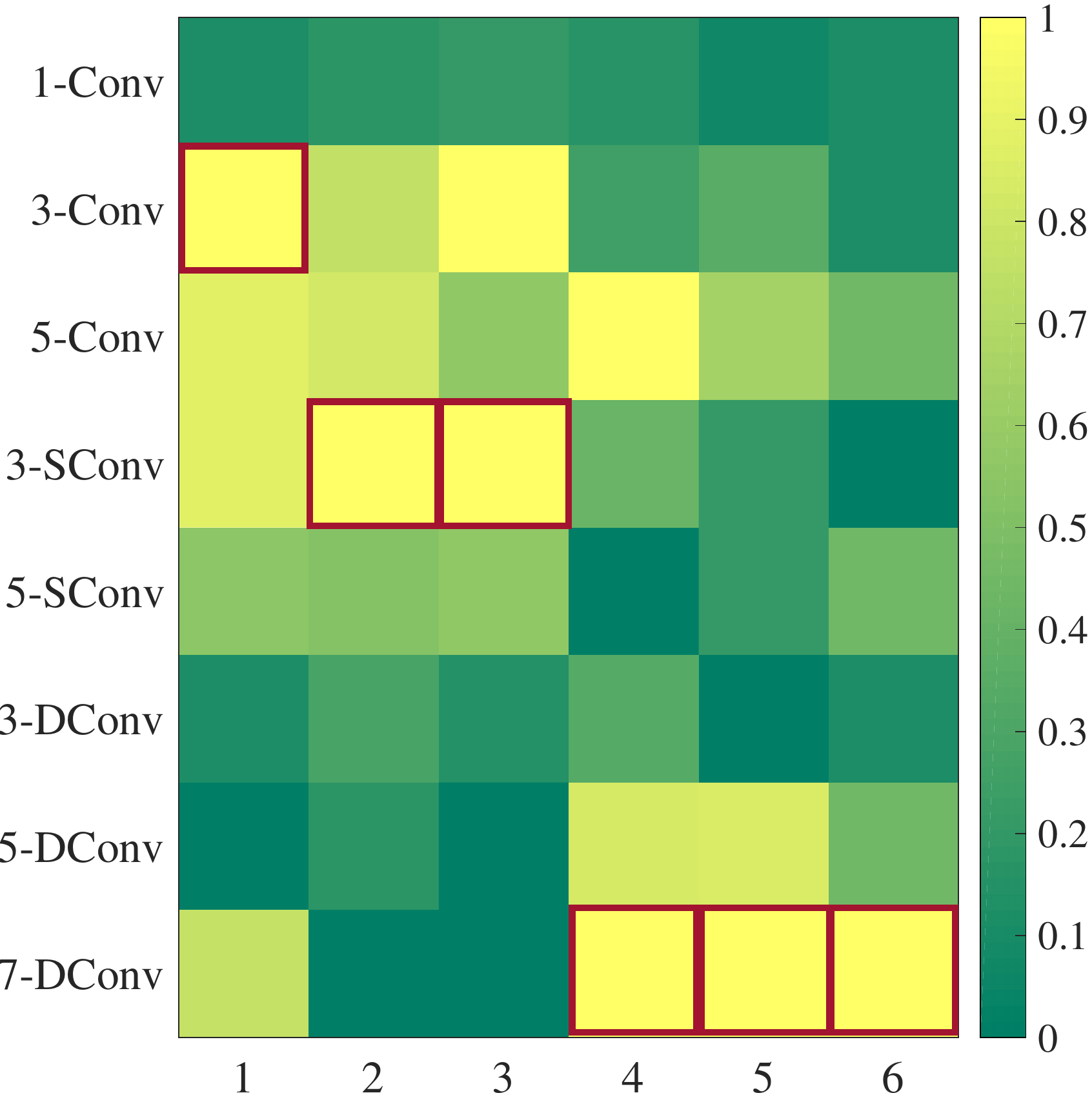}&
			\includegraphics[width=0.222\linewidth]{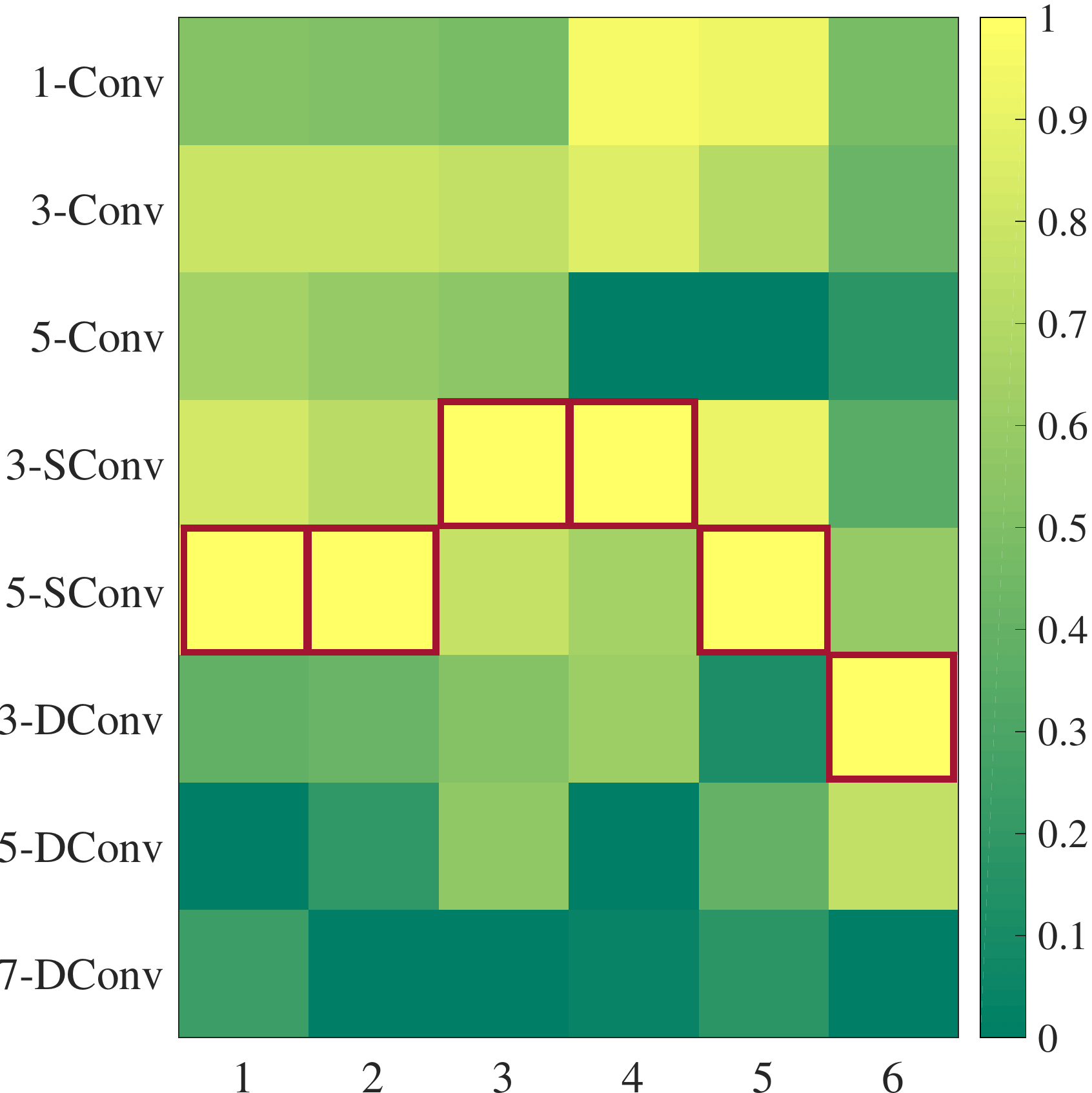}&
			\includegraphics[width=0.222\linewidth]{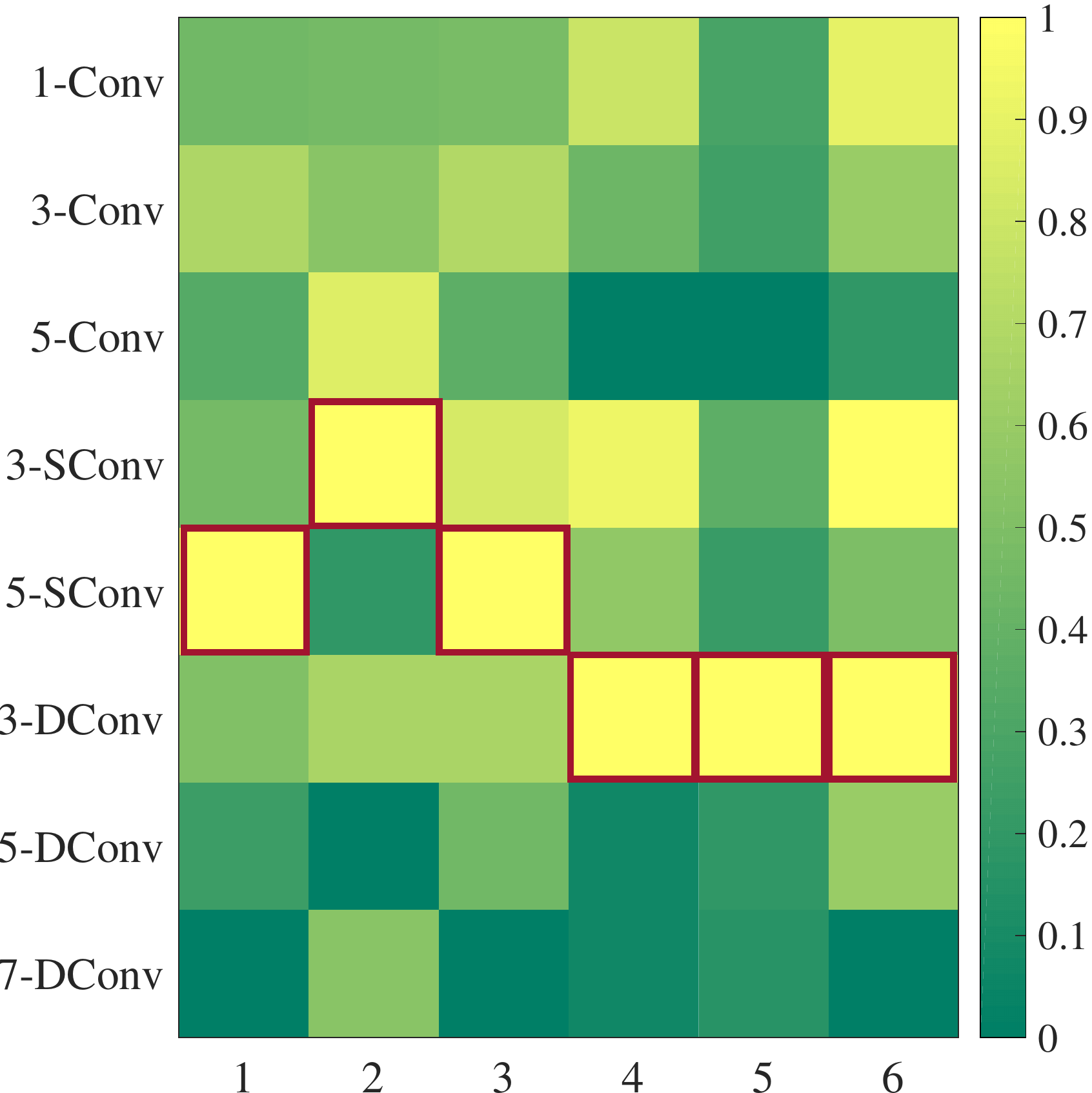}&
			\includegraphics[width=0.222\linewidth]{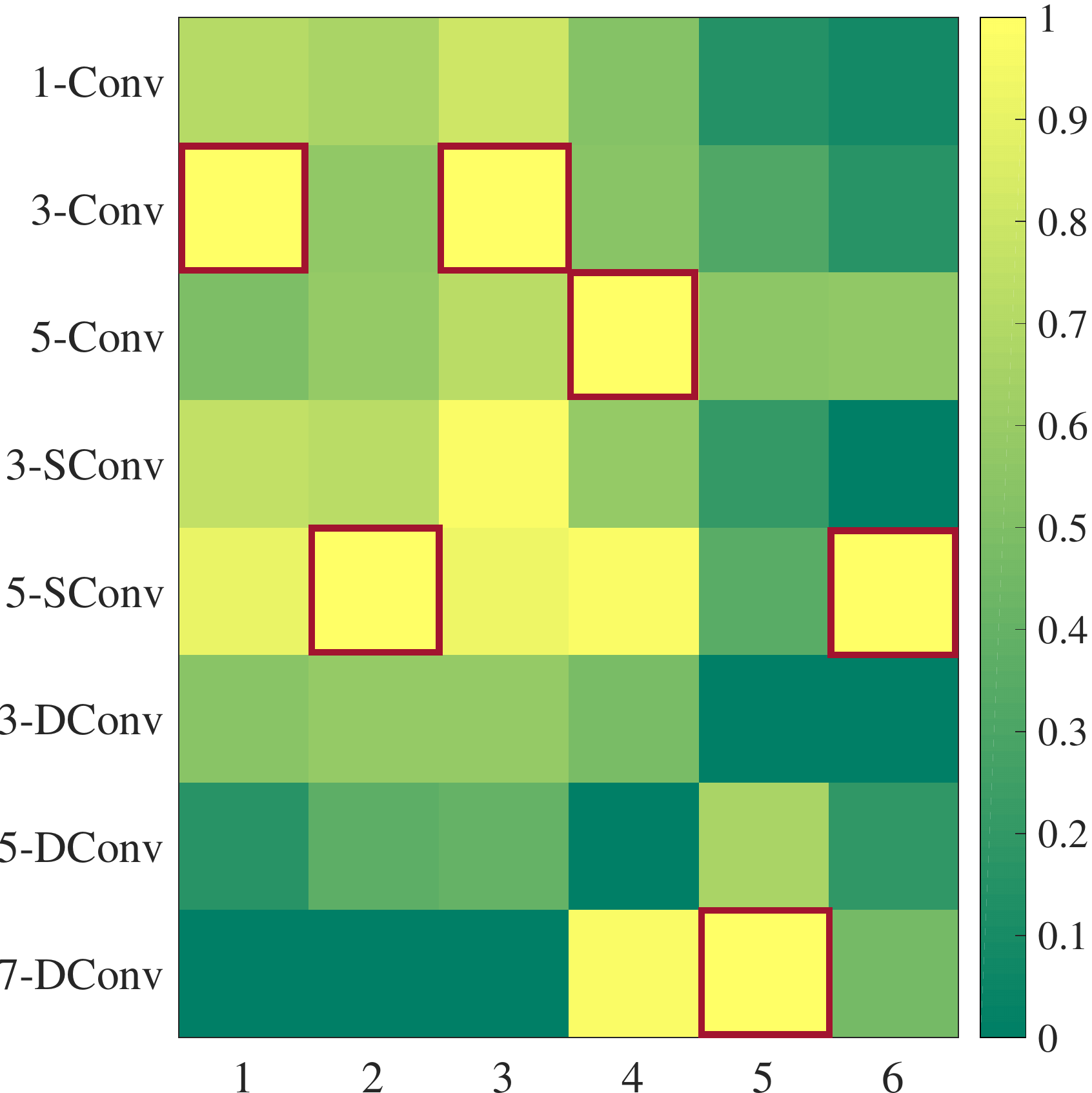}
			\\
			$W/ \x:=\{0.91, 1.6, 0.25, 0.15\}$	 &   	$W/ \x:=\{0.01, 0.2, 0.3, 0.04\}$ &  	$W/ \x:=\{0.27, 0.46, 0.35, 0.08\}$ &  	$W/ \x:=\{0.02, 0.1, 0.8, 0.3\}$   
			\\
				Brain T1-to-T1 & Brain T2-to-T2 & Knee T1-to-T1 &  Brain T2-to-T1 \\
		\end{tabular}	
	\caption{ Heatmaps of candidate architectures $\a := \{\a_F, \a_D\}$ in the last searching epoch. In each sub-figure, horizontal axis represents six convolution operations of two cells to be searched, and vertical axis represents eight operation types, red boxes indicate searched architectures with the highest score. 
		Top:  the searched $\a_F$. Bottom: corresponding searched $\a_D$.  
	} 
	\label{fig:architectures}
\end{figure*}

\subsubsection{Automatic learning across registration tasks}
To explore the influence of model weights, architectures, and training objectives in registration performance and verify the benefit from adaptive feature extraction cell, task-aware deformation estimation cell design, and scenario-oriented training objectives, we made groups of contrast experiments, covering diverse registration scenarios. 
Specifically, we first train a registration model on PPMI brain T1 MR image-to-atlas registration tasks, then take the model as initialization to adopt our auto-search strategy to register images of other different scenarios. The scenarios cover registration on another ADNI dataset, Brain T1-to-T1, Brain T2-to-T2, Knee T1-to-T1 and  Brain T2-to-T1 setups.

In Table.~\ref{tab:ablation-tasks}, we directly apply the model \textcolor{update}{(with $\a_F$, $\a_D$, $\x$ and $\y$)} on the PPMI dataset to other scenarios, corresponding to the 1st column result, showing poor performance. Whereas performances of re-trained models \textcolor{update}{(with task-specific model weight $\y$)} on these tasks correspond to the 2nd column.
%
We also demonstrate the performance of searched networks with auto-learned feature cells $\a_F$, deformation cells $\y_D$, and training objectives $\x$, corresponding to the 3rd to 5th columns. 
To fully capture the benefit of the proposed technique, we further report the increase in registration accuracy for cases where all hyperparameters are searched in the last column. We can observe that, \emph{firstly}, retraining the model for different alignment tasks will result in better performance. \emph{Secondly}, searched tailored architecture and training objectives largely improve numerical results, which means automatic learning combining training objectives, architectures and hyperparameters can achieves excellent alignment performance in different alignment scenarios.

Also, the models on the diagonal with a blue background perform second best which can be seen from Table.~\ref{tab:ablation-tasks}, providing meaningful indications and conclusions.
\emph{Firstly}, when transferring to another dataset or image contrast, feature extraction plays a dominant role in model performance. \emph{Secondly}, whereas transferring to another anatomical structure such as the knee data, regulating the deformation estimation section has a more significant impact on the performance of the model. \emph{Lastly}, adjusting the training objective plays a more important role in the performance of a registration network when transferring to multi-modal datasets.

\begin{table*}[t]
	\footnotesize
	\centering
	\caption{ \textcolor{update}{Comparison results in terms of Dice scores of different network architectures on multiple registration tasks.  }}
	\begin{tabular}{|m{2.2cm}<{\centering}| m{2.4cm}<{\centering}| m{2.4cm}<{\centering}| m{2.4cm}<{\centering}| m{2.4cm}<{\centering}|}
		\hline
		Method           & Brain T1-to-T1  & Brain T2-to-T2      & Knee T1-to-T1   & Brain T2-to-T1    \\
		\hline
		\textcolor{update}{Manual-Small}     & 0.700 $\pm$ 0.036  & 0.610 $\pm$ 0.009 & 0.395 $\pm$ 0.110  &  0.579 $\pm$ 0.005  \\
		\textcolor{update}{Manual-Medium}     & \textcolor{blue}{0.769 $\pm$ 0.025}  & \textcolor{blue}{0.636 $\pm$ 0.010} & 0.605 $\pm$ 0.131  &  \textcolor{blue}{0.617 $\pm$ 0.006} \\
		\textcolor{update}{Manual-Large}     & 0.761 $\pm$ 0.025  & 0.610 $\pm$ 0.009 & \textcolor{blue}{0.614 $\pm$ 0.091} & 0.613 $\pm$ 0.007 \\ 
		AutoReg    & \textcolor{red}{0.778 $\pm$ 0.023}  &  \textcolor{red}{0.646 $\pm$ 0.010} & \textcolor{red}{0.616 $\pm$ 0.150}  & \textcolor{red}{0.622 $\pm$ 0.007} \\ 
		\hline
	\end{tabular}
	\label{tab:ab_optimalstructure}
\end{table*}

\begin{figure*}[t]
	\centering 
	\begin{tabular}{@{\extracolsep{0em}}c@{\extracolsep{0.6em}}c@{\extracolsep{0.6em}}c@{\extracolsep{0.6em}}c@{\extracolsep{0.6em}}c}
		\includegraphics[width=0.178\textwidth]{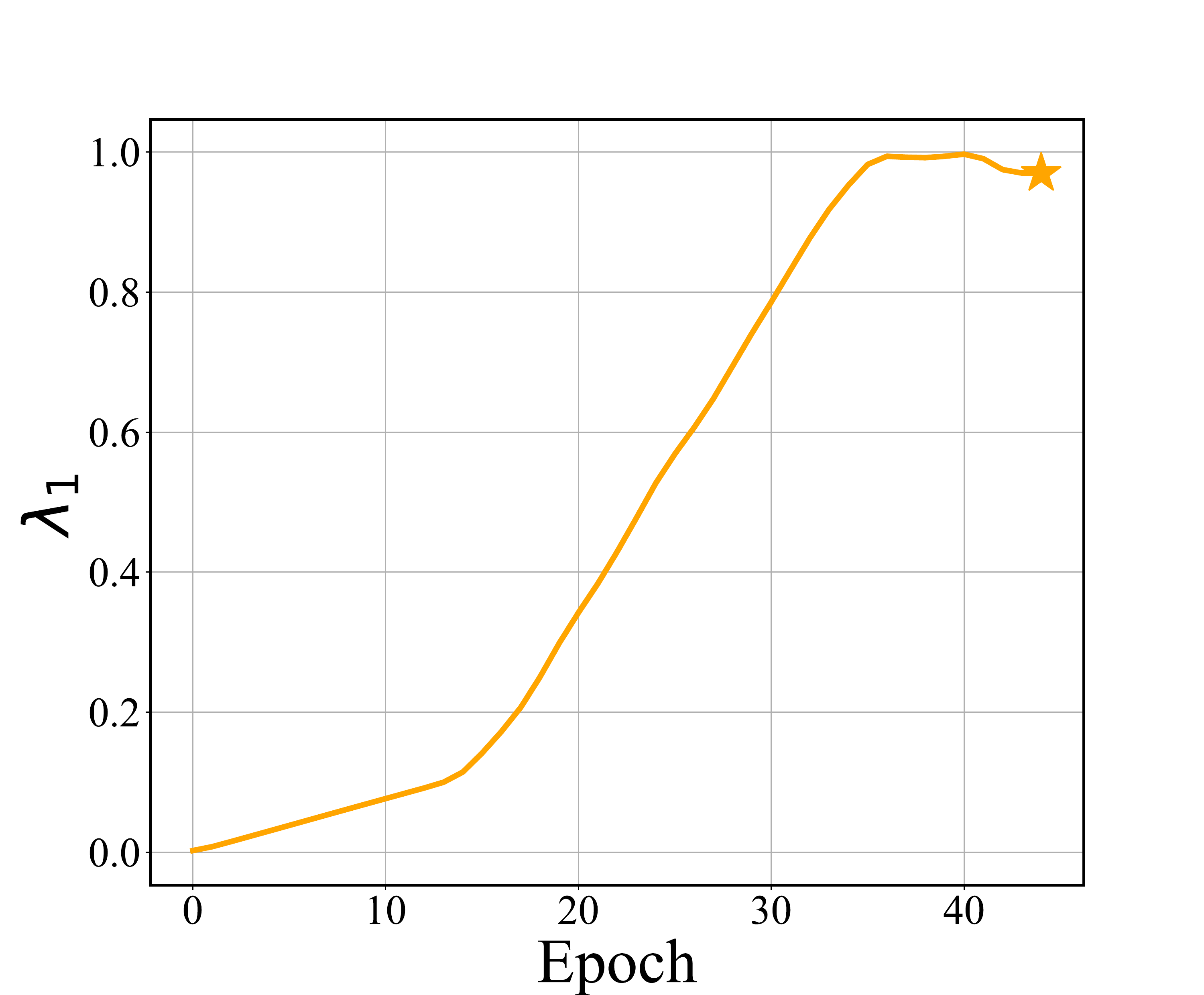}  &
		\includegraphics[width=0.178\textwidth]{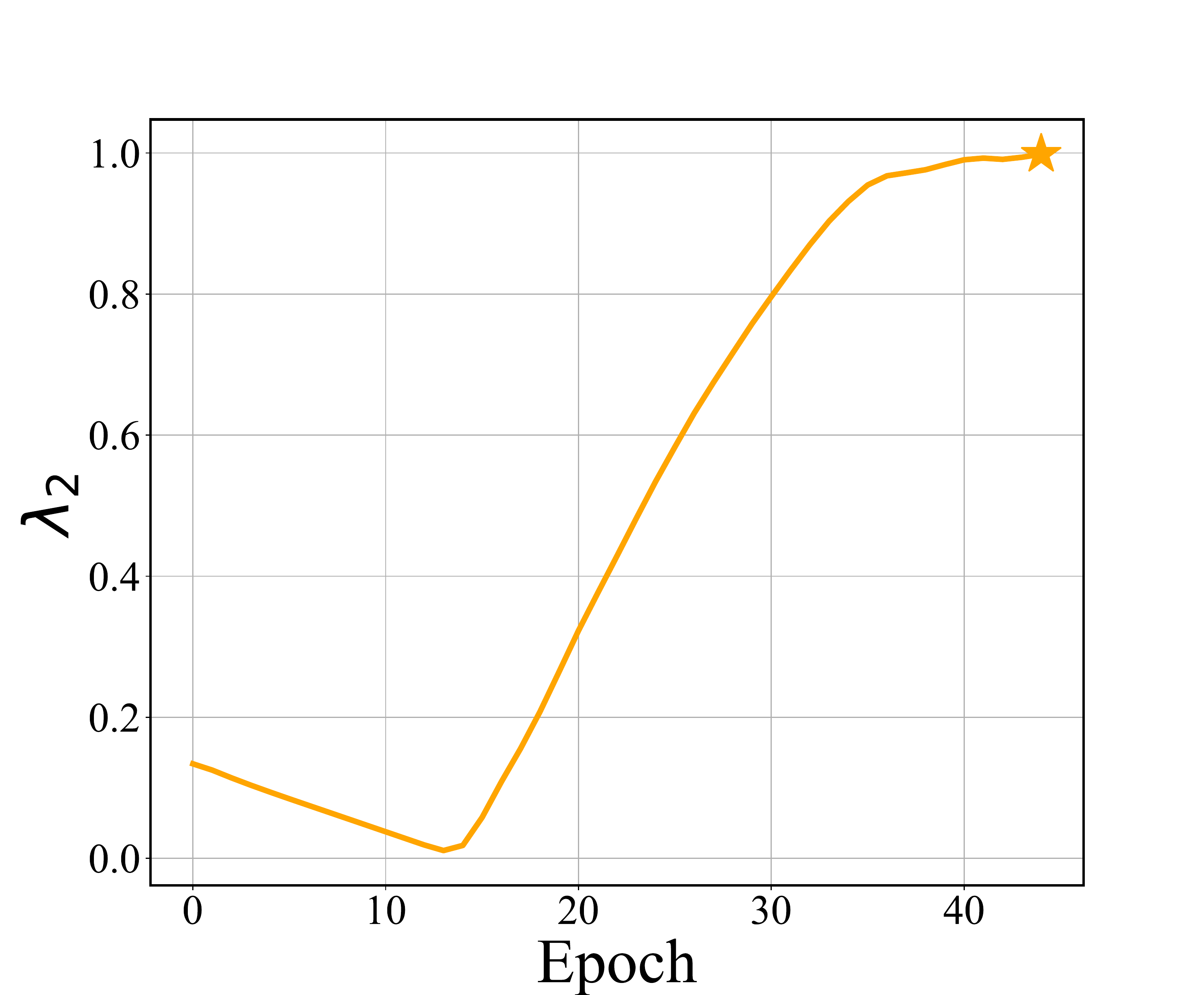}  &
		\includegraphics[width=0.178\textwidth]{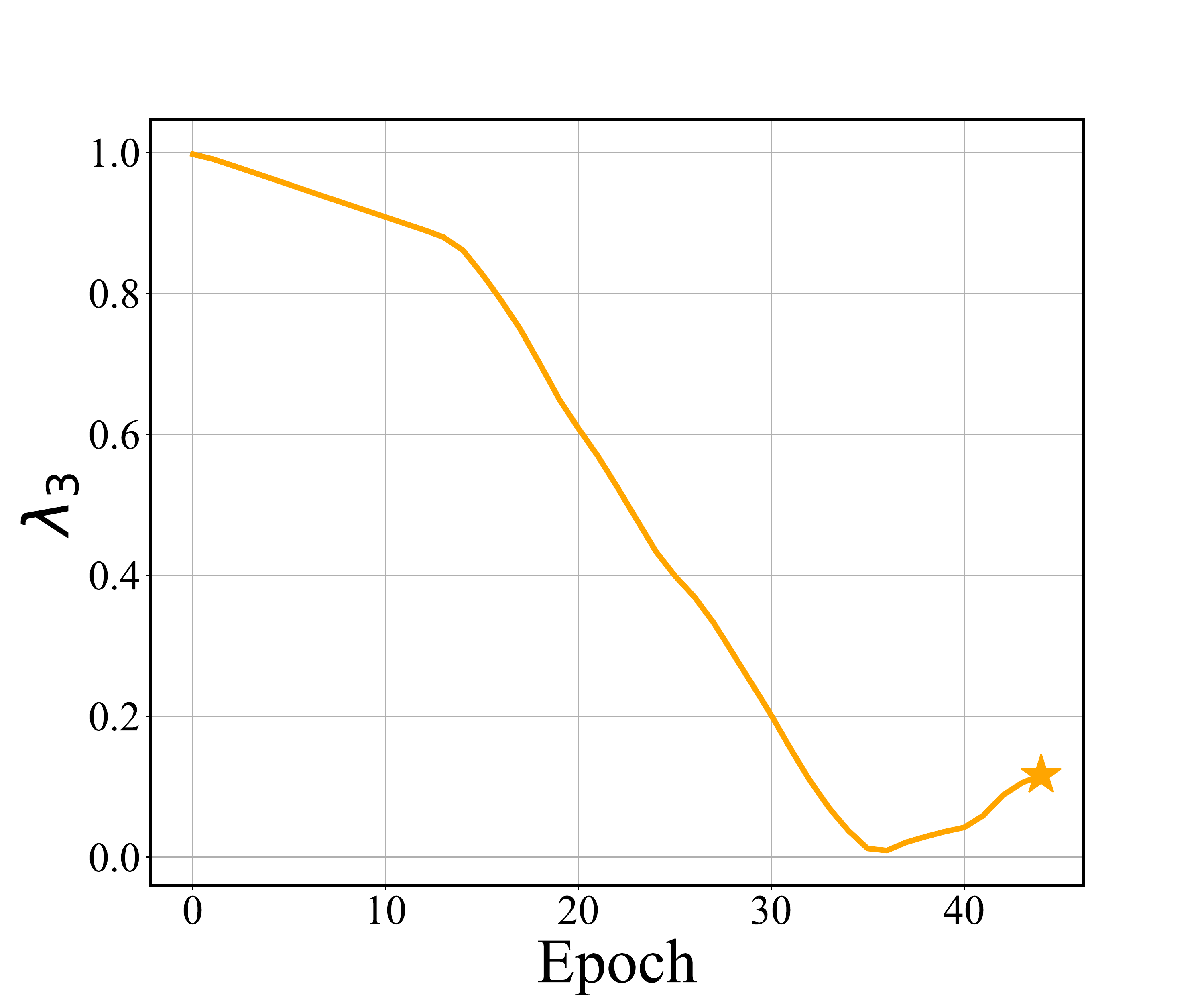}  &
		\includegraphics[width=0.178\textwidth]{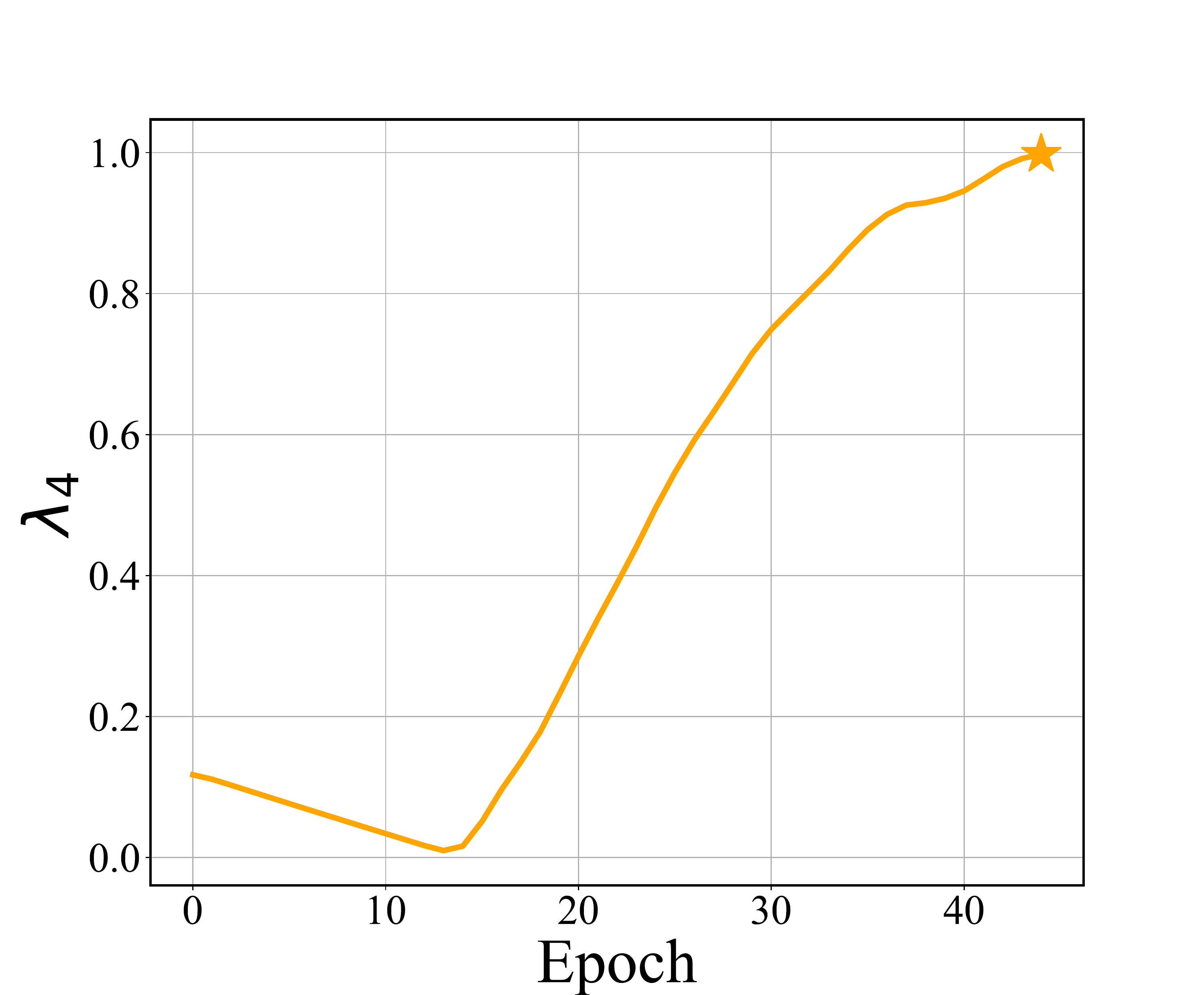}  &
		\includegraphics[width=0.178\textwidth]{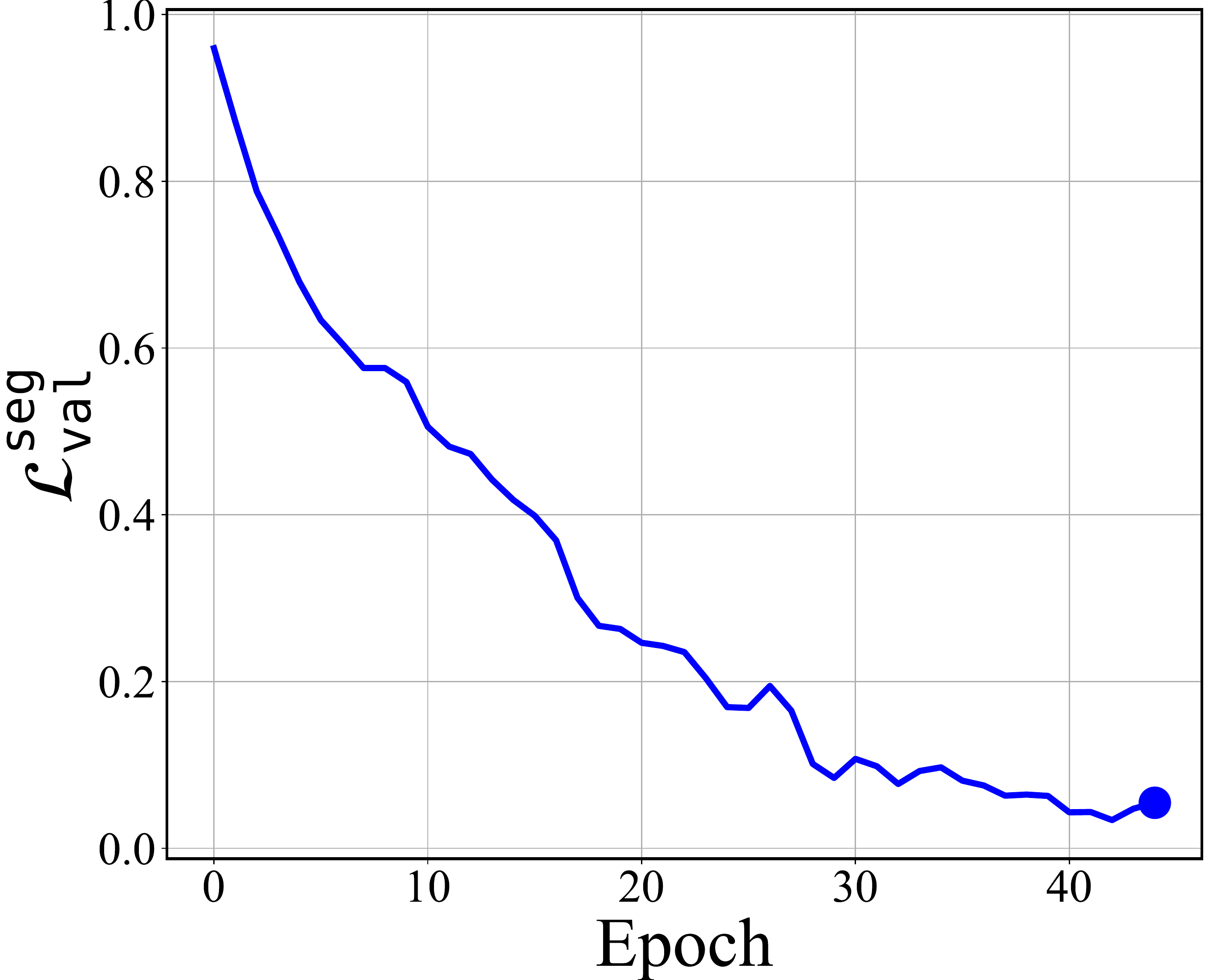}
	\end{tabular}
	\caption{ Hyperparameters values and loss values of hyperparameters optimization across training epoch on the Image-to-Atlas registration of brain MRI.  We rescale these values to [0,1]. As the epoch number increases, the change of values becomes smoother.} 
	\label{fig:lossconverge}
\end{figure*}

\begin{table*}[t]
	\footnotesize
	\centering
	\caption{ \textcolor{update}{Ablation experiments on our scenario-oriented objective design with NAS not involved in terms of Dice scores. }}
	\begin{tabular}{|p{2.2cm}<{\centering}| p{2.4cm}<{\centering}|p{2.4cm}<{\centering}| p{2.4cm}<{\centering}| p{2.4cm}<{\centering}|}
		\hline
		Application & W/ NCC & W/ MIND & W/ UNIFIED  & W/ UNIFIED*\\
		\hline	
		Image-to-Atlas  & 0.757 $\pm$ 0.014  & 0.737 $\pm$ 0.019 & \textcolor{blue}{0.758 $\pm$ 0.012} &  \textcolor{red}{ 0.764 $\pm$ 0.011} \\ 
		
		Brain T2-to-T1     & 0.591 $\pm$ 0.004 & \textcolor{blue}{0.601 $\pm$ 0.008}  & 0.587 $\pm$ 0.006 & \textcolor{red}{0.604 $\pm$ 0.005} \\ \hline
	\end{tabular}
	\label{tab:ab_optimalHO}
\end{table*}

\subsubsection{Searched architectures and hyperparameters}
From the searched architecture in Fig.~\ref{fig:architectures}, 
\emph{firstly}, we observe that in the case of the same convolution kernel size, our searched cells contain more SConv and DConv than ordinary convolution. Overall, 3-SConv has been selected the most times while 1-Conv and 3-Conv are selected the least.
\emph{Secondly}, we find that the first convolution of each feature-cell and the last convolution of each deformation-cell favor the convolution type with large receptive fields, while these correspond to the high-resolution positions in the registration network. We may also draw interesting conclusions: compared to the position at smaller resolutions, the ones on larger resolution prefer larger receptive fields such as high dilation rates and large kernels. 
\emph{Thirdly}, we can observe that 1-Conv only appears in the searched feature cells. It is reasonable because the number of feature maps often increases with the depth of the network whereas the 1-Conv can be used to offer a channel-wise pooling, often called feature map pooling or a projection layer. This simple technique can be used for dimensionality reduction, decreasing the number of feature maps whilst retaining their salient features, being a good choice for feature extraction. 
These observations provide us with meaningful and insightful cues for manually designing registration networks.
Fig.~\ref{fig:architectures} also gives the searched hyper-parameters under these registration scenarios. 
Obviously, we can observe that the optimal hyper-parameter varies substantially across registration tasks. For example, uni-modal tasks such as brain T1 MR image-to-image require a significantly different hyper-parameters value than multi-modal data.

\subsubsection{Optimality verification}

To verify the searched architectural design is optimal for different deformable registration tasks, we report the registration performance of the searched architecture compared to diverse architectural designs of existing deformable registration methods in Table.~\ref{tab:ab_optimalstructure}. To be specific, we explore how much of a difference the choice of particular architecture actually makes by comparing experimental results among the networks with all-1-Conv \textcolor{update}{(Manual-Small)}, all-3-Conv \textcolor{update}{(Manual-Medium)}, all-7-Conv \textcolor{update}{(Manual-Large)}, and searched architecture. 
%
\textcolor{update}{As Table.~\ref{tab:ab_optimalstructure} shows, the artificially designed all-3-Conv can achieve better performance than all-1-Conv and all-7-Conv in most scenarios. However, the all-7-Conv design is more suitable for the knee T1-to-T1 scenario, rather than all-3-Conv. In contrast, our searched ones beat all these manually designed architectures and obtain suitable convolution types with appropriate feature expression capabilities for different scenarios.}

Fig.~\ref{fig:lossconverge} gives the variability of hyperparameters values and loss values of hyperparameters optimization across training epoch. We normalize the hyperparameters by adding regularization to the magnitude of the hyperparameters.
As the figure shows, with the convergence of loss values, the change of hyperparameters values gradually stabilized. The yellow stars and blue dot indicate the optimal value as identified by automatic hyperparameter optimization.
As shown in Table.~\ref{tab:ab_optimalHO}, we compare the results of experiments using NCC loss, MIND loss, the unified loss with manually selected hyperparameters, and the unified loss using searched hyperparameters.
\textcolor{update}{Note that these results are obtained across the same network with all-3-Conv and NAS is not involved.}
Results indicate that the networks trained with the searched loss functions deliver accuracy on par or even superior to those with the handcrafted losses. 
When switching to multi-modal data, generally, manually loss function tuning will be executed, which requires many training runs. The proposed self-tuned training could auto-adapt to this new scene and achieve satisfying performance.

\begin{table*}[!t]
	\footnotesize
	\centering
	\caption{ \textcolor{update}{Ablation analysis of AutoReg strategy for VM models on multiple registration tasks in terms of Dice scores.} }
	\begin{tabular}{|m{2.2cm}<{\centering}| m{2.4cm}<{\centering}| m{2.4cm}<{\centering}| m{2.4cm}<{\centering}| m{2.4cm}<{\centering}|}
		\hline
		Method           & Brain T1-to-T1  & Brain T2-to-T2      & Knee T1-to-T1   & Brain T2-to-T1    \\
		\hline
		VM   & 0.757 $\pm$ 0.035      & 0.638 $\pm$ 0.012  &  0.440 $\pm$ 0.132  & 0.579 $\pm$ 0.013 \\
		VM + AutoReg   & 0.761 $\pm$ 0.010   & 0.640  $\pm$ 0.013 & 0.482 $\pm$ 0.151 & 0.596 $\pm$ 0.006 \\ 
		\hline
	\end{tabular}
	\label{tab:ab_strategy}
\end{table*}

\begin{figure*}[!t]
		\footnotesize
	\centering 
	\begin{tabular}{c@{\extracolsep{0.22em}}c@{\extracolsep{0.22em}}c@{\extracolsep{0.22em}}c}
		\includegraphics[width=0.206\textwidth]{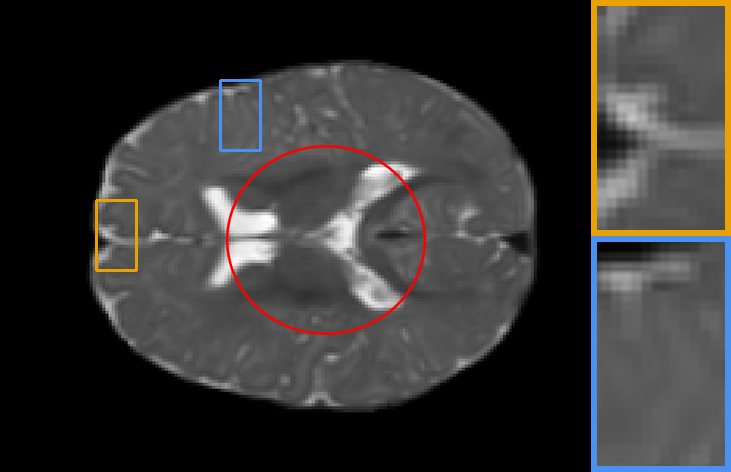}  &
		\includegraphics[width=0.206\textwidth]{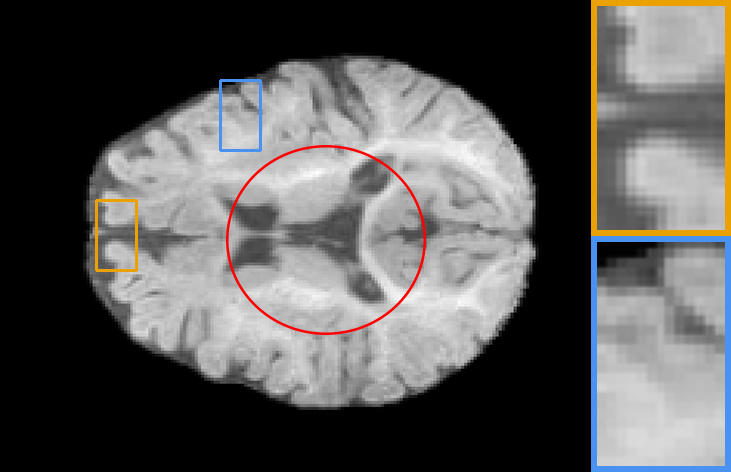}  &
		\includegraphics[width=0.206\textwidth]{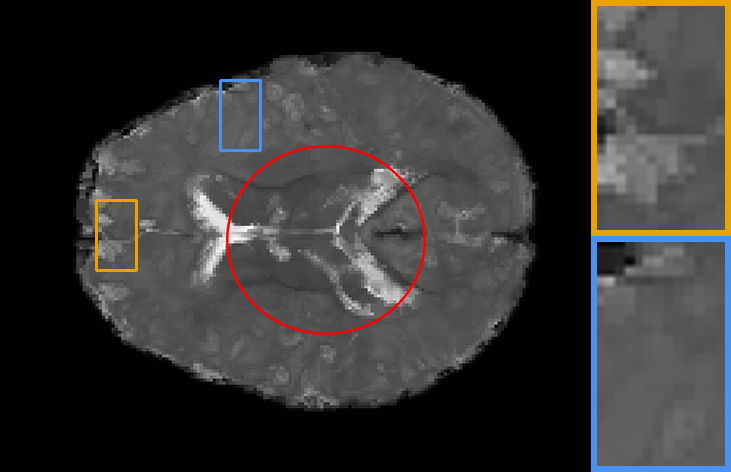}  &
		\includegraphics[width=0.206\textwidth]{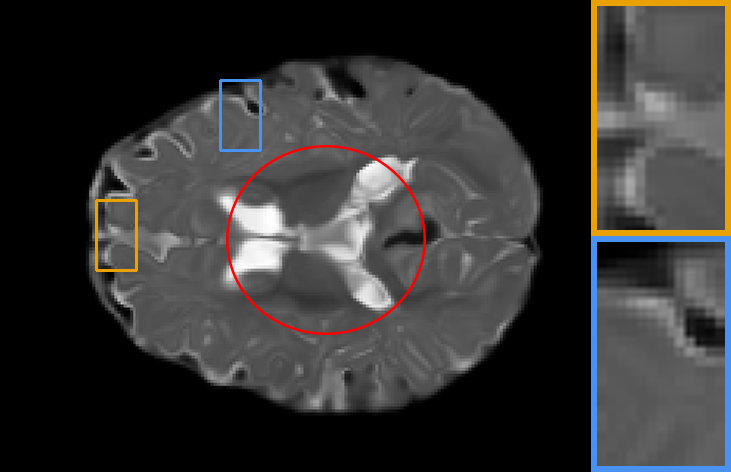} 
		\\
		\includegraphics[width=0.206\textwidth]{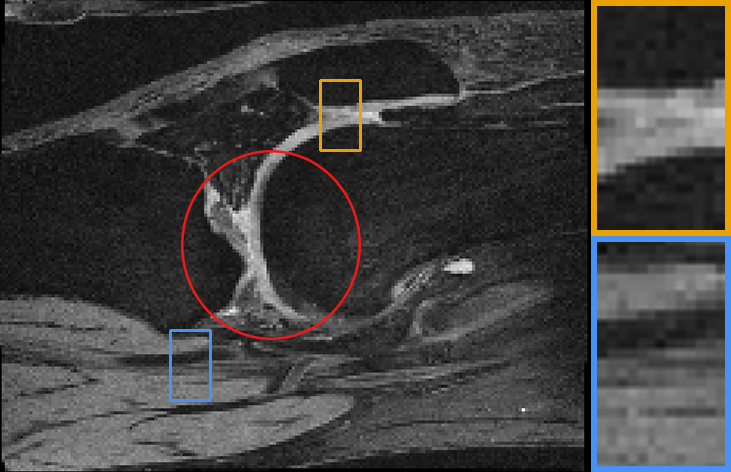}  &
		\includegraphics[width=0.206\textwidth]{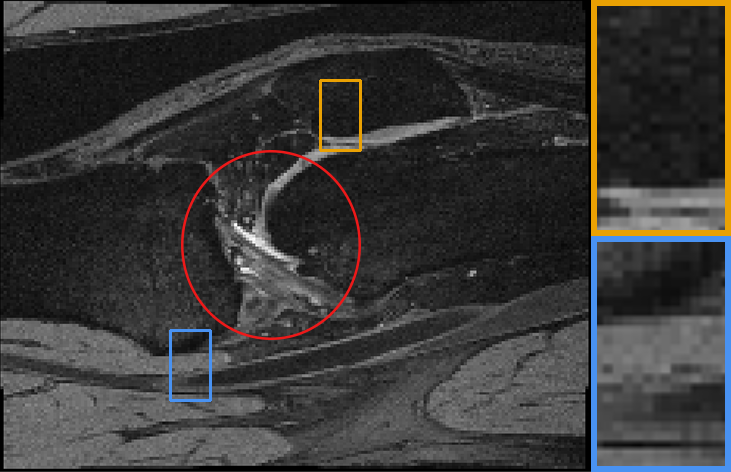}  &
		\includegraphics[width=0.206\textwidth]{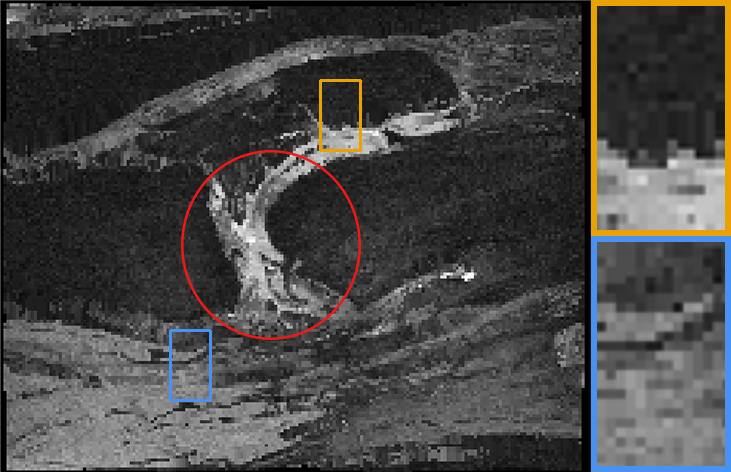}  &
		\includegraphics[width=0.206\textwidth]{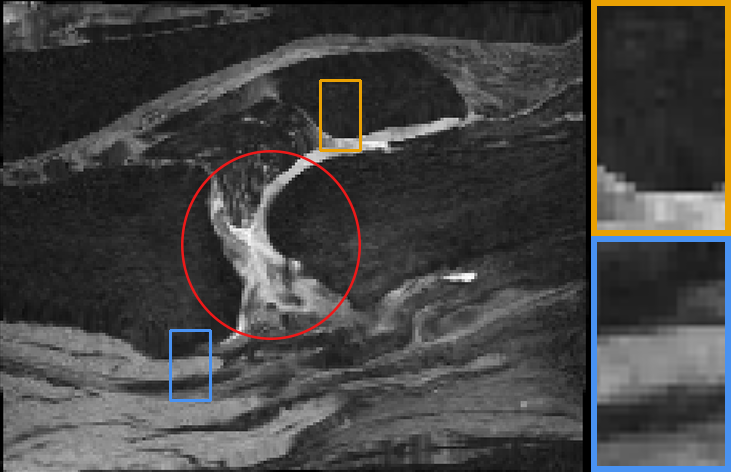}
		\\
		Source  &  Target 	&   VM  &   VM + AutoReg  \\
	\end{tabular}
	\caption{Sample result of registering different images on VM and autoreg strategy for VM model. Each row refers to an example registration case, including target, source and registered images. }
	\label{fig:ab_strategy}
\end{figure*}

\subsubsection{Generalizability analysis}

To further demonstrate the importance of applying auto-tuning methods across model types and not relying on previously-published hyperparameters for different applications and network designs, we involve the application of our AutoReg technique to the voxelmorph, a UNet-like registration model.
As quantitative and qualitative ablation comparisons in Table.~\ref{tab:ab_strategy} and Fig.~~\ref{fig:ab_strategy}, our AutoReg startegy could effectively facilitate the tuning of existing highly-parameterized registration models.

\begin{table*}[t]
	\footnotesize
	\centering
	\caption{ Comparison results of image-to-atlas registration in terms of Dice scores of different methods on multiple datasets.  }
	\begin{tabular}{|m{2.6cm}<{\centering}|  m{2.8cm}<{\centering}| m{2.8cm}<{\centering}| m{2.8cm}<{\centering} |m{2.8cm}<{\centering} |}
		\hline
		Method & ABIDE         & ADNI    & OASIS   & PPMI     \\
		\hline
		Initial     & 0.624 $\pm$ 0.024  & 0.571 $\pm$ 0.049  &  0.580 $\pm$ 0.028  & 0.610 $\pm$ 0.033     \\
		ANTs    & 0.728 $\pm$ 0.029  & 0.761 $\pm$ 0.021  &  0.765 $\pm$ 0.010  & 0.778 $\pm$ 0.013   \\
		NiftyReg  & 0.747 $\pm$ 0.026  & 0.737 $\pm$ 0.035   & 0.748 $\pm$ 0.017  & 0.765 $\pm$ 0.015     \\ 
		VM    & 0.754 $\pm$ 0.016  & 0.761 $\pm$ 0.024 & 0.765 $\pm$ 0.010  & 0.775 $\pm$ 0.013   \\
		VM-diff  & \textcolor{blue}{0.773 $\pm$ 0.009}  & 0.768 $\pm$ 0.020   & 0.757 $\pm$ 0.011  & 0.781 $\pm$ 0.011     \\
		MultiPropReg  & 0.764 $\pm$ 0.016  & \textcolor{blue}{0.773 $\pm$ 0.017}  & \textcolor{blue}{0.777 $\pm$ 0.006}  & \textcolor{blue}{0.787 $\pm$ 0.010}   \\
		Our AutoReg   & \textcolor{red}{0.784 $\pm$ 0.008}  & \textcolor{red}{0.774 $\pm$ 0.022}   & \textcolor{red}{0.788 $\pm$ 0.010}  & \textcolor{red}{0.788 $\pm$ 0.012} \\
		\hline 
	\end{tabular}
	\label{tab:compare_dice}
\end{table*}

\begin{table*}[t]
	\footnotesize
	\centering
	\caption{ Comparison results in terms of regularity of the transformation of different methods on multi brain MR datasets.  }
	\begin{tabular}{|m{2.6cm}<{\centering}| m{2.6cm}<{\centering}| m{2.6cm}<{\centering}| m{2.6cm}<{\centering} |m{2.6cm}<{\centering} |}
		\hline
		Method & ABIDE         & ADNI         & OASIS   & PPMI     \\
		\hline
		ANTs  & 27288 $\pm$ 3411  & 30737 $\pm$ 9537 & 29094 $\pm$ 8772 & 25452 $\pm$ 6490     \\
		NiftyReg & 11.4 $\pm$ 13  & 572.2 $\pm$ 878 &  416 $\pm$ 416 & 314.3 $\pm$ 353    \\
		VM & 28861 $\pm$ 1616  & 33047 $\pm$ 4667  & 32029 $\pm$ 3498 & 30192 $\pm$ 3375    \\
		VM-diff & 25 $\pm$ 13.1  & 43 $\pm$ 33.1 &  35 $\pm$ 13 & 29 $\pm$ 24   \\
		MultiPropReg  &  \textcolor{blue}{1 $\pm$ 0.9} & \textcolor{blue}{5.3 $\pm$ 6}   &  \textcolor{blue}{6.2 $\pm$ 4.6}  & \textcolor{blue}{0.1 $\pm$ 0.7}    \\
		AutoReg &  \textcolor{red}{0} & \textcolor{red}{5 $\pm$ 6}   & \textcolor{red}{0} & \textcolor{red}{0.1 $\pm$ 0.7}     \\
		\hline 
	\end{tabular}
	\label{tab:Flods}
\end{table*}

\vspace{-1.0em}
\begin{table}[b]
	\footnotesize
	\centering
	\caption{ Quantitative comparisons in terms of running time and model size on brain MR datasets of size $ 160 \times 192 \times 224$.}  
	\begin{tabular}{|m{1.8cm}<{\centering}| m{1.8cm}<{\centering}| m{1.6cm}<{\centering}|}
		\hline
		Methods &  Runtime (s) & \# (MB)  \\
		\hline
		Elastix & 83 $\pm$ 10    & N/A   \\ 	
		ANTs  & 4614 $\pm$ 1030   & N/A  \\  
		NiftyReg  & 435 $\pm$ 39  & N/A  \\ 
		VM & 0.558 $\pm$ 0.017   & 1.146   \\ 	
		VM-diff  & 0.423 $\pm$ 0.011   & 1.016   \\  
		MultiPropReg & \textcolor{blue}{0.360 $\pm$ 0.010} & \textcolor{red}{0.526} \\ 
		AutoReg  &  \textcolor{red}{0.270 $\pm$ 0.010}  & \textcolor{blue}{0.852}  \\  \hline	
	\end{tabular}
	\label{tab:compare_time}
\end{table}

\begin{table*}[!t]
	\footnotesize
	\centering
	\caption{ \textcolor{update}{Comparison results in terms of Dice scores of different methods on multiple registration tasks.  }}
	\begin{tabular}{|m{2.0cm}<{\centering}| m{2.0cm}<{\centering}| m{2.0cm}<{\centering}| m{2.0cm}<{\centering}| m{2.0cm}<{\centering}|m{2.0cm}<{\centering}|m{2.0cm}<{\centering}|}
		\hline
		Method           & Brain T1-to-T1  & Brain T2-to-T2      & Knee T1-to-T1   & Brain T1-to-T2  & Brain T2-to-T1  & Lung CT-to-CT   \\
		\hline
		Initial      & 0.613 $\pm$ 0.057  & 0.563 $\pm$ 0.018 & 0.340 $\pm$ 0.077   & 0.539 $\pm$ 0.007 &  0.539 $\pm$ 0.007 &  0.863 $\pm$ 0.035\\
		ANTs     & \textcolor{blue}{0.777 $\pm$ 0.030} & 0.625 $\pm$ 0.013 & 0.498 $\pm$ 0.210   & 0.537 $\pm$ 0.006 & 0.538 $\pm$ 0.019 & 0.926 $\pm$ 0.049\\
		NiftyReg   & 0.773 $\pm$ 0.026  & \textcolor{blue}{0.645 $\pm$ 0.006} & 0.340 $\pm$ 0.078  & 0.601 $\pm$ 0.007 & \textcolor{blue}{0.639 $\pm$ 0.011} & \textcolor{red}{0.940 $\pm$ 0.039}\\ 
		VM   & 0.757 $\pm$ 0.035      & 0.638 $\pm$ 0.012  &  0.440 $\pm$ 0.132  & 0.579 $\pm$ 0.013 & 0.579 $\pm$ 0.013  & 0.927 $\pm$ 0.034\\
		VM-diff    & 0.765 $\pm$ 0.023   & 0.412 $\pm$ 0.009  & 0.320 $\pm$ 0.007  & 0.515 $\pm$ 0.010  & 0.334 $\pm$ 0.006  & - \\
		MultiPropReg &  0.775 $\pm$ 0.027  & 0.610 $\pm$ 0.012 & \textcolor{blue}{0.578 $\pm$ 0.136}  & \textcolor{red}{0.625 $\pm$ 0.009}  & \textcolor{red}{0.644 $\pm$ 0.007} & 0.931 $\pm$ 0.014\\
		AutoReg    &  \textcolor{red}{0.778 $\pm$ 0.023}  &  \textcolor{red}{0.646 $\pm$ 0.010} & \textcolor{red}{0.616 $\pm$ 0.150}   & \textcolor{blue}{0.610 $\pm$ 0.011} & 0.622 $\pm$ 0.007  & \textcolor{blue}{0.937 $\pm$ 0.014}\\
		\hline 
	\end{tabular}
	\label{tab:compare_added}
\end{table*}

\subsection{Comparison Experiments}

\subsubsection{Baseline methods}

We compare our approach with five state-of-the-art registration methods, including  two optimization-based tools: Symmetric Normalization (SyN)~\cite{SyN} and NiftyReg~\cite{SunNK14}, three learning-based methods: VoxelMorph~\cite{BalakrishnanZSG19}, diffeomorphic variant~\cite{DalcaBGS19} (referred as VM and VM-diff, respectively) and MultiPropReg~\cite{9551747}. 
We train deep methods with recommended hyper-parameters on the same datasets from scratch.
The parameter settings of the conventional methods are as follows.
For SyN, we use the version implemented in ANTs~\cite{AvantsTSCKG11} and obtain hyper-parameters from~\cite{BalakrishnanZSG19}, which uses a wide parameter sweep across datasets same to ours. We take cross-correlation as the measurement metric and use a step size of 0.25, Gaussian parameters (9, 0.2), at three scales with 201 iterations each. 
As for NiftyReg, we use the Cross Correlation as the similarity measure. We run it with 12 threads through $1500$ iterations.

\begin{figure*}[t]
	\centering 
	\begin{tabular}{@{\extracolsep{0em}}c@{\extracolsep{0.4em}}c@{\extracolsep{0.4em}}c@{\extracolsep{0.4em}}c}
		\includegraphics[width=0.266\textwidth]{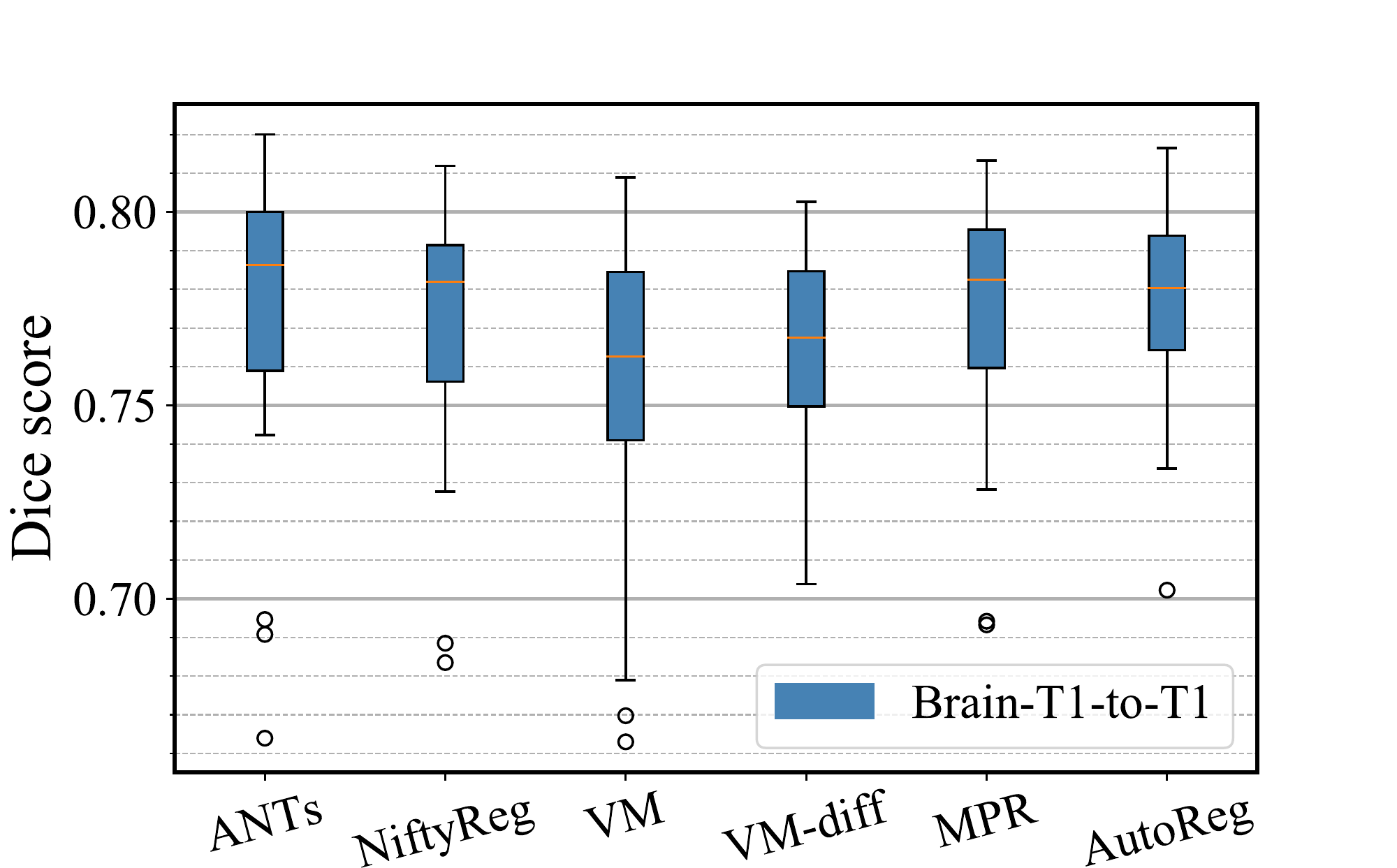}  &
		\includegraphics[width=0.234\textwidth]{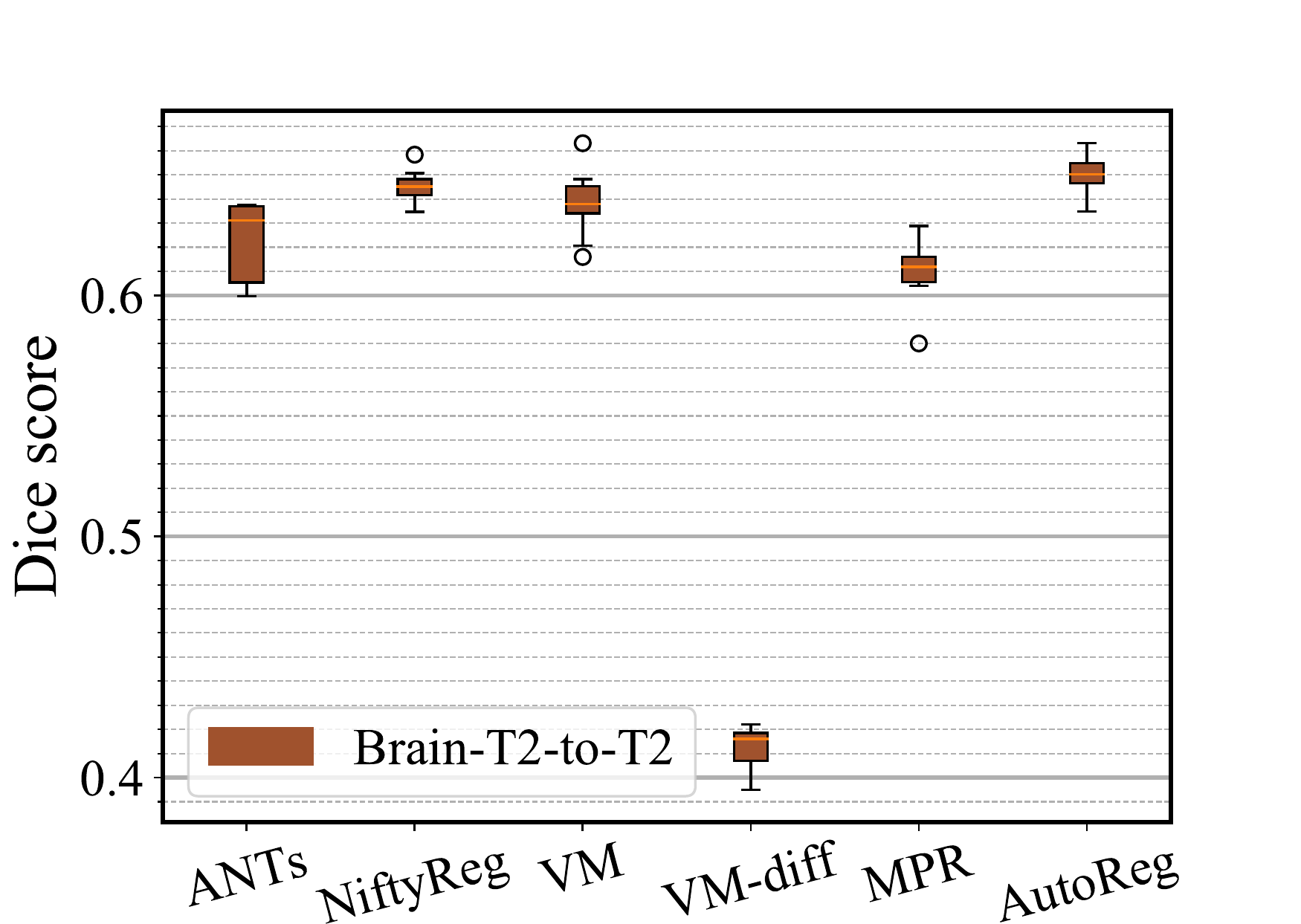}  &
		\includegraphics[width=0.234\textwidth]{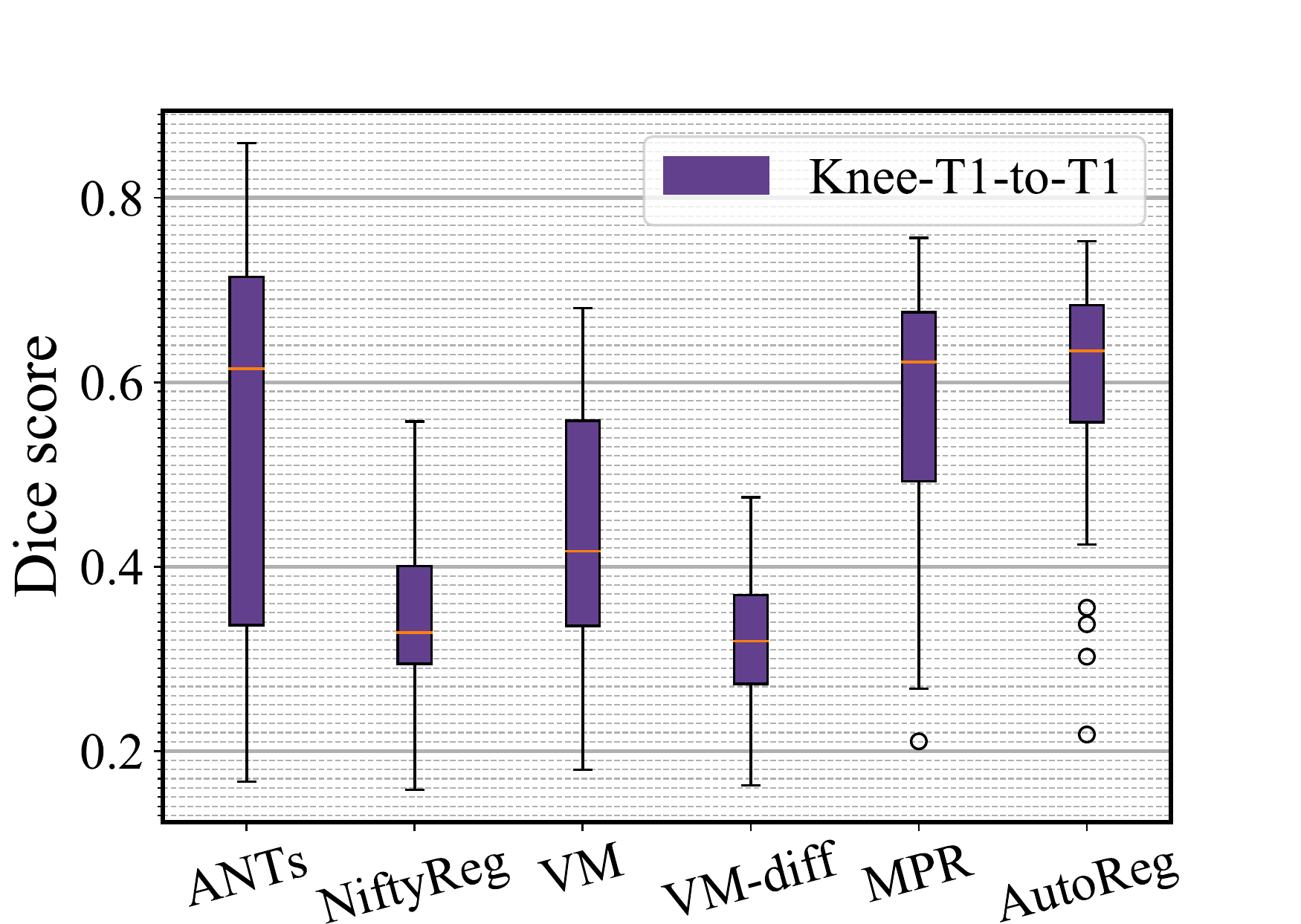}  &
		\includegraphics[width=0.234\textwidth]{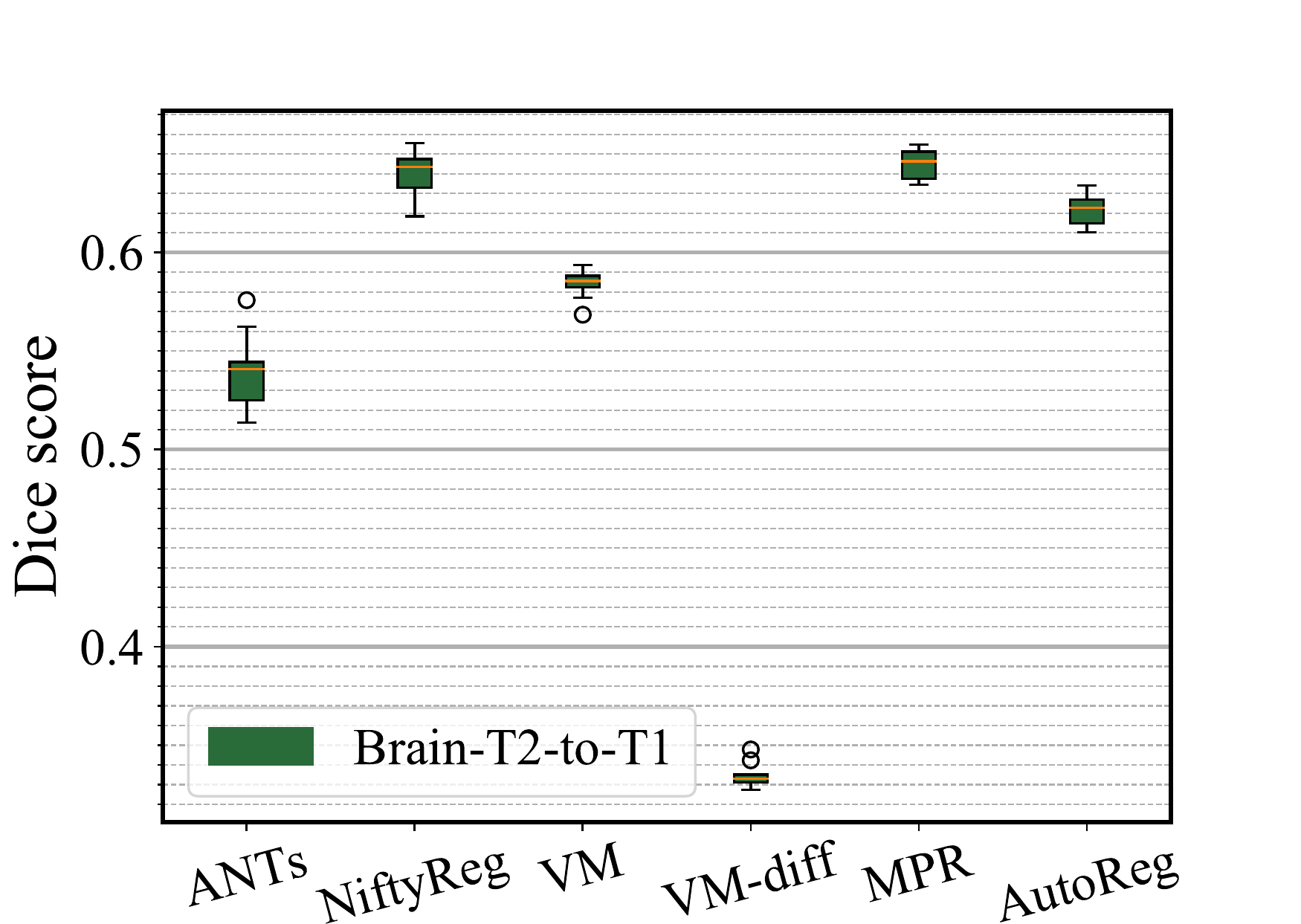}
	\end{tabular}
	\caption{ Registration accuracy compared to different registration methods on diverse registration tasks in terms of Dice score. Each box shows mean accuracy over all anatomical structures for all test-image pairs.} 
	\label{fig:boxplot1}
\end{figure*}

\begin{figure*}[!htp]
	\centering
	\begin{tabular}{@{\extracolsep{0.2em}}c@{\extracolsep{0.2em}}c@{\extracolsep{0.2em}}c@{\extracolsep{0.2em}}c@{\extracolsep{0.2em}}c@{\extracolsep{0.2em}}c@{\extracolsep{0.2em}}}
		
		\includegraphics[width=0.144\textwidth]{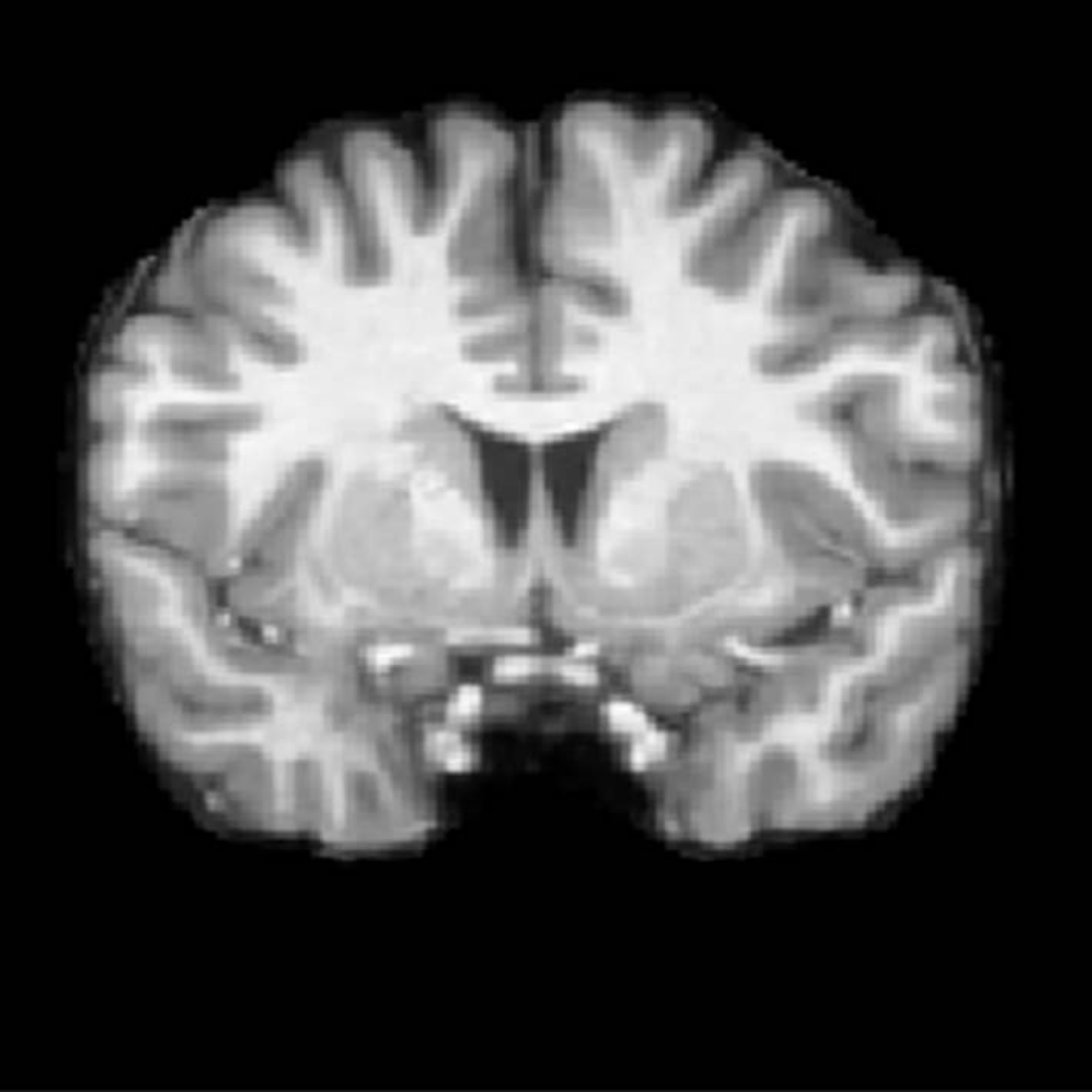}
		&\includegraphics[width=0.144\textwidth]{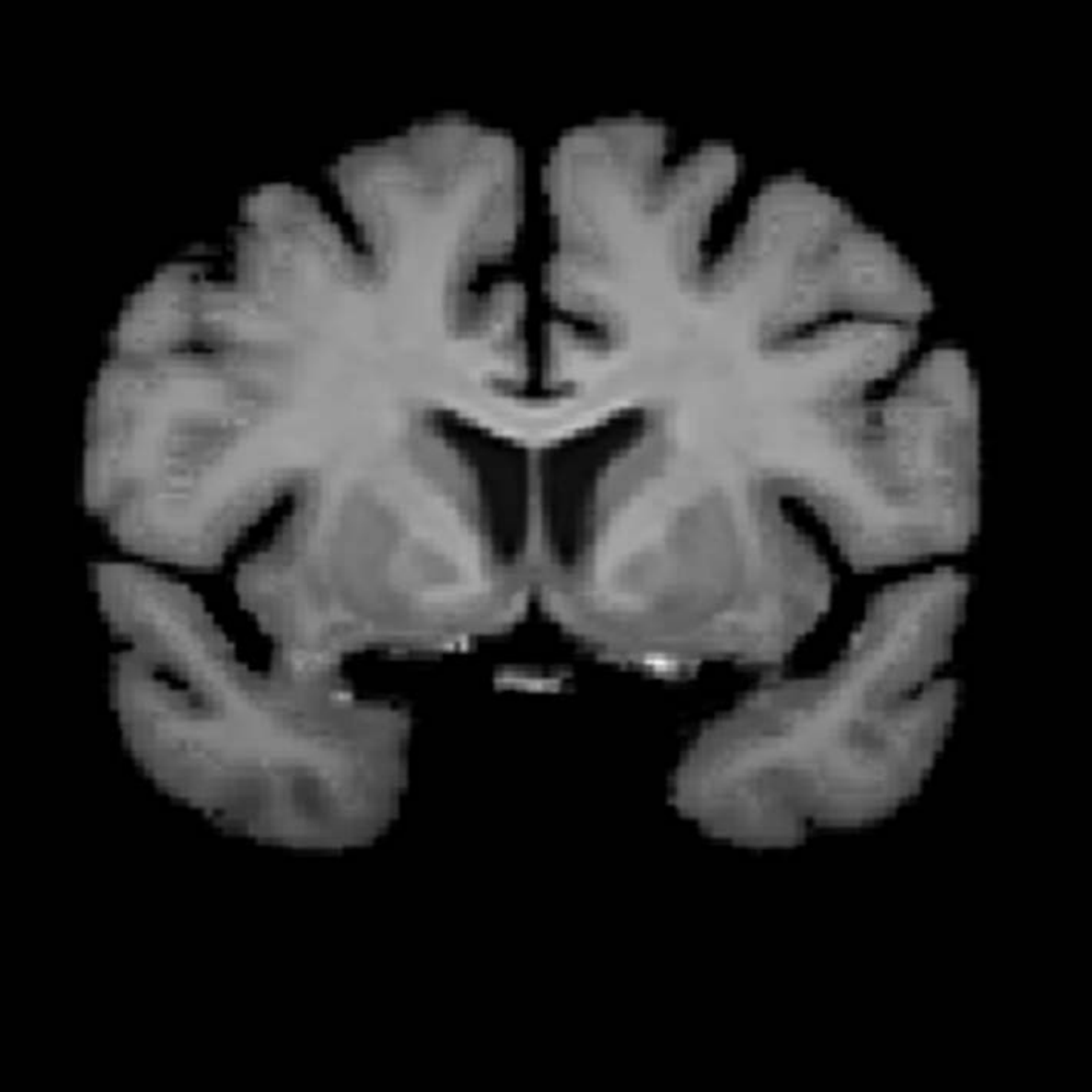}
		&\includegraphics[width=0.144\textwidth]{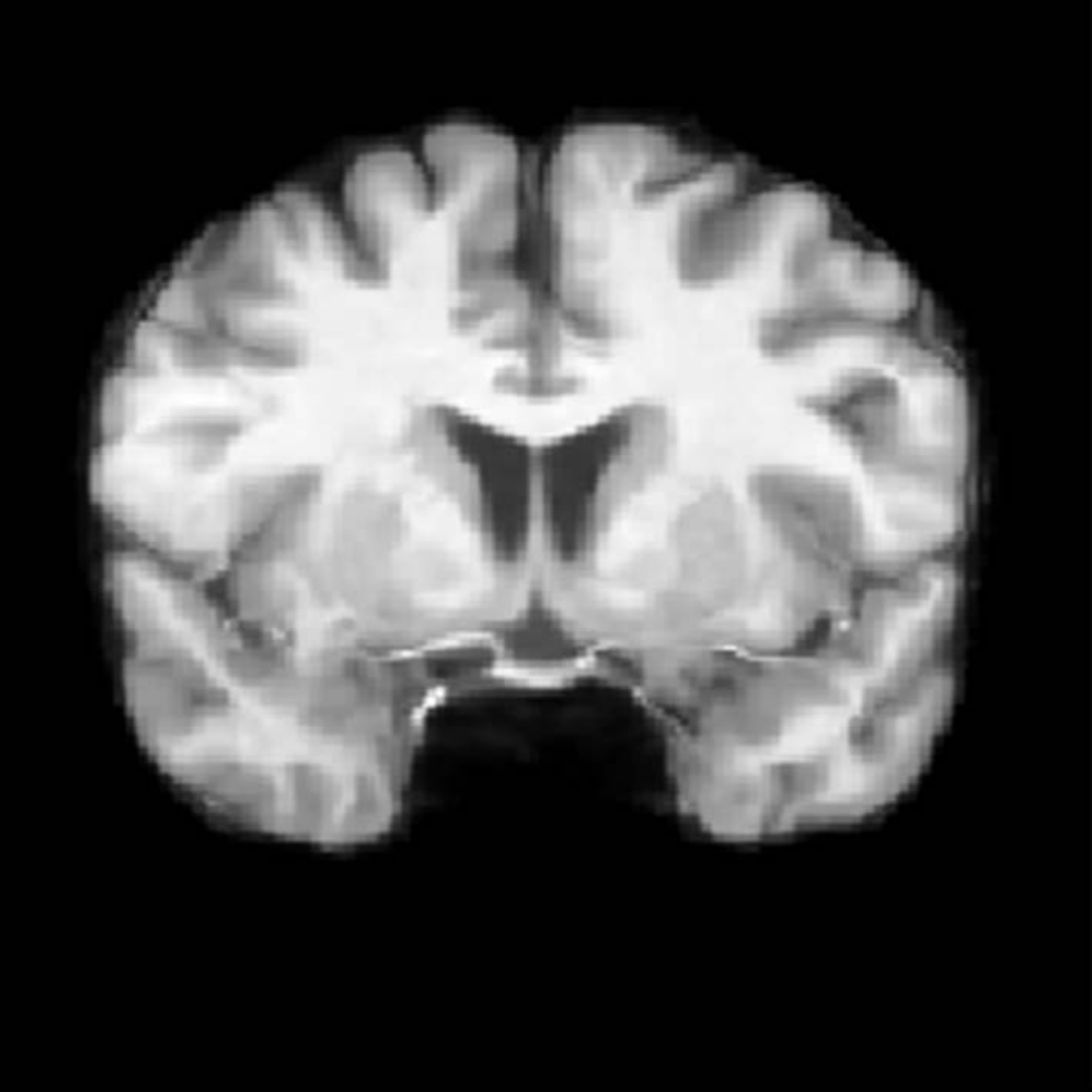}
		&\includegraphics[width=0.144\textwidth]{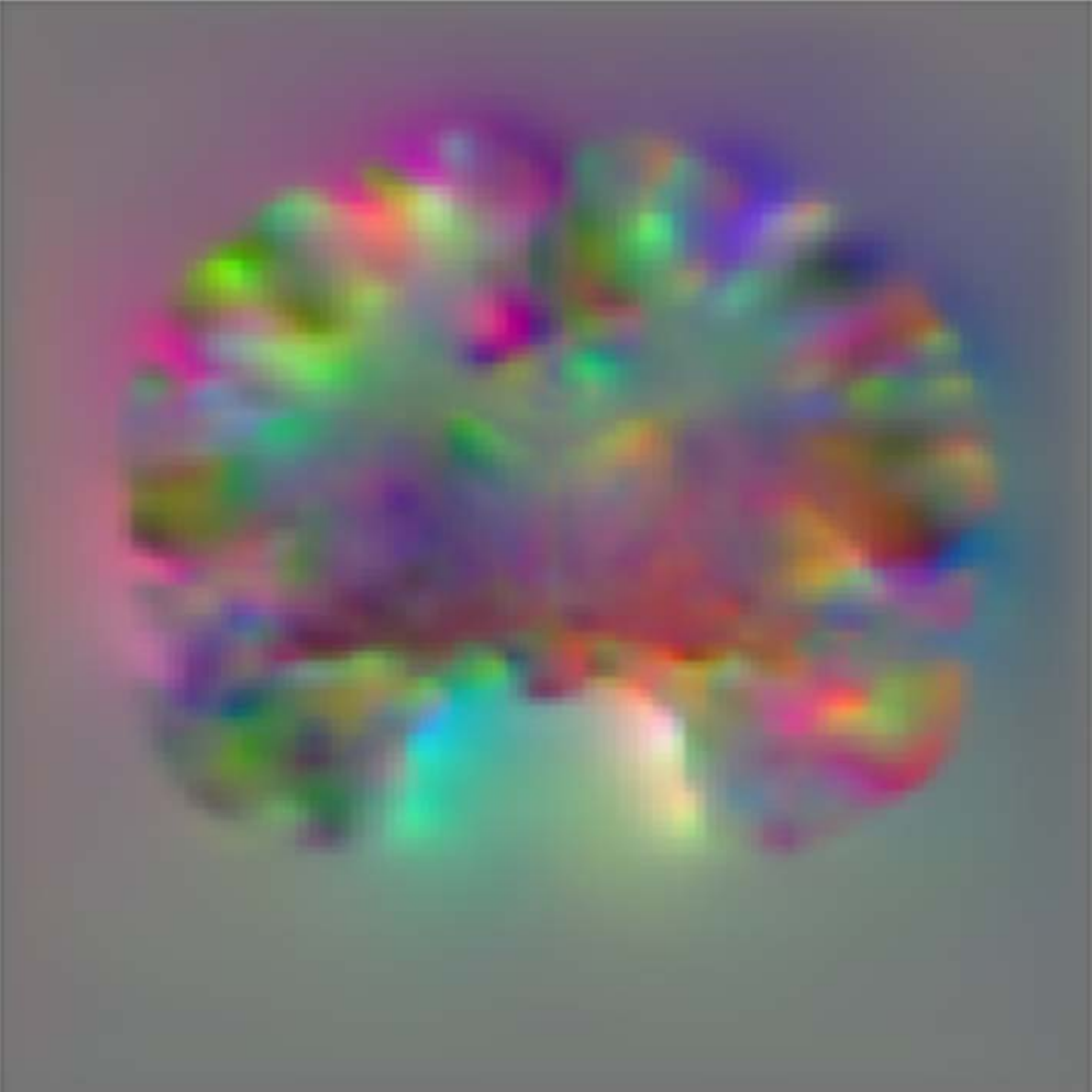}
		&\includegraphics[width=0.144\textwidth]{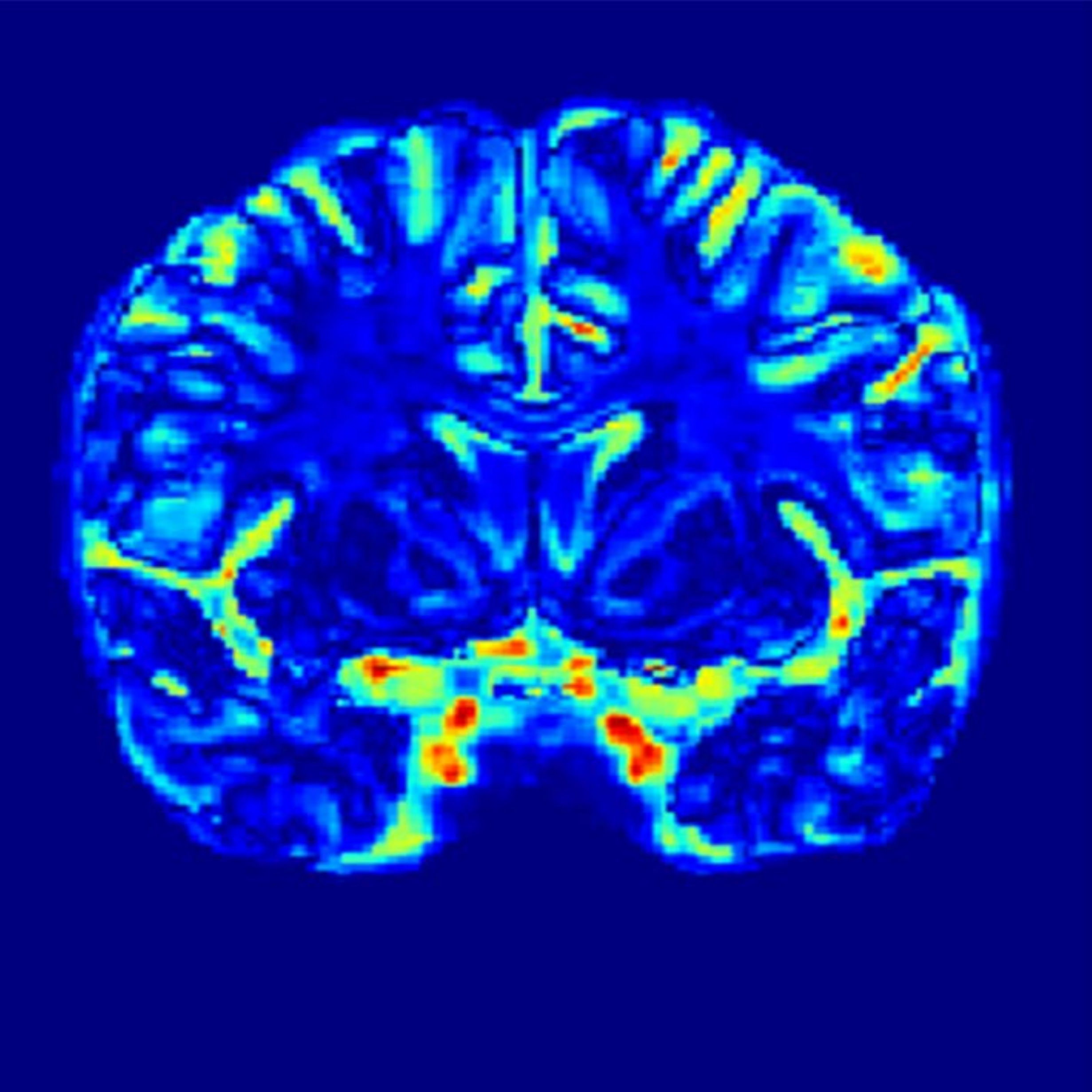} 
		&\includegraphics[width=0.144\textwidth]{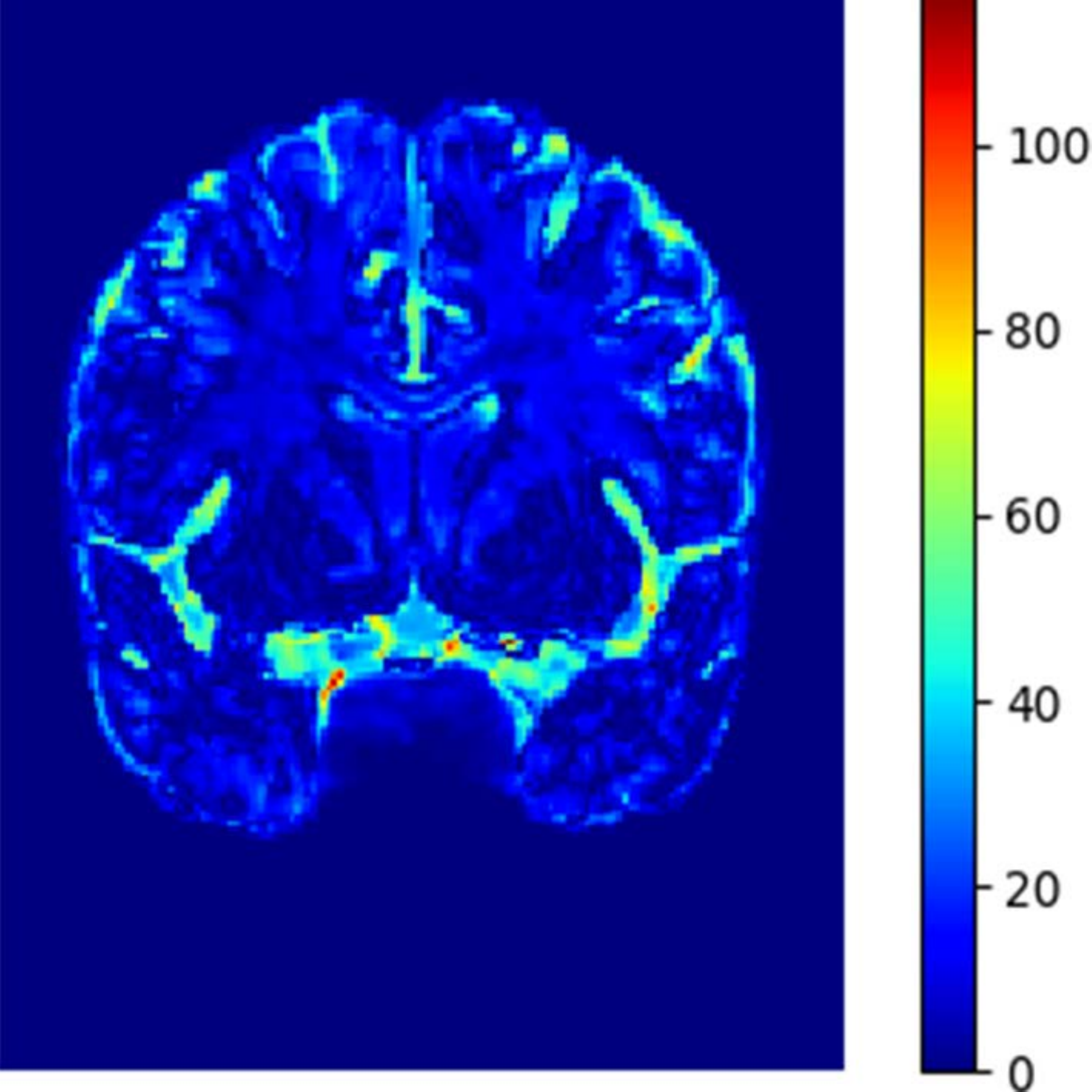} \\
		
		\includegraphics[width=0.144\textwidth]{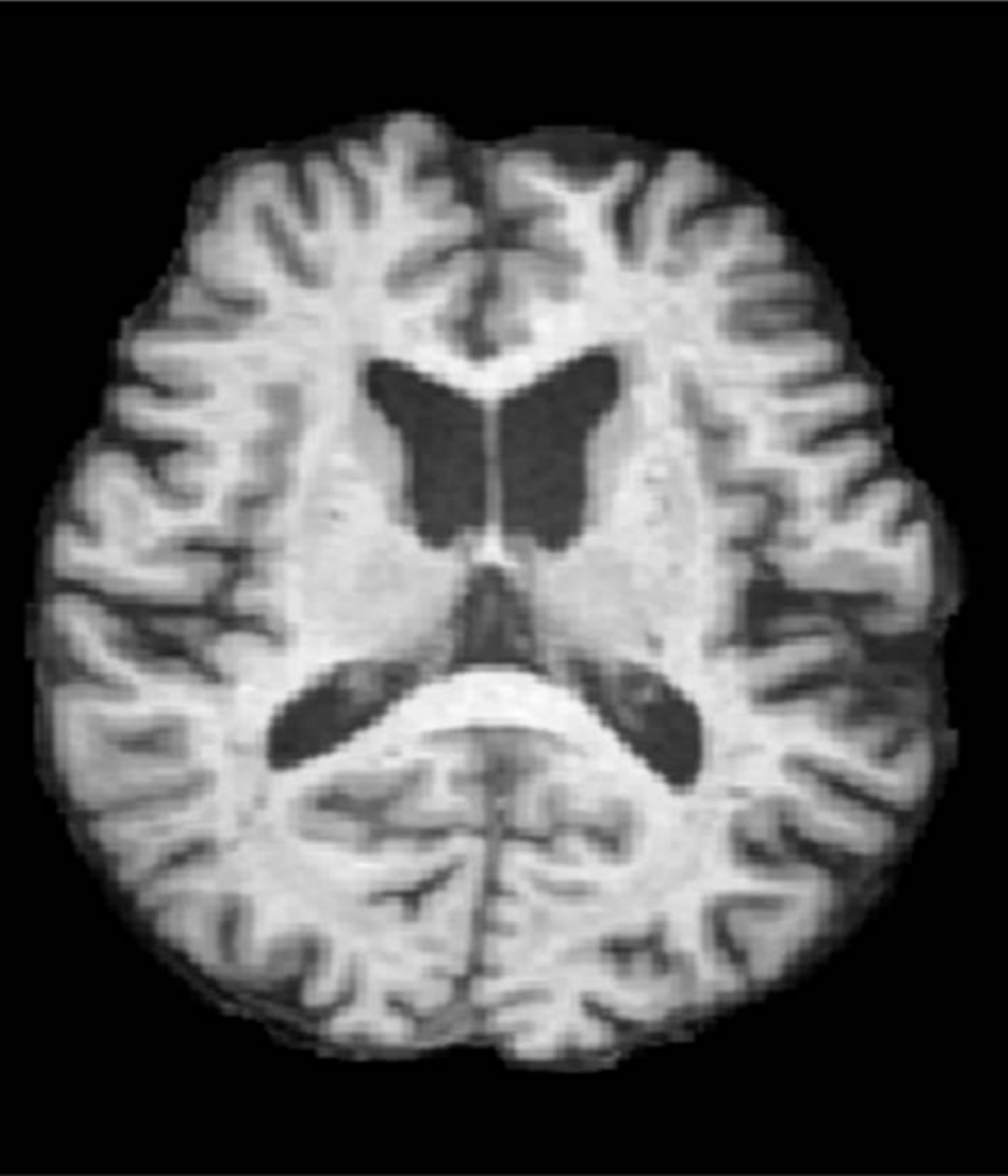}
		&\includegraphics[width=0.144\textwidth]{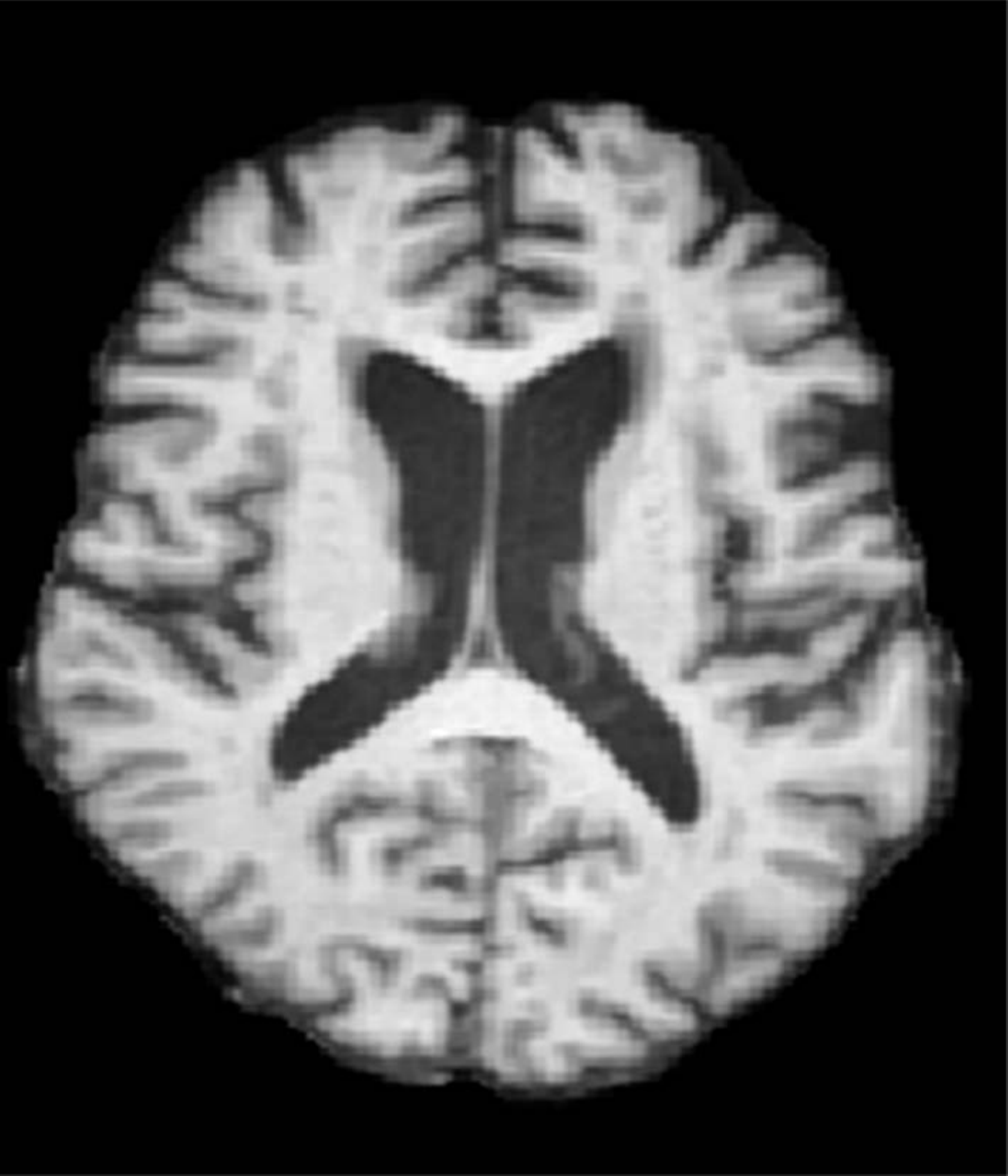}
		&\includegraphics[width=0.144\textwidth]{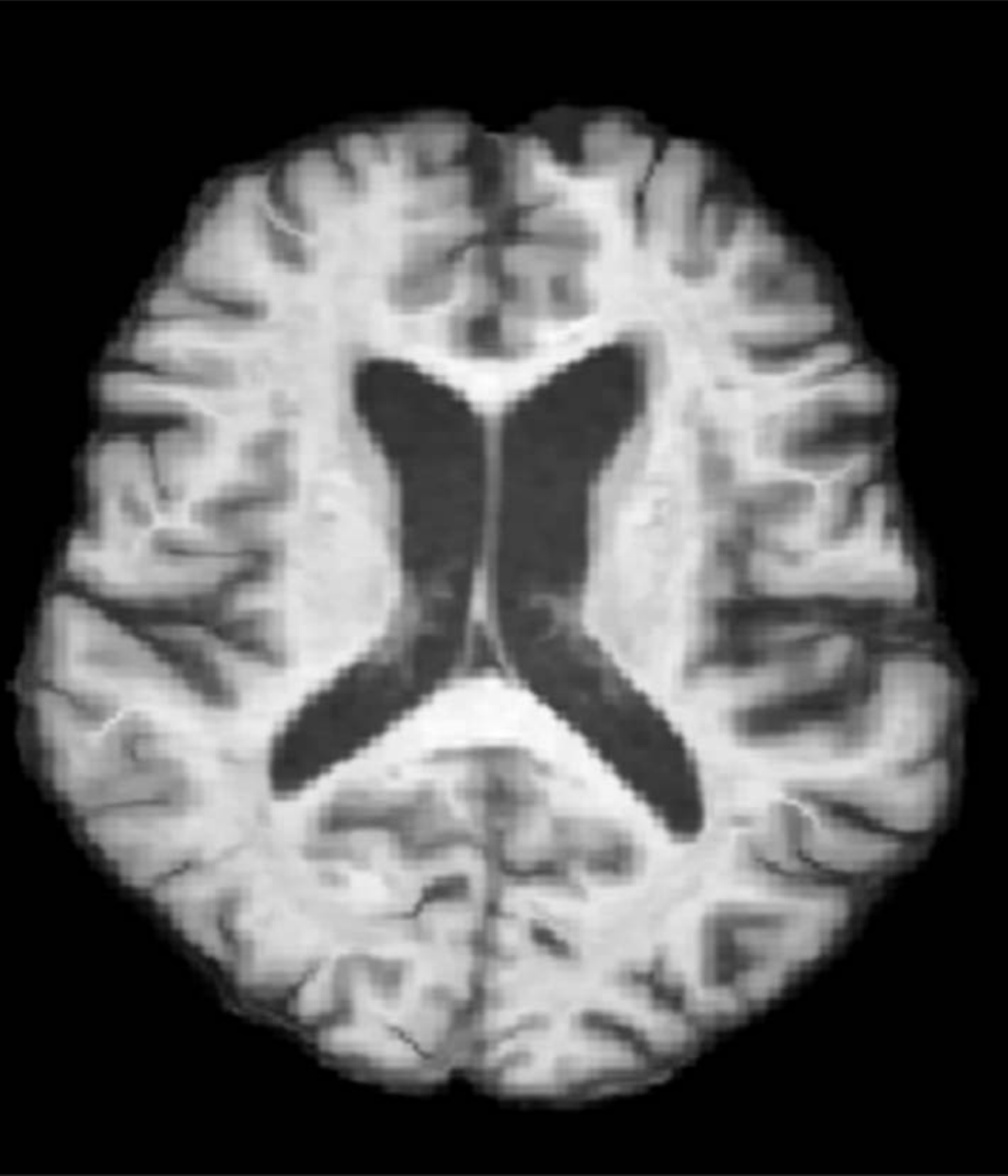}
		&\includegraphics[width=0.144\textwidth]{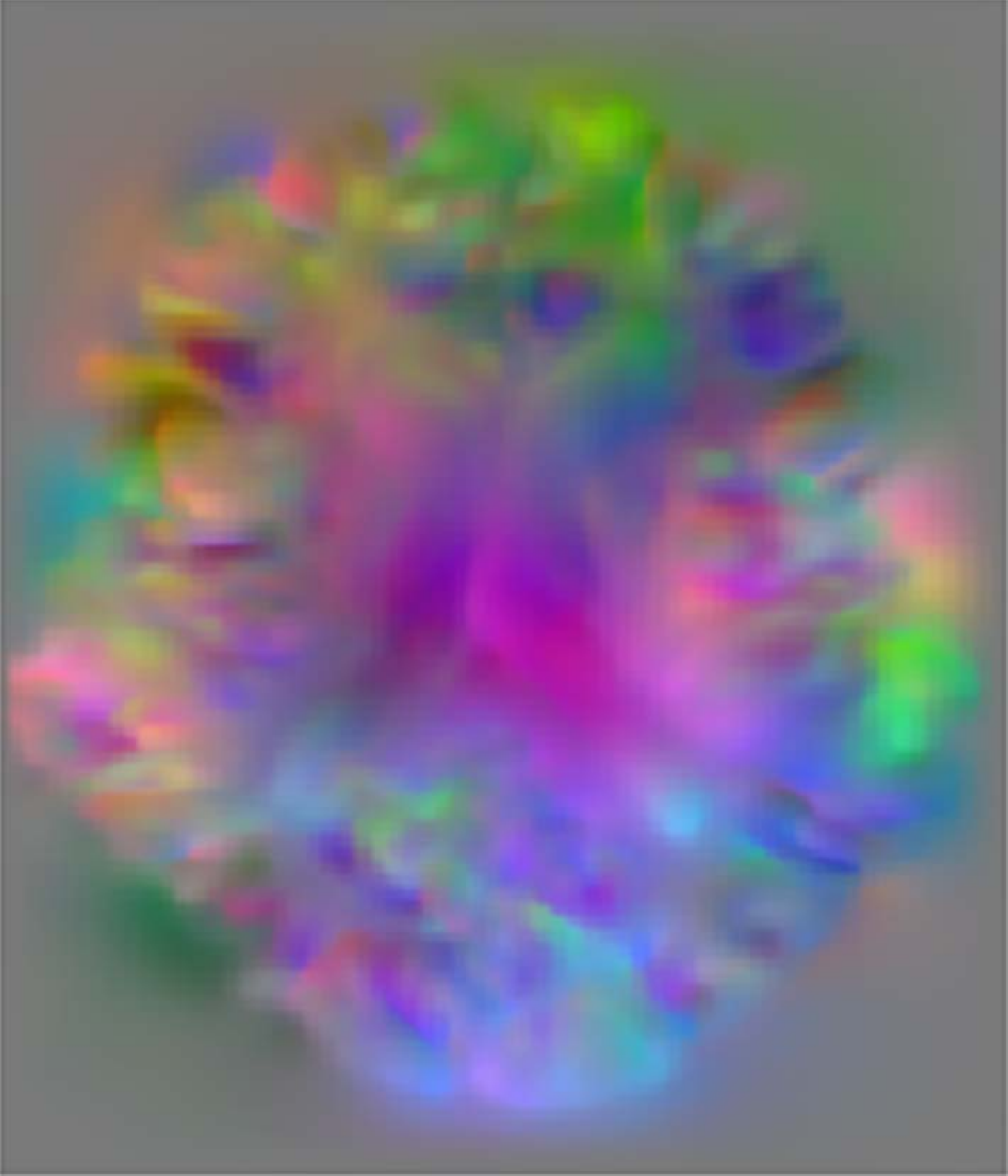}	
		&\includegraphics[width=0.144\textwidth]{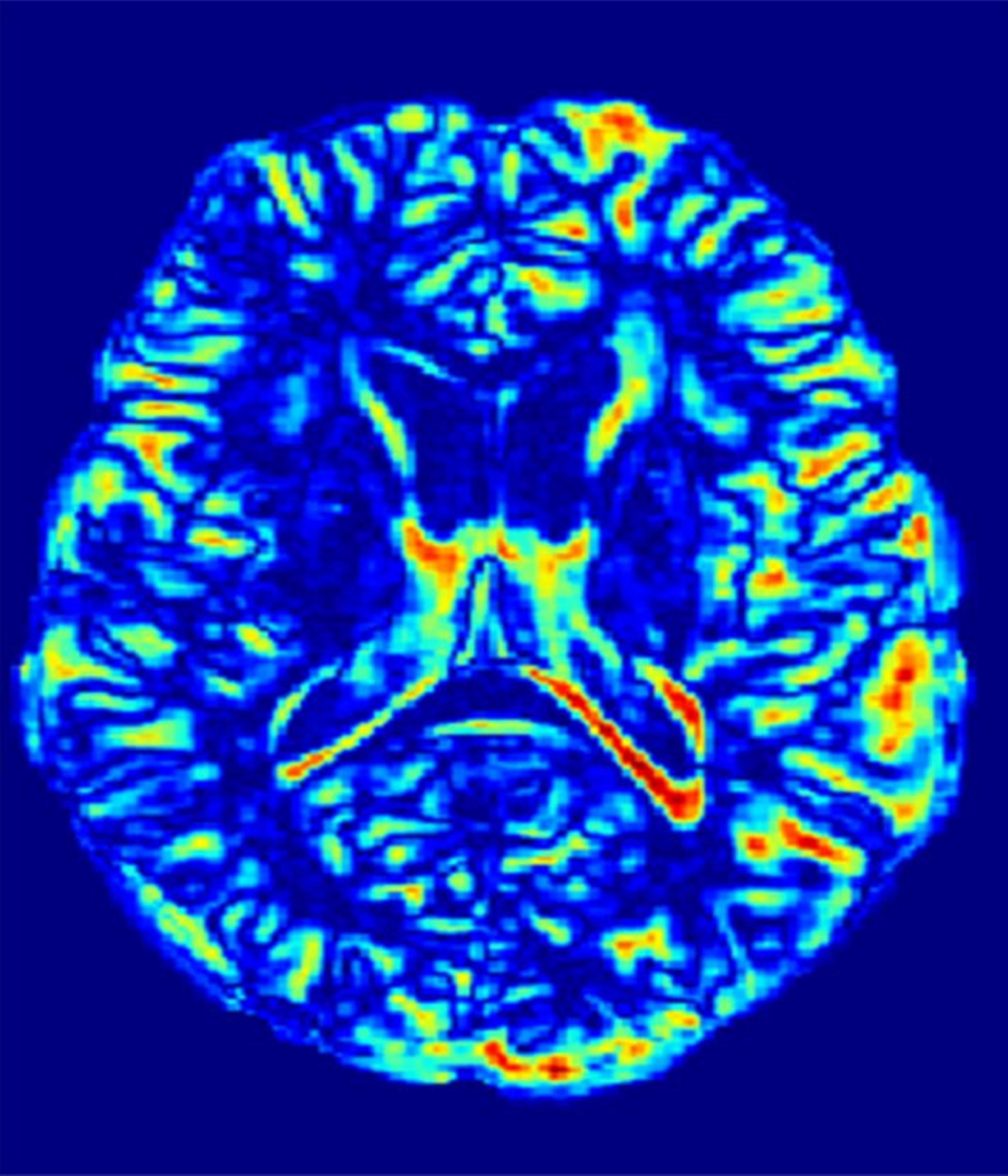}
		&\includegraphics[width=0.144\textwidth]{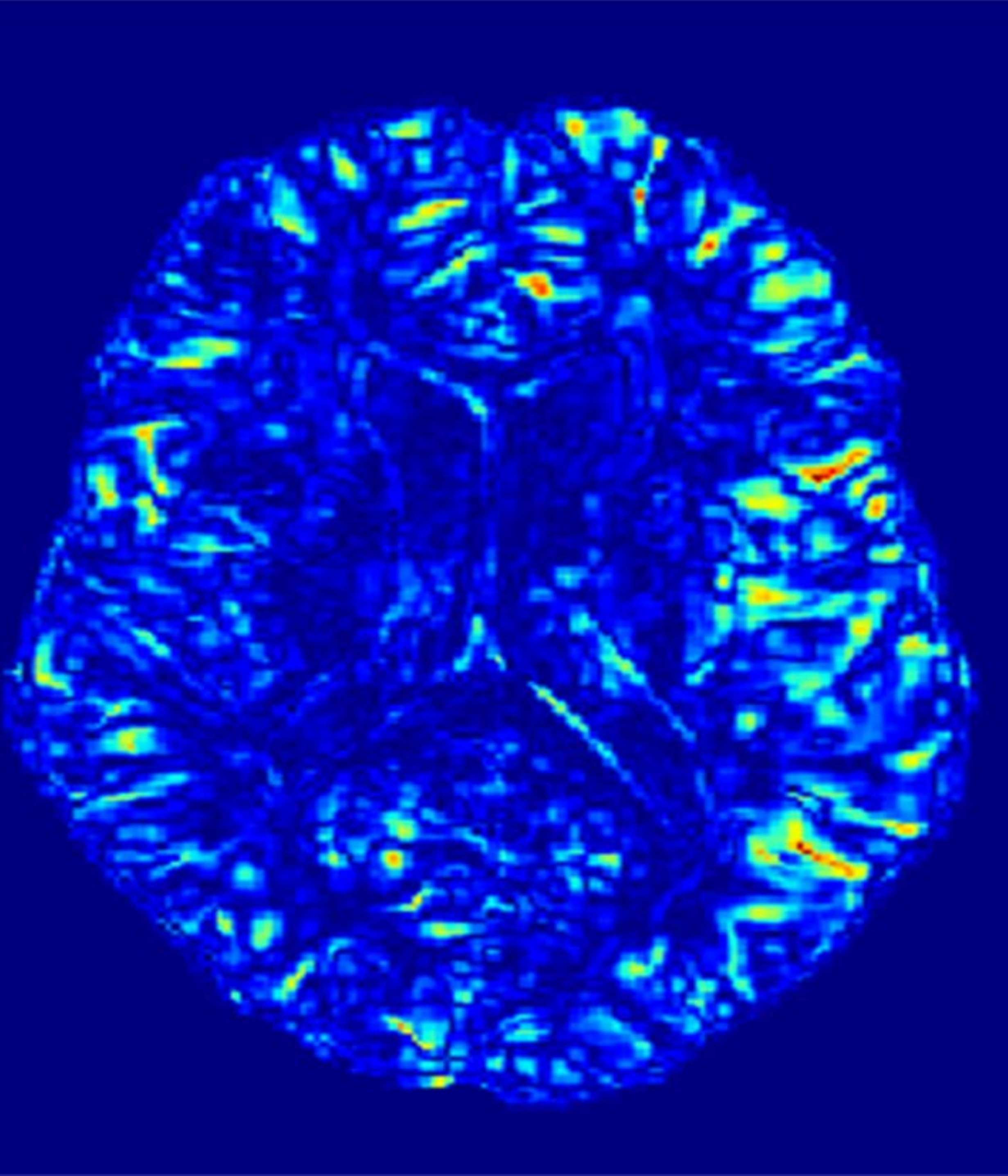} \\
		
		\includegraphics[width=0.144\textwidth]{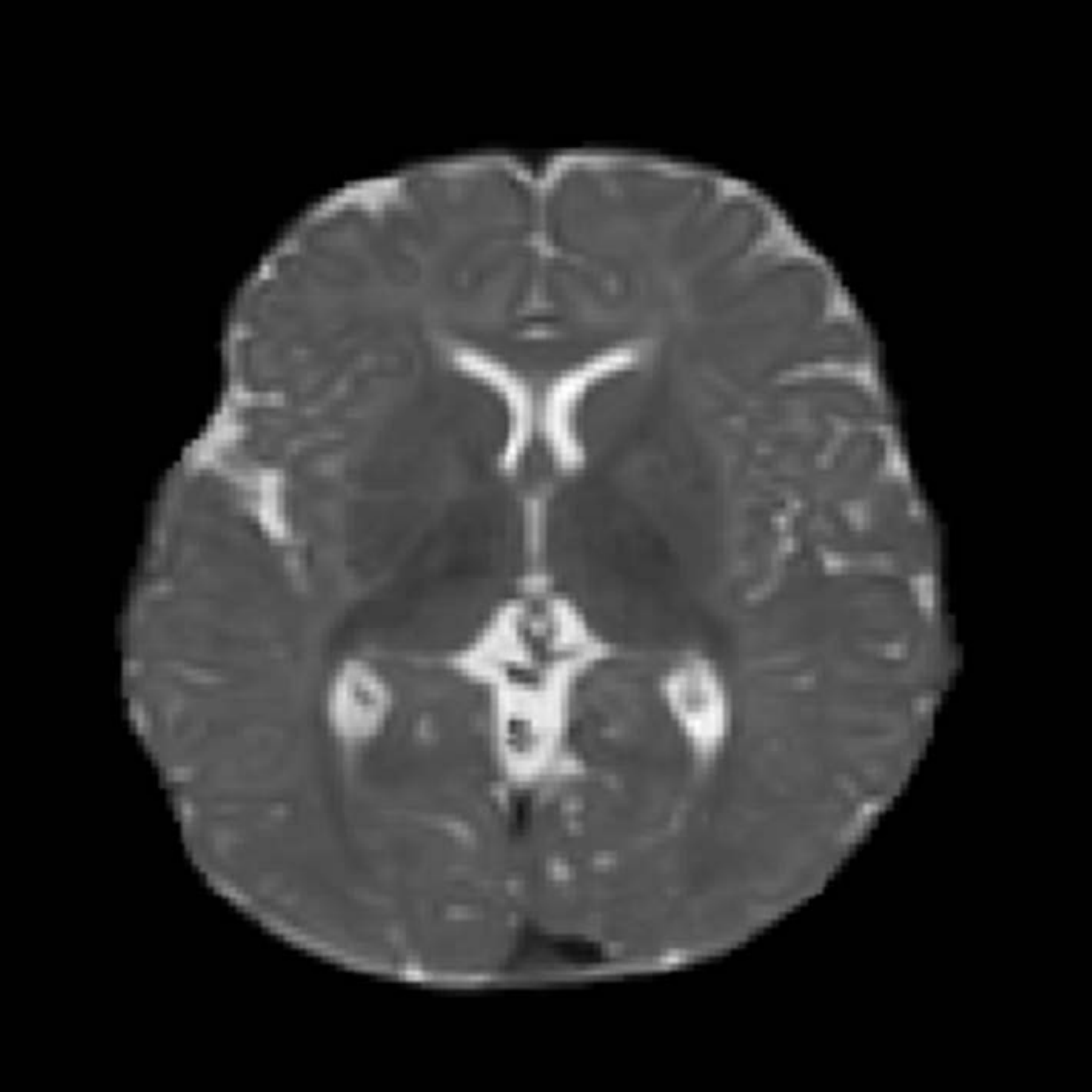}
		&\includegraphics[width=0.144\textwidth]{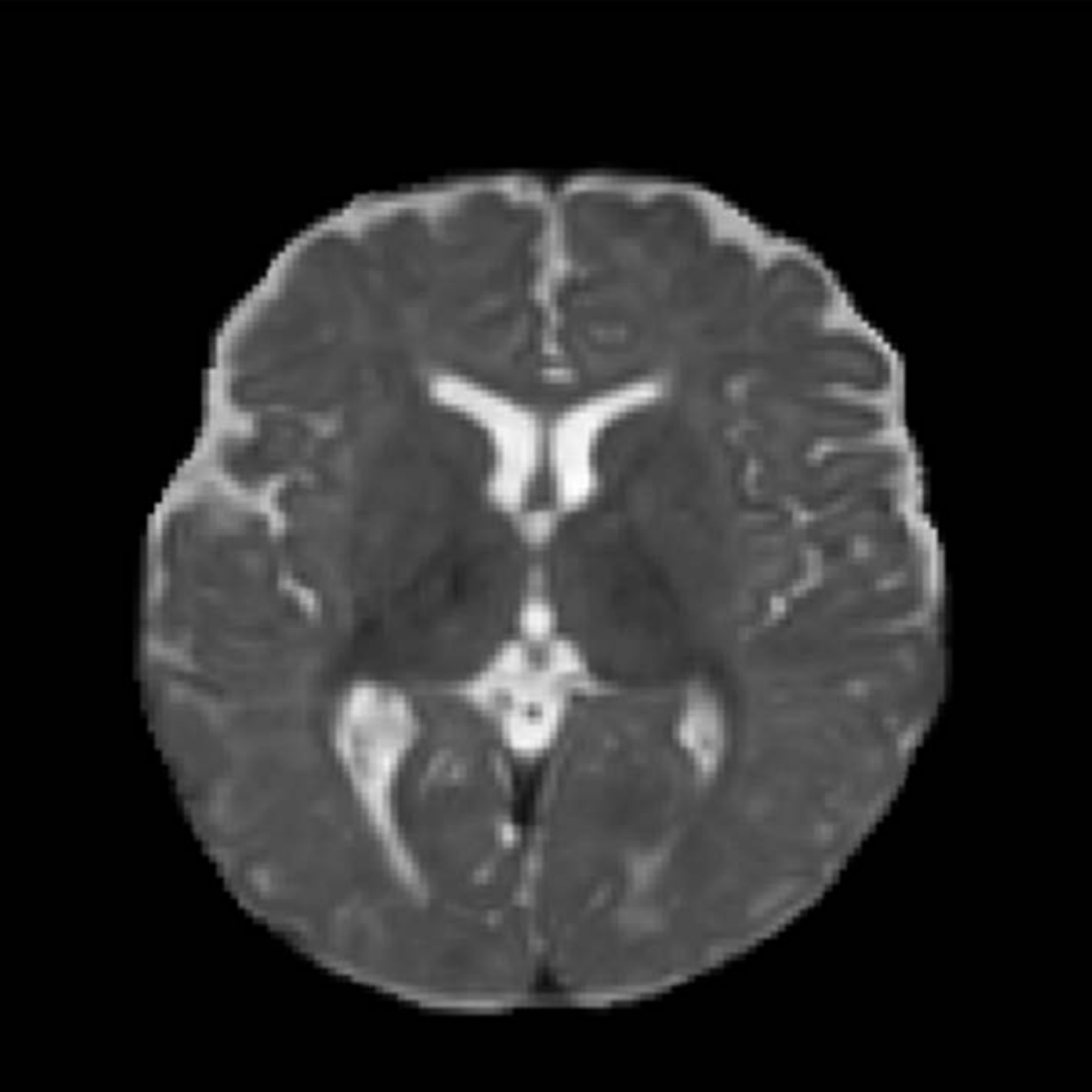}
		&\includegraphics[width=0.144\textwidth]{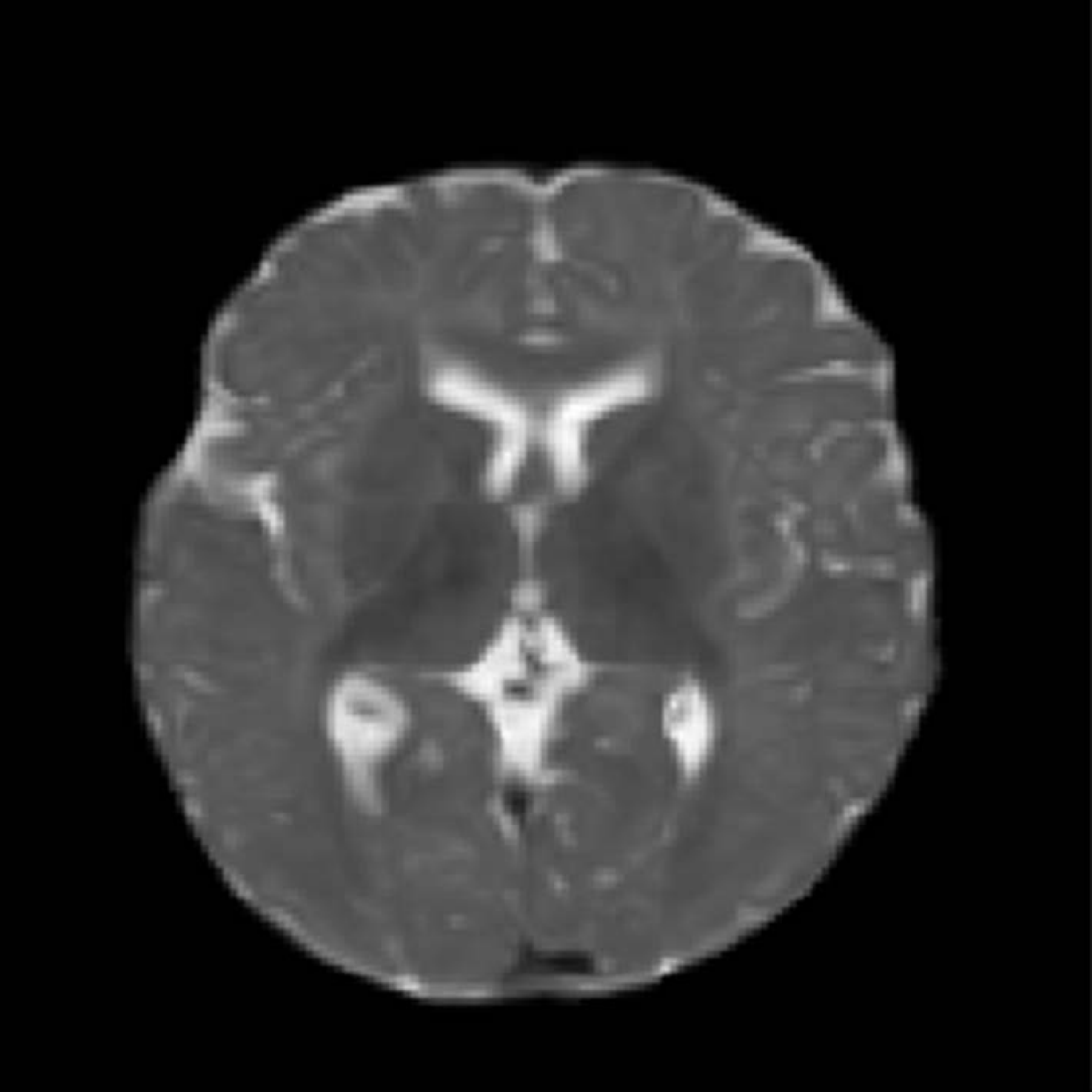}
		&\includegraphics[width=0.144\textwidth]{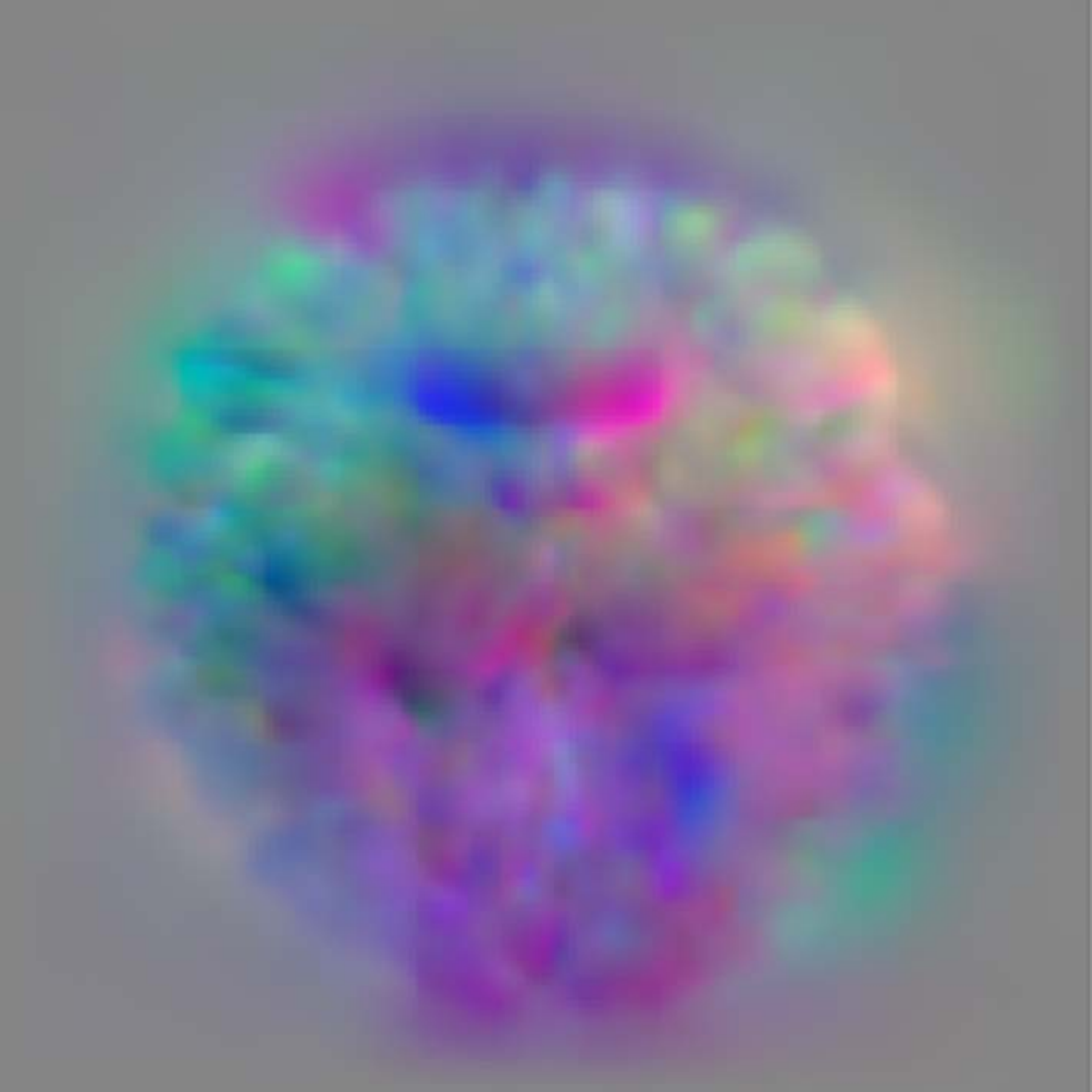}	
		&\includegraphics[width=0.144\textwidth]{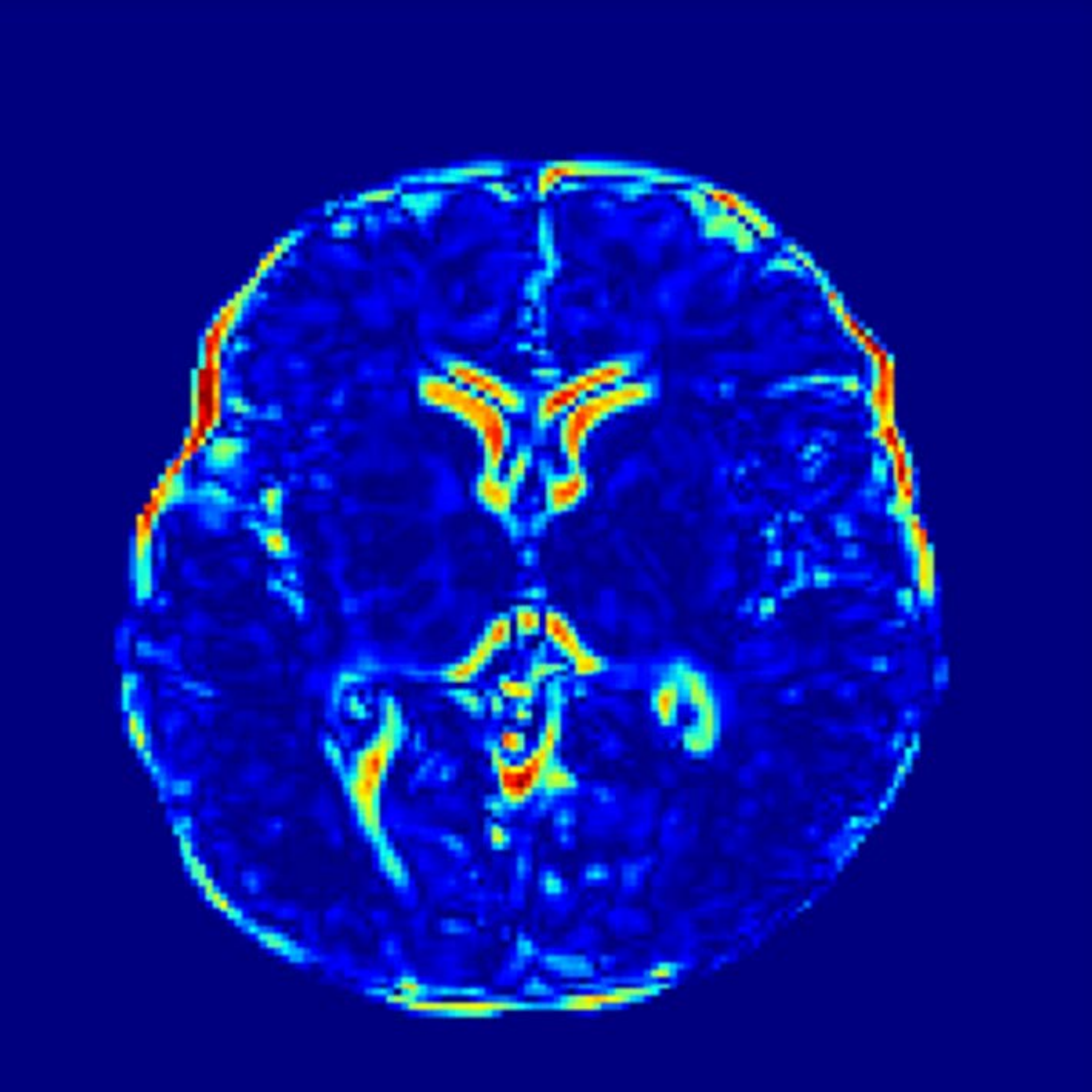}
		&\includegraphics[width=0.144\textwidth]{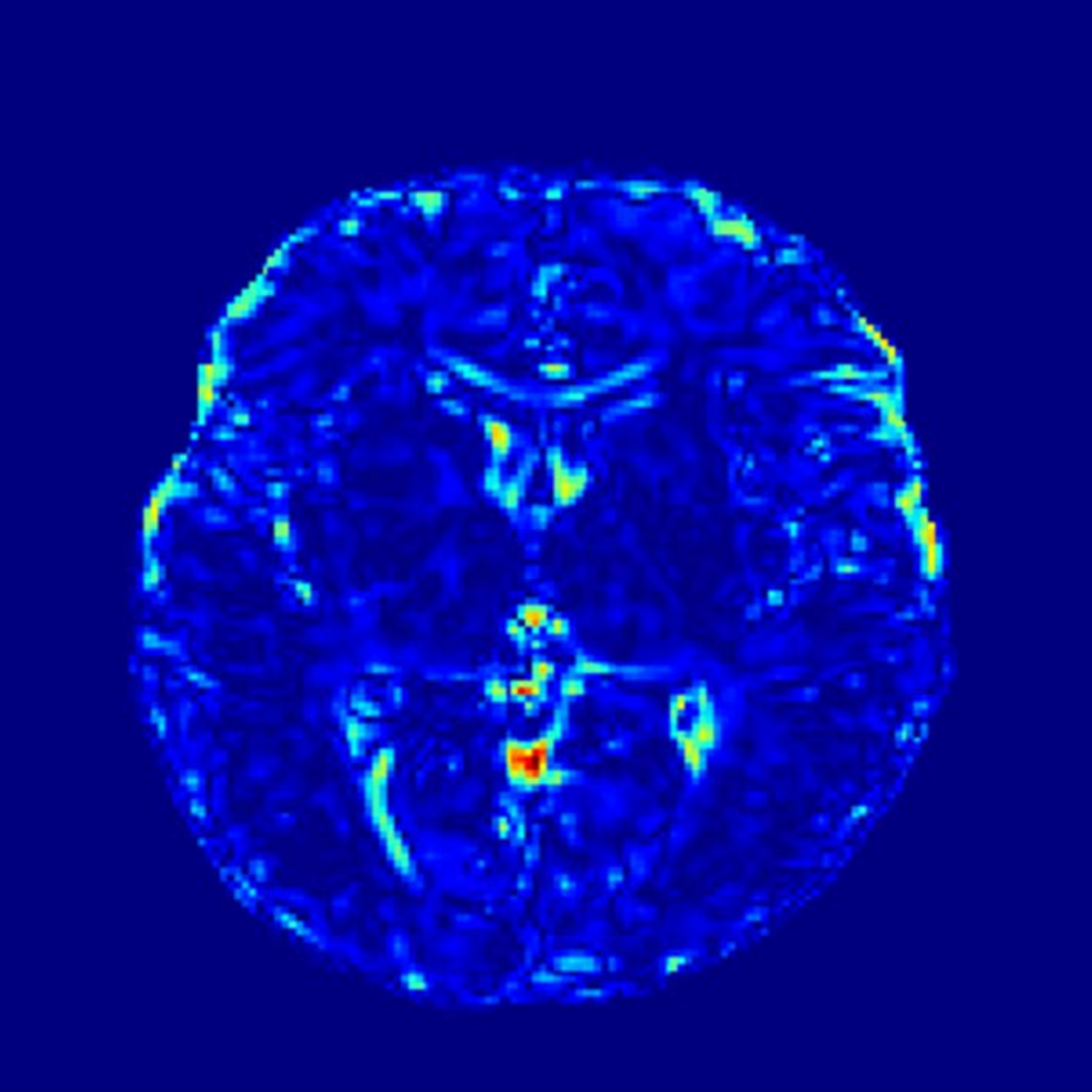} \\
			
		Source  &  Target  &  Warped  & Flow 	& Intensity Error B &  Intensity Error A \\
		
	\end{tabular}
	\caption{ Sample result of registering different images. Each row refers to an example registration case.  Example 2D slices of Intensity difference Before registration and Intensity difference After registration. The registration field is visualized by RGB images with each channel representing dimension. 
	}
	\label{fig:examples} 
\end{figure*}

\begin{figure*}[!htp]
	\centering
	\begin{tabular}{@{\extracolsep{0.2em}}c@{\extracolsep{0.2em}}c@{\extracolsep{0.2em}}c@{\extracolsep{0.2em}}c@{\extracolsep{0.2em}}c@{\extracolsep{0.2em}}c@{\extracolsep{0.2em}}}

		\includegraphics[width=0.144\textwidth]{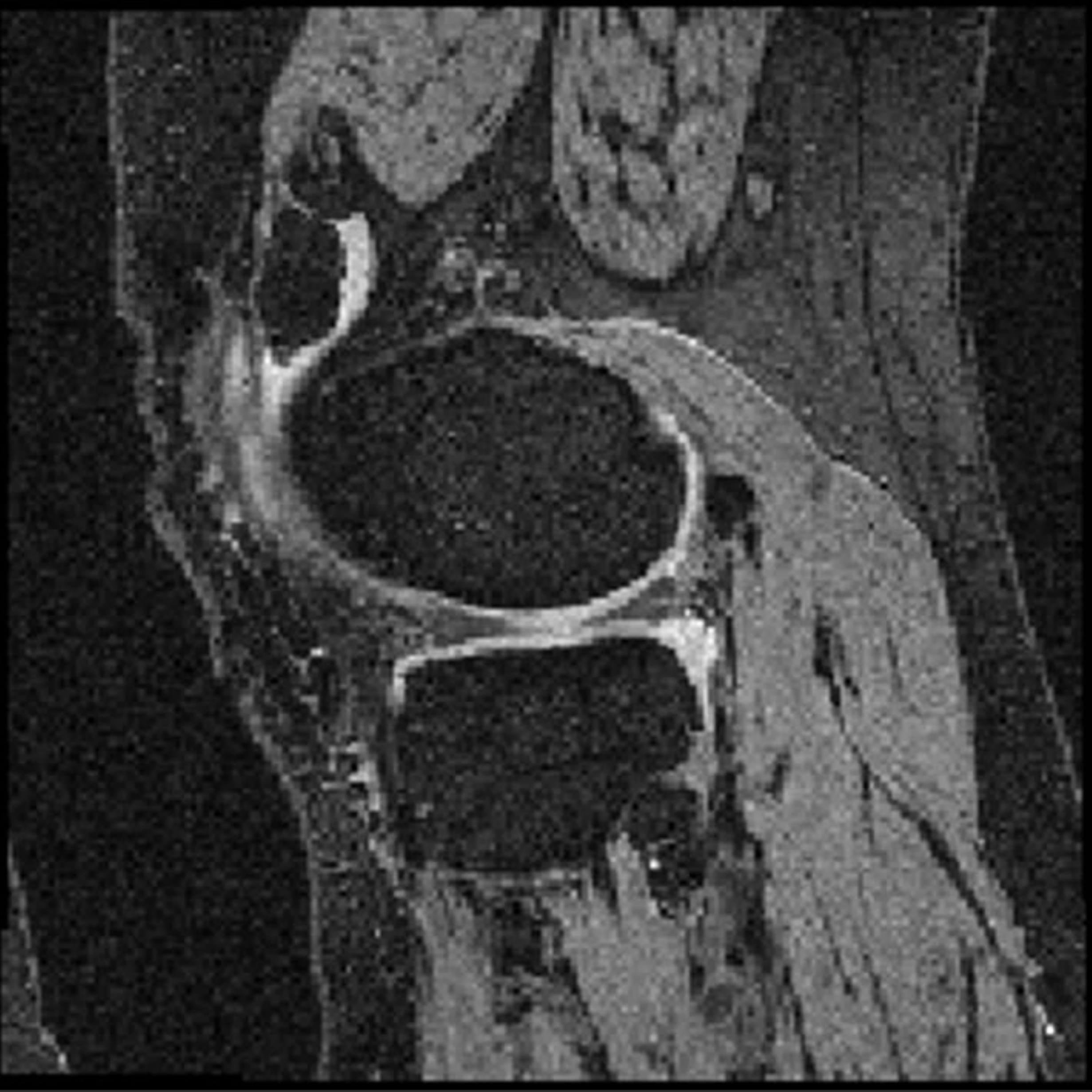}
		&\includegraphics[width=0.144\textwidth]{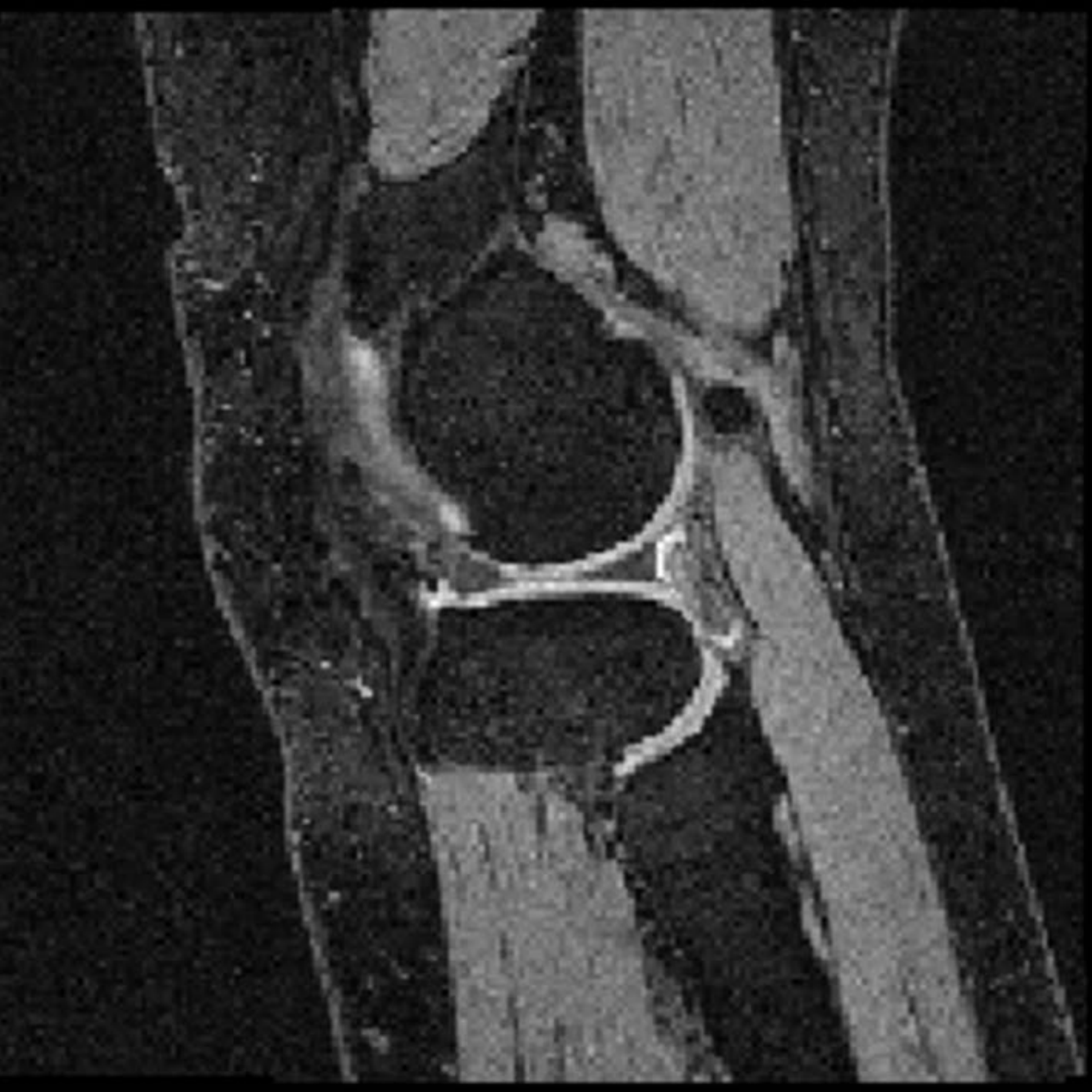}
		&\includegraphics[width=0.144\textwidth]{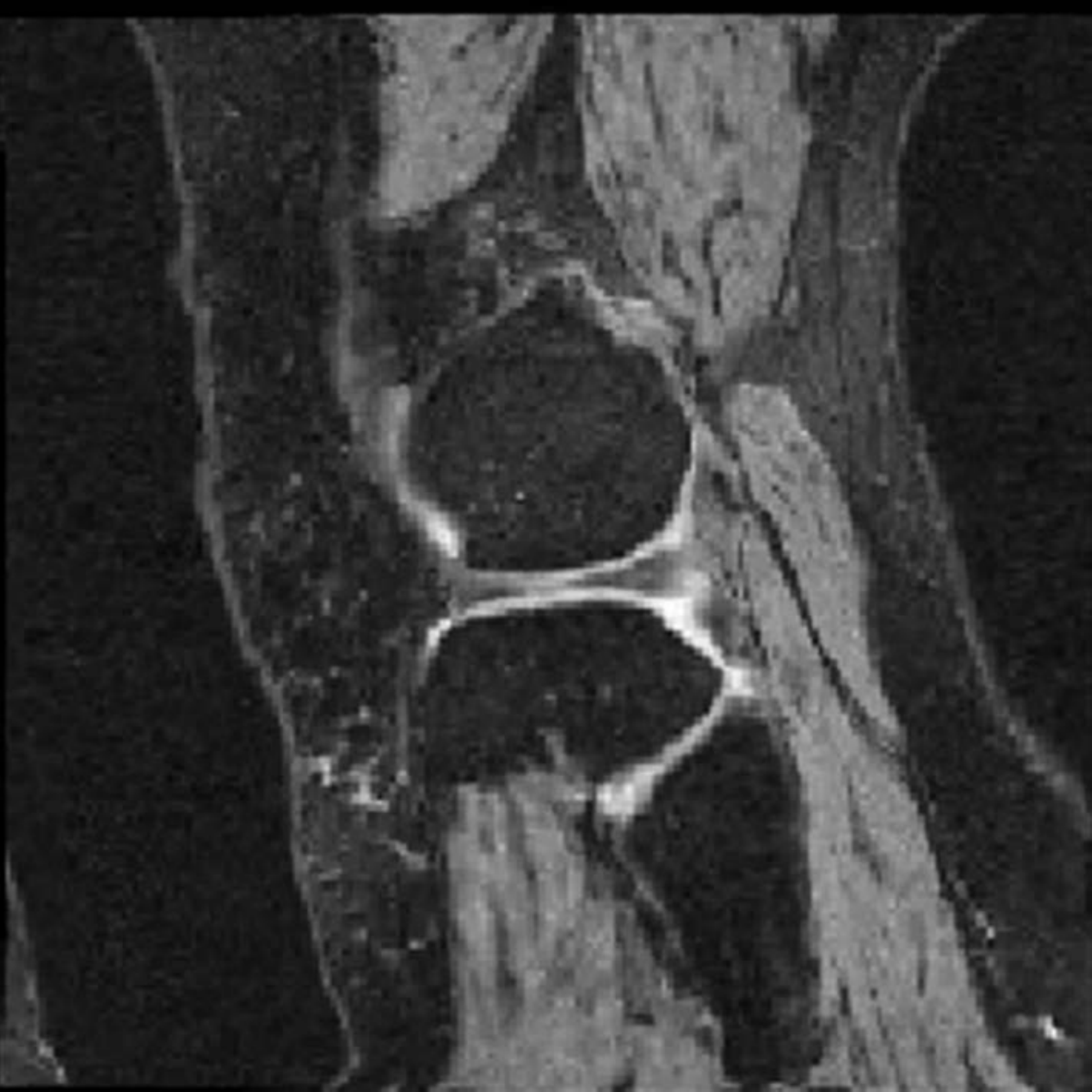}
		&\includegraphics[width=0.144\textwidth]{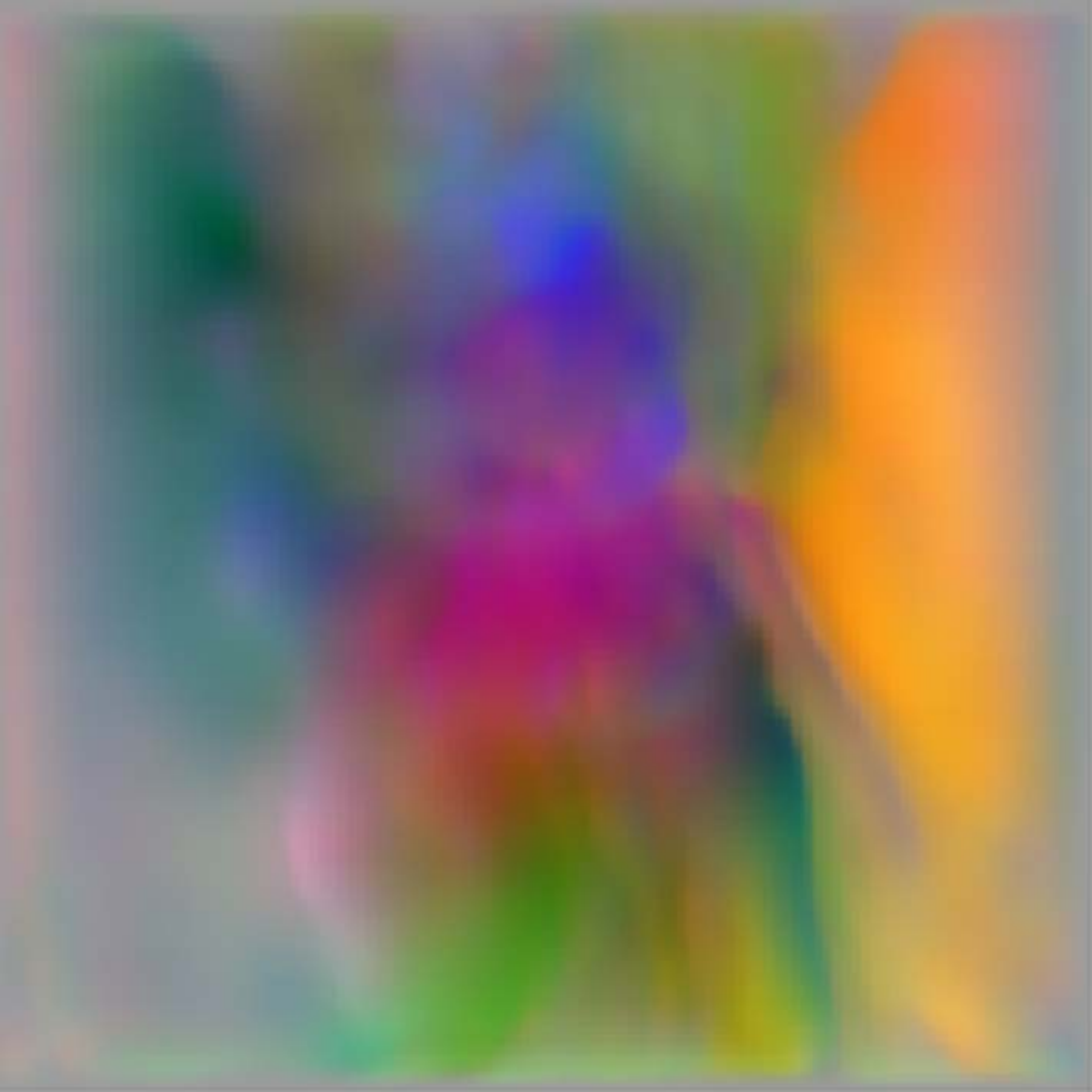}	
		&\includegraphics[width=0.144\textwidth]{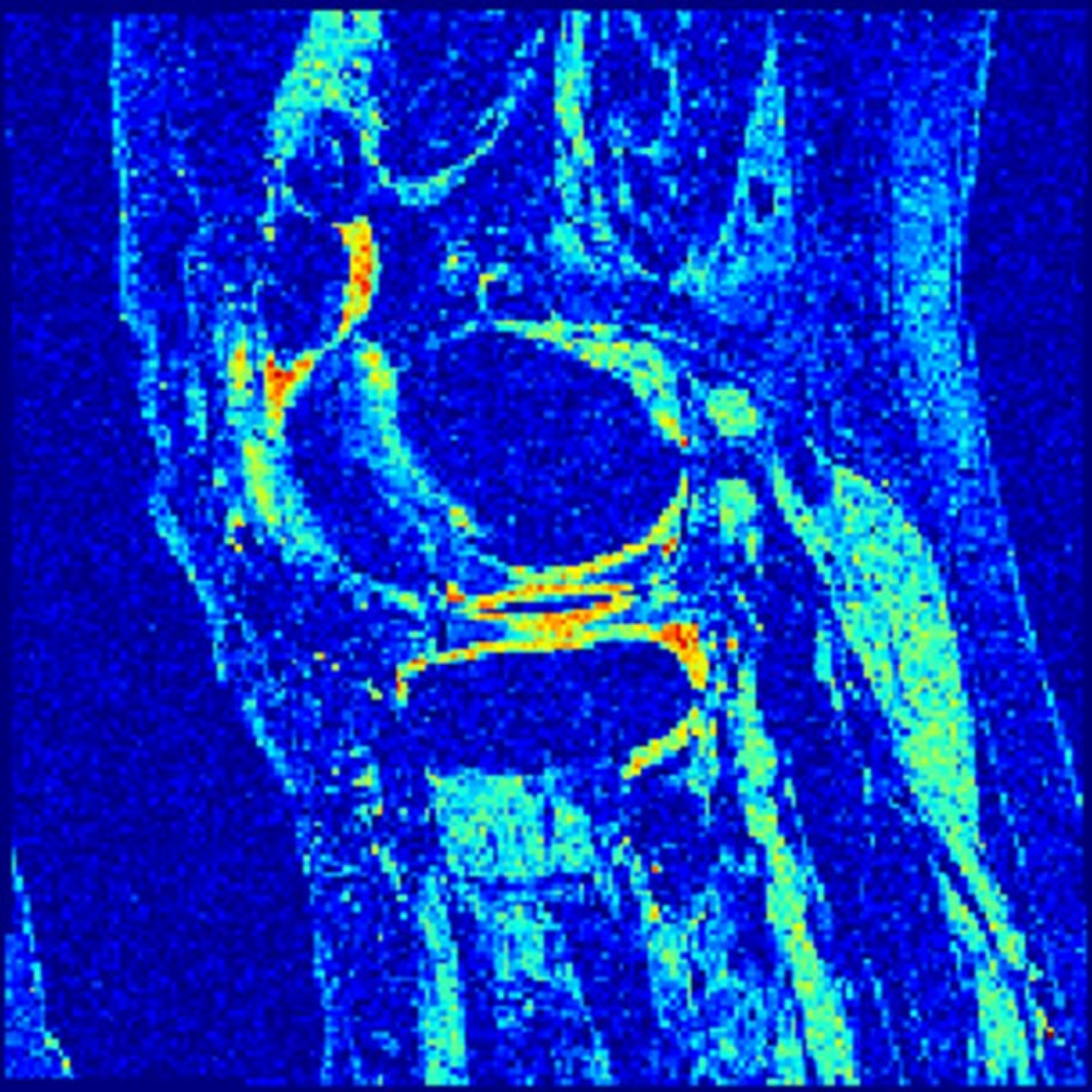}
		&\includegraphics[width=0.144\textwidth]{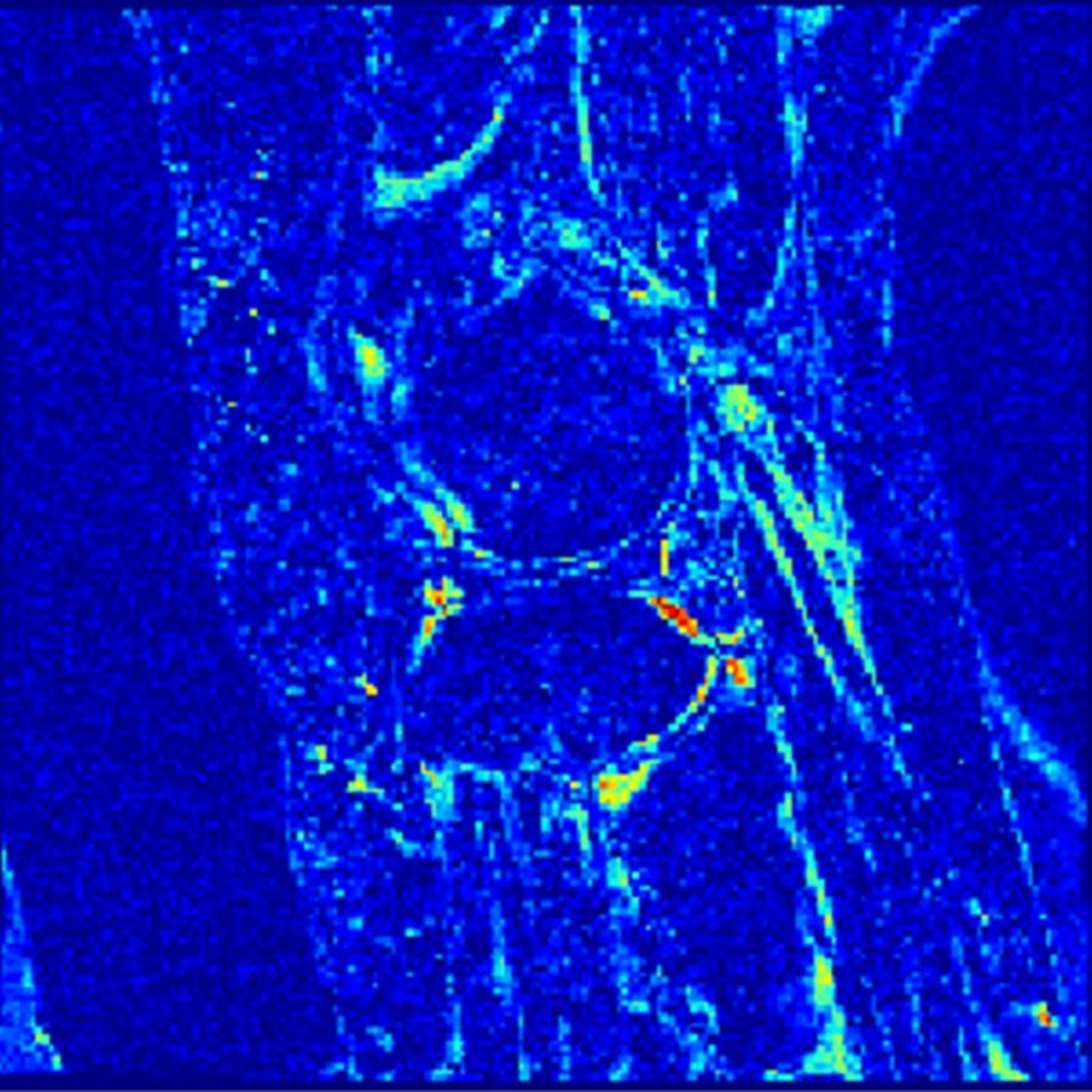} \\
		
		\includegraphics[width=0.144\textwidth]{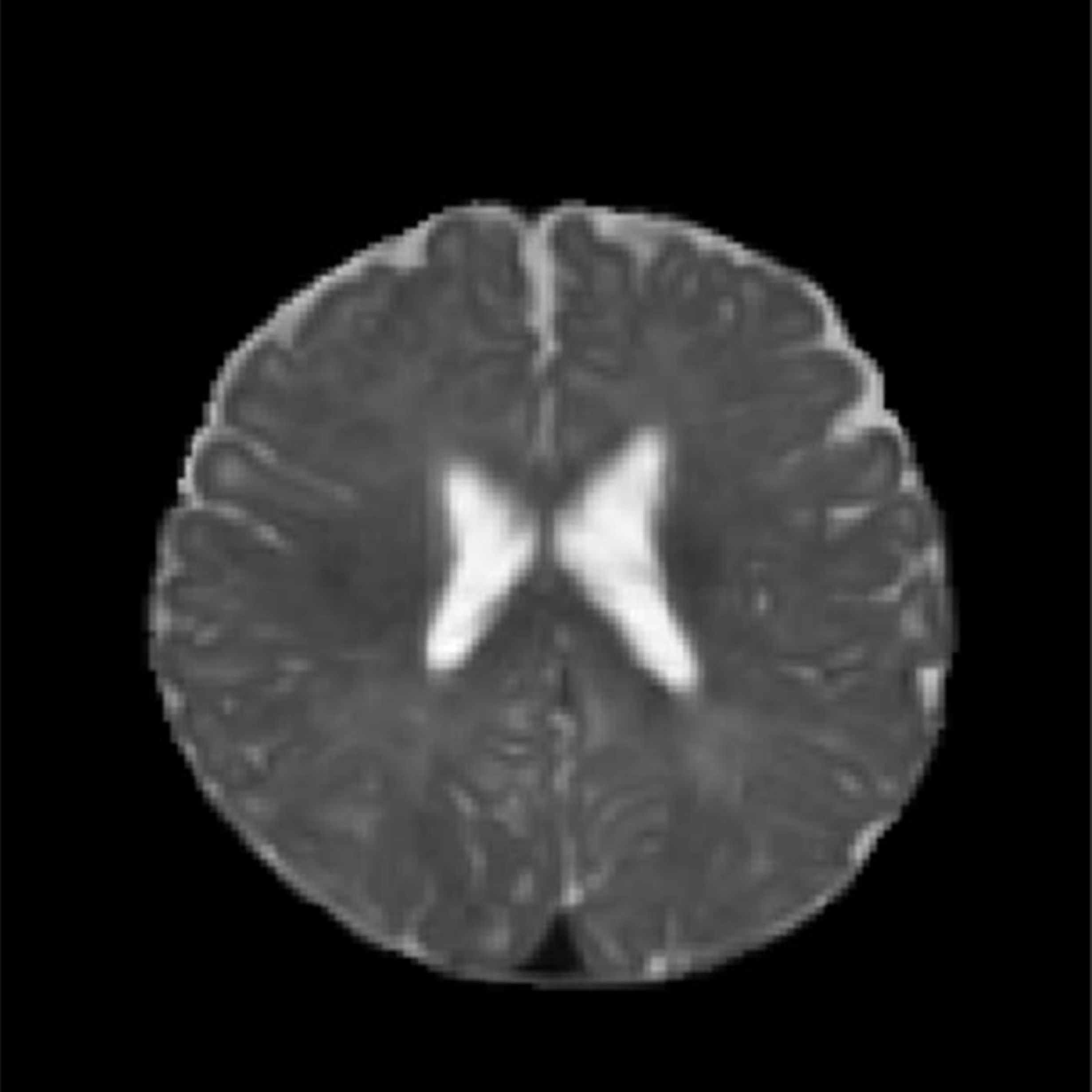}
		&\includegraphics[width=0.144\textwidth]{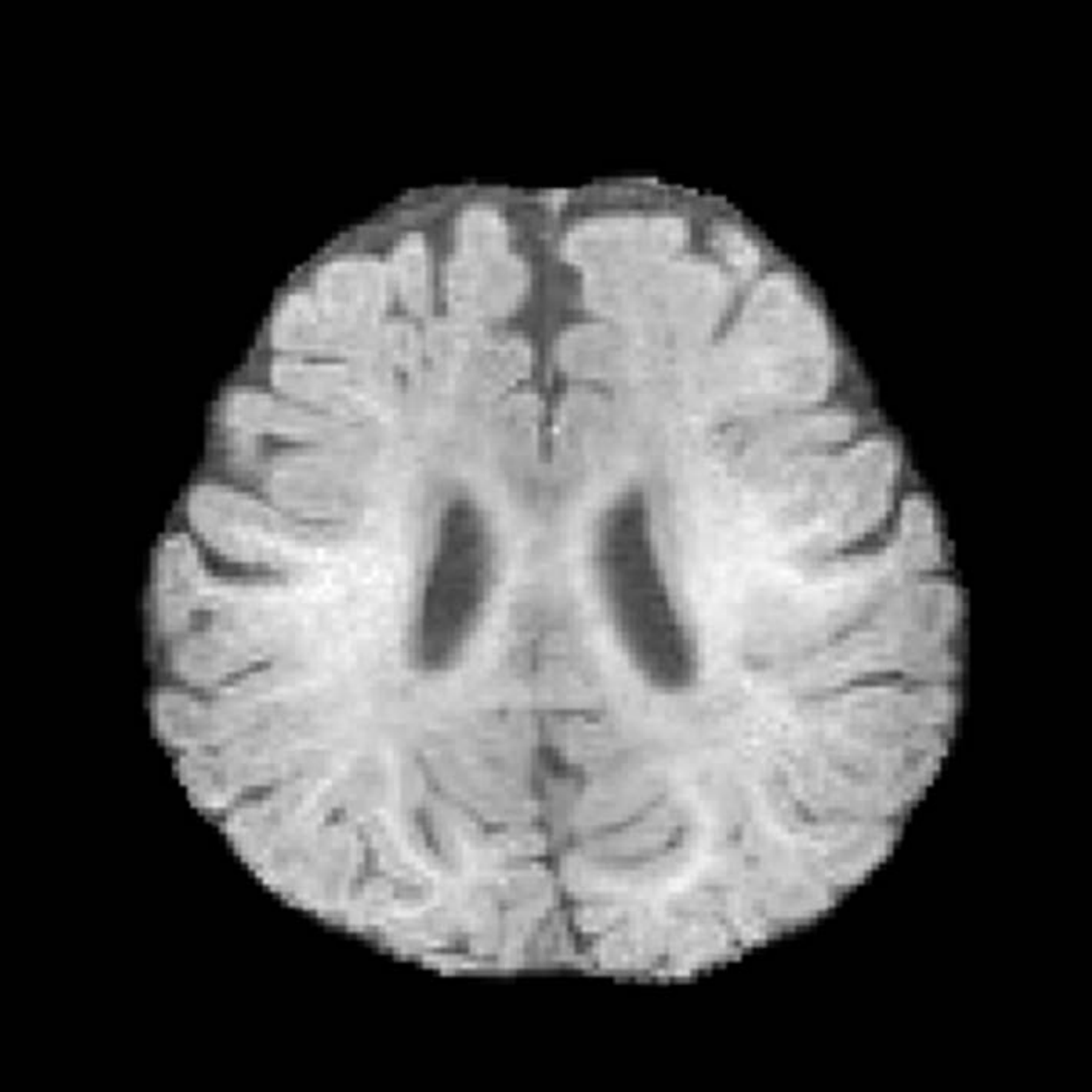}
		&\includegraphics[width=0.144\textwidth]{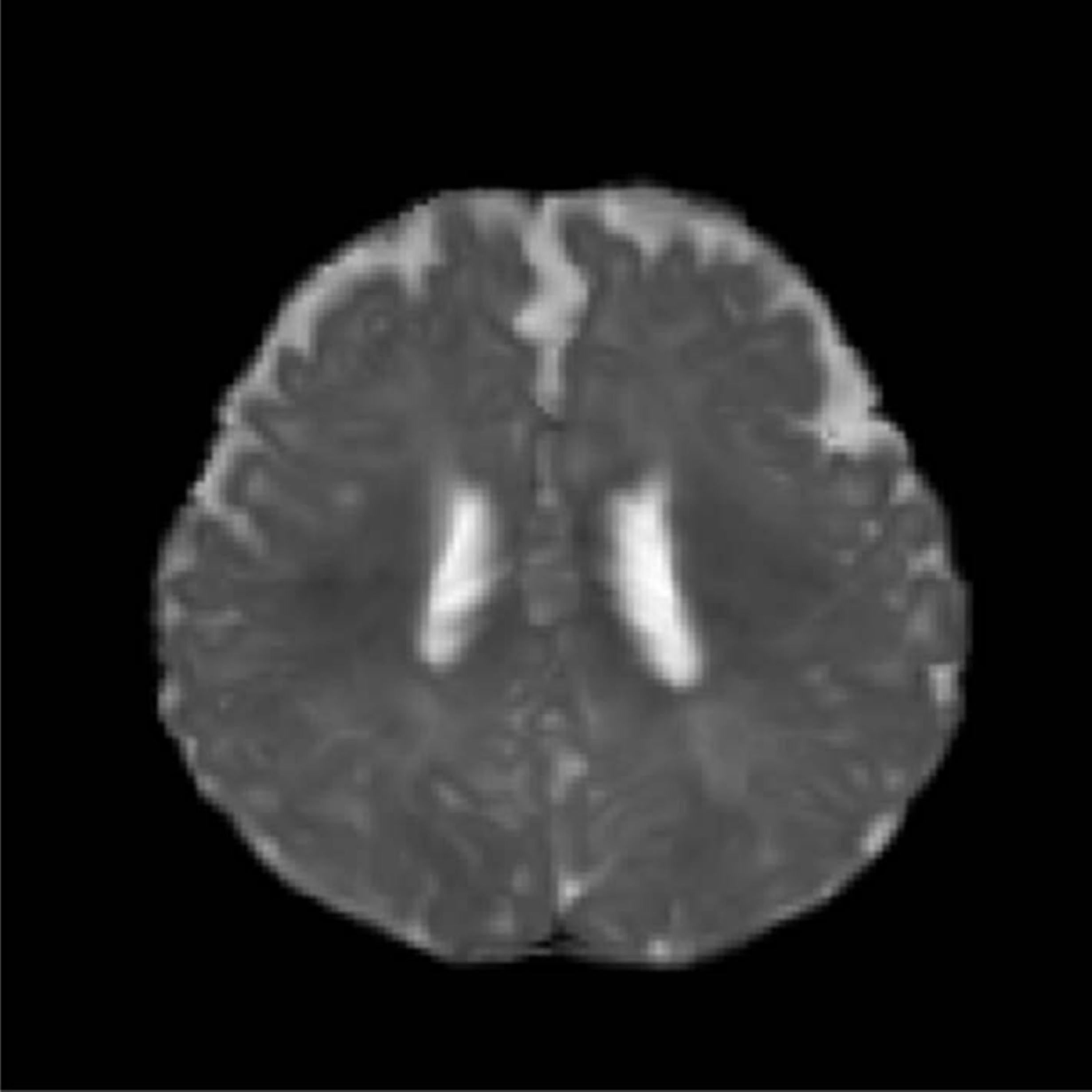}
		&\includegraphics[width=0.144\textwidth]{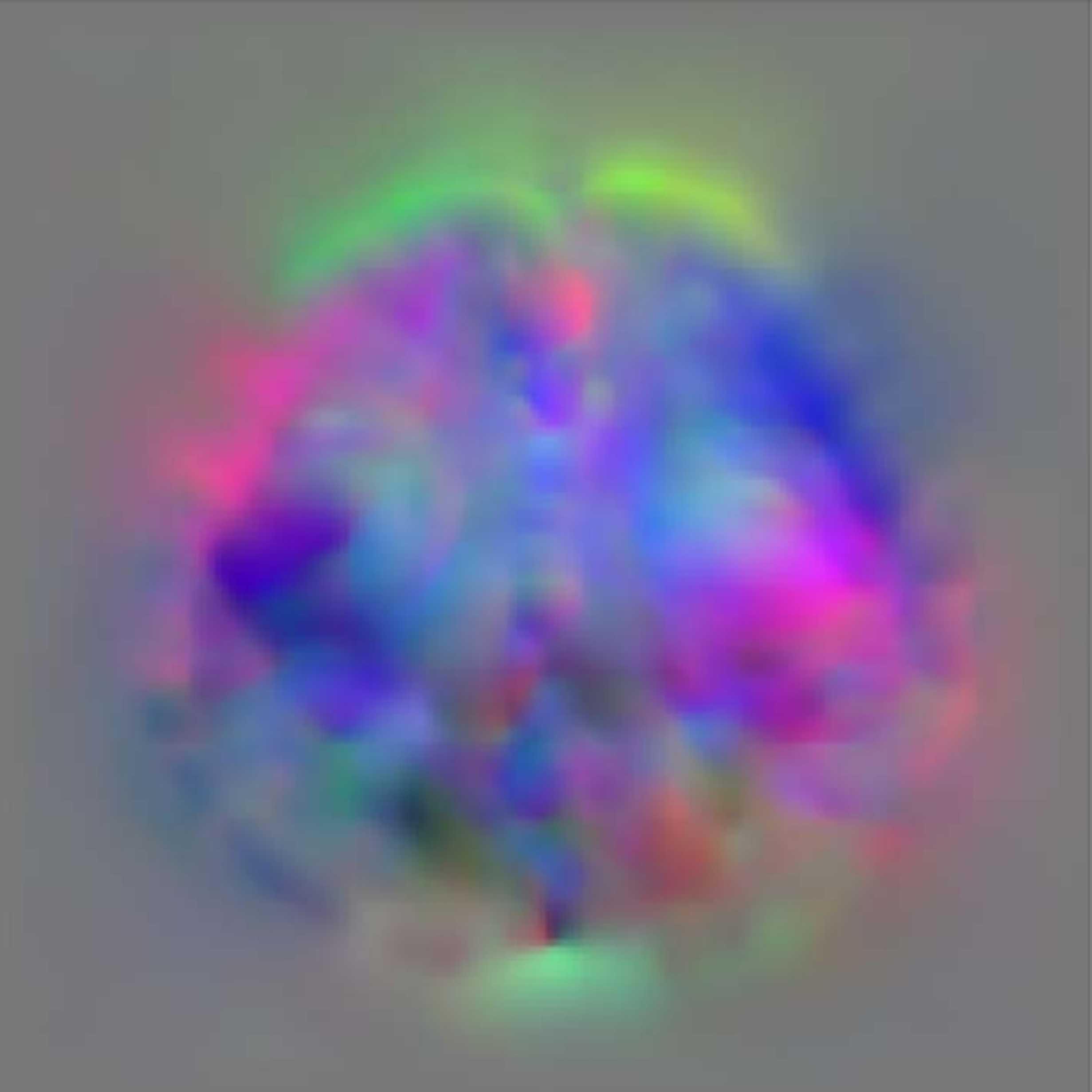}	
		&\includegraphics[width=0.144\textwidth]{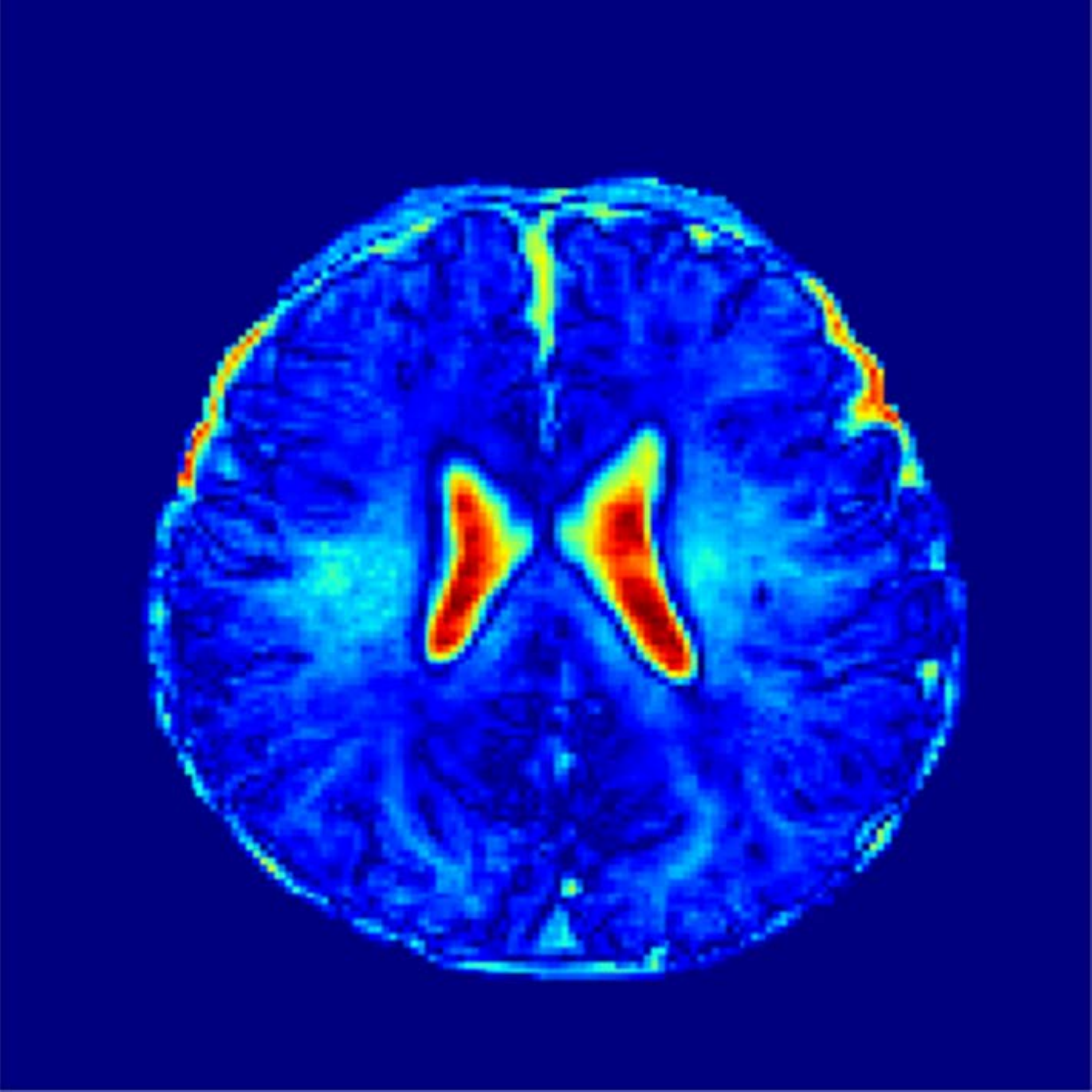}
		&\includegraphics[width=0.144\textwidth]{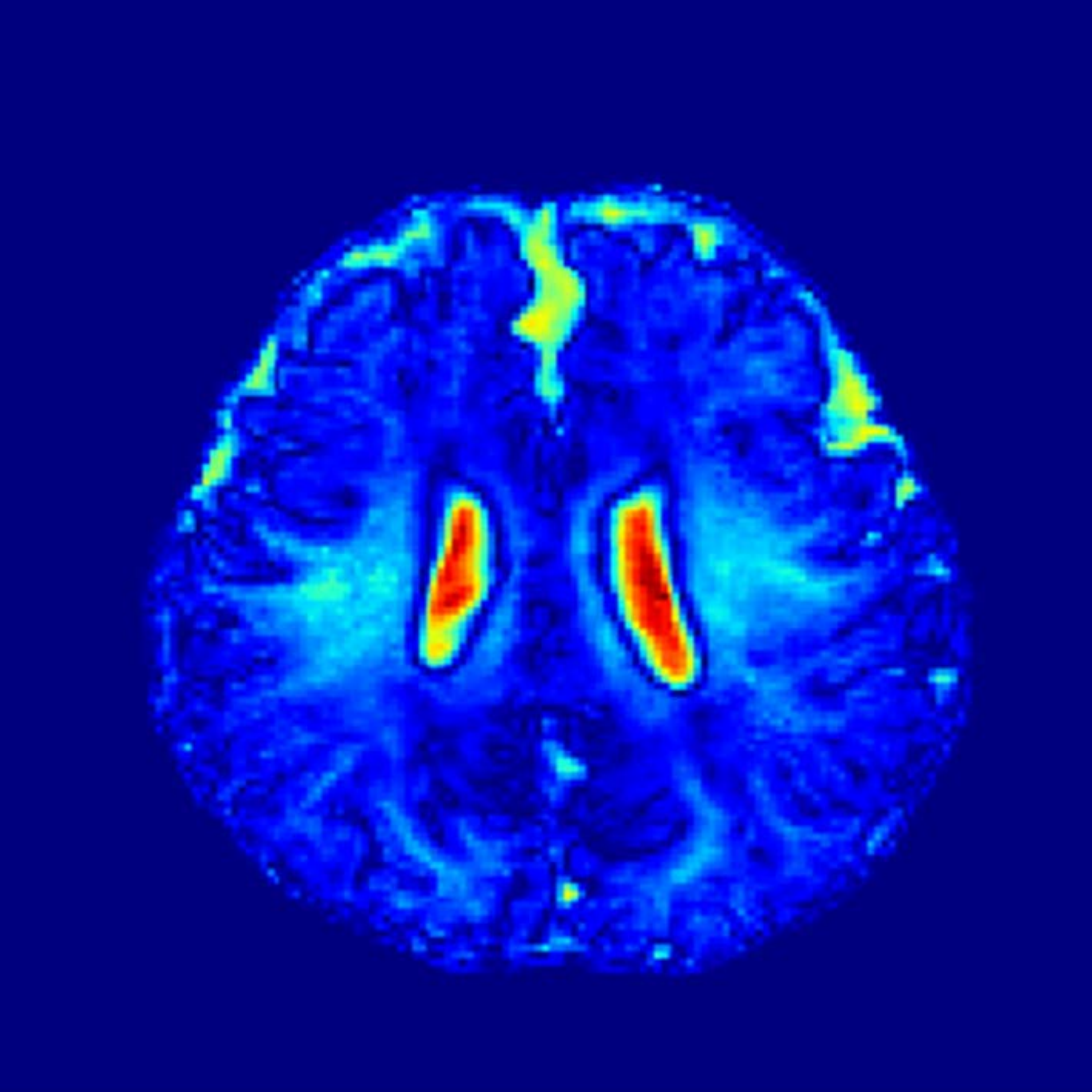} \\
		
			\includegraphics[width=0.144\textwidth]{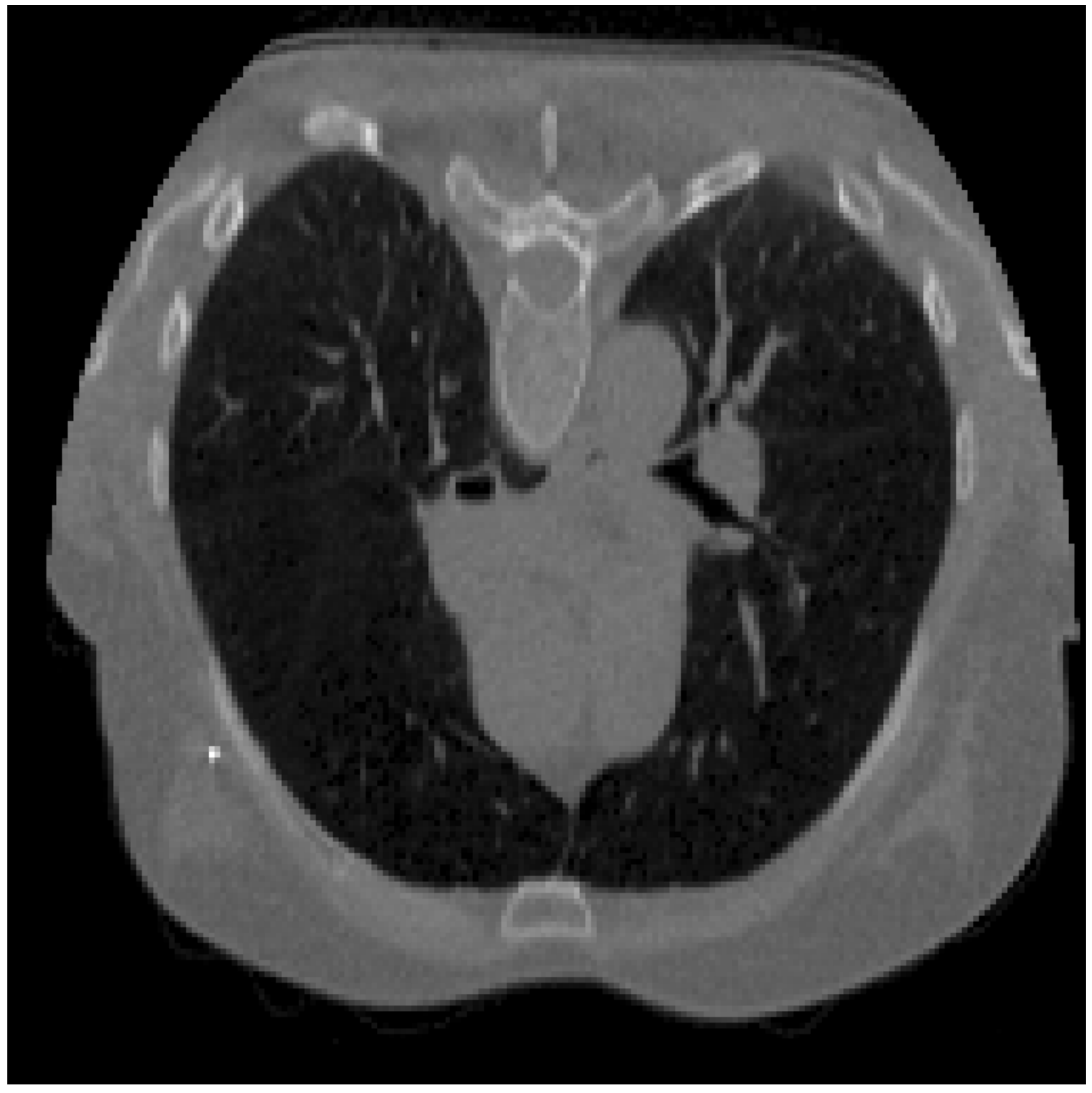}
		&\includegraphics[width=0.144\textwidth]{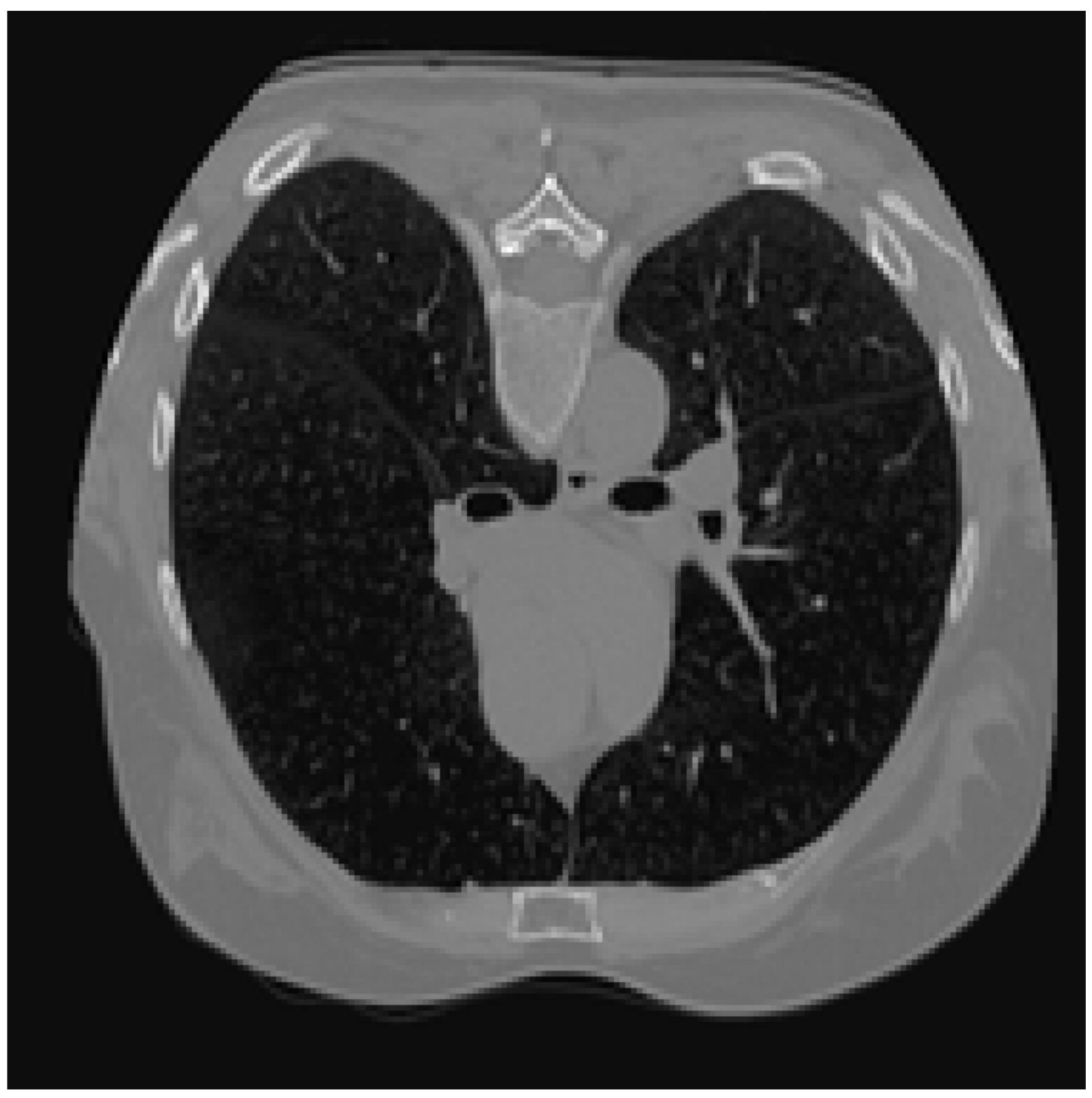}
		&\includegraphics[width=0.144\textwidth]{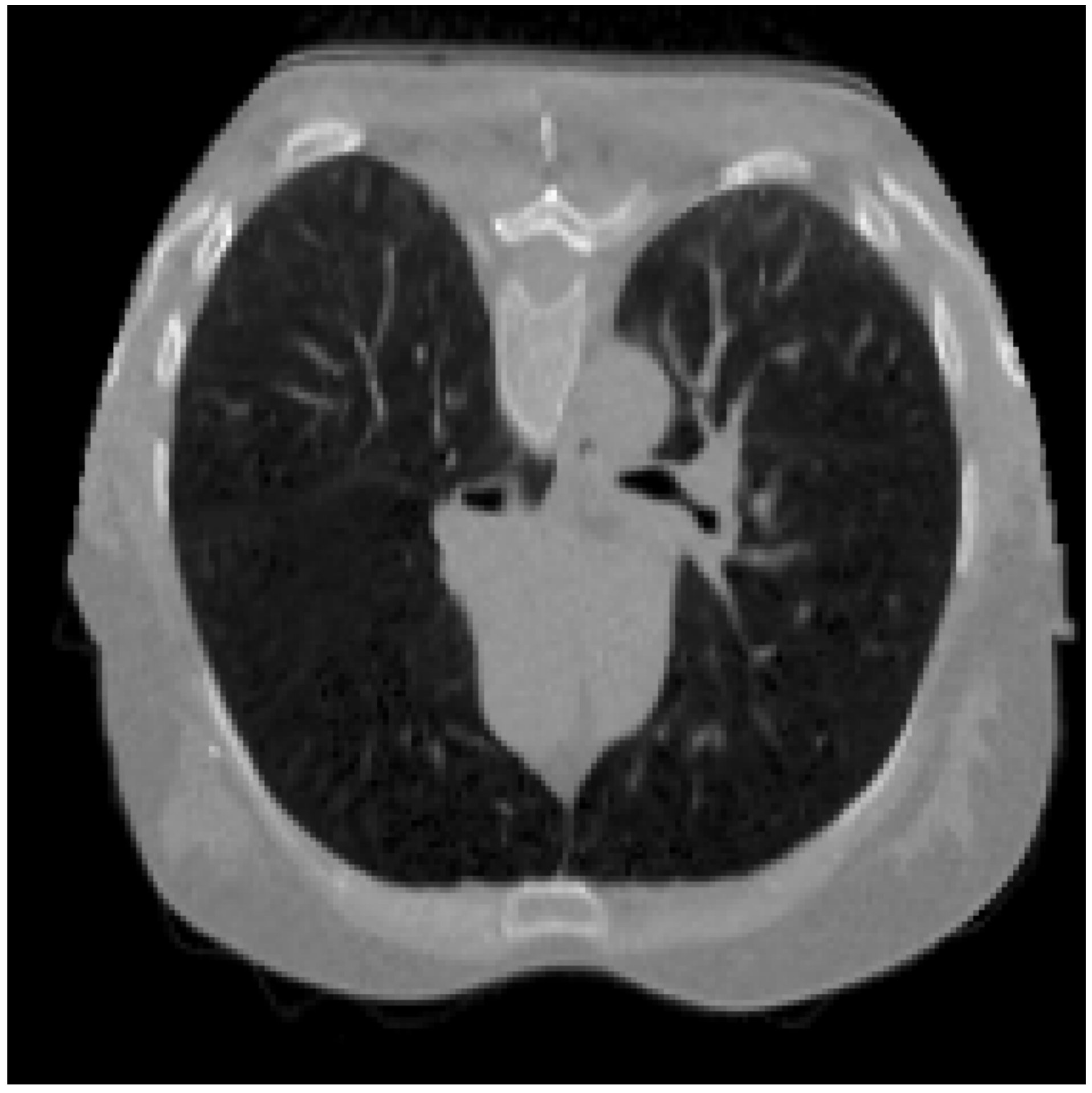}
		&\includegraphics[width=0.144\textwidth]{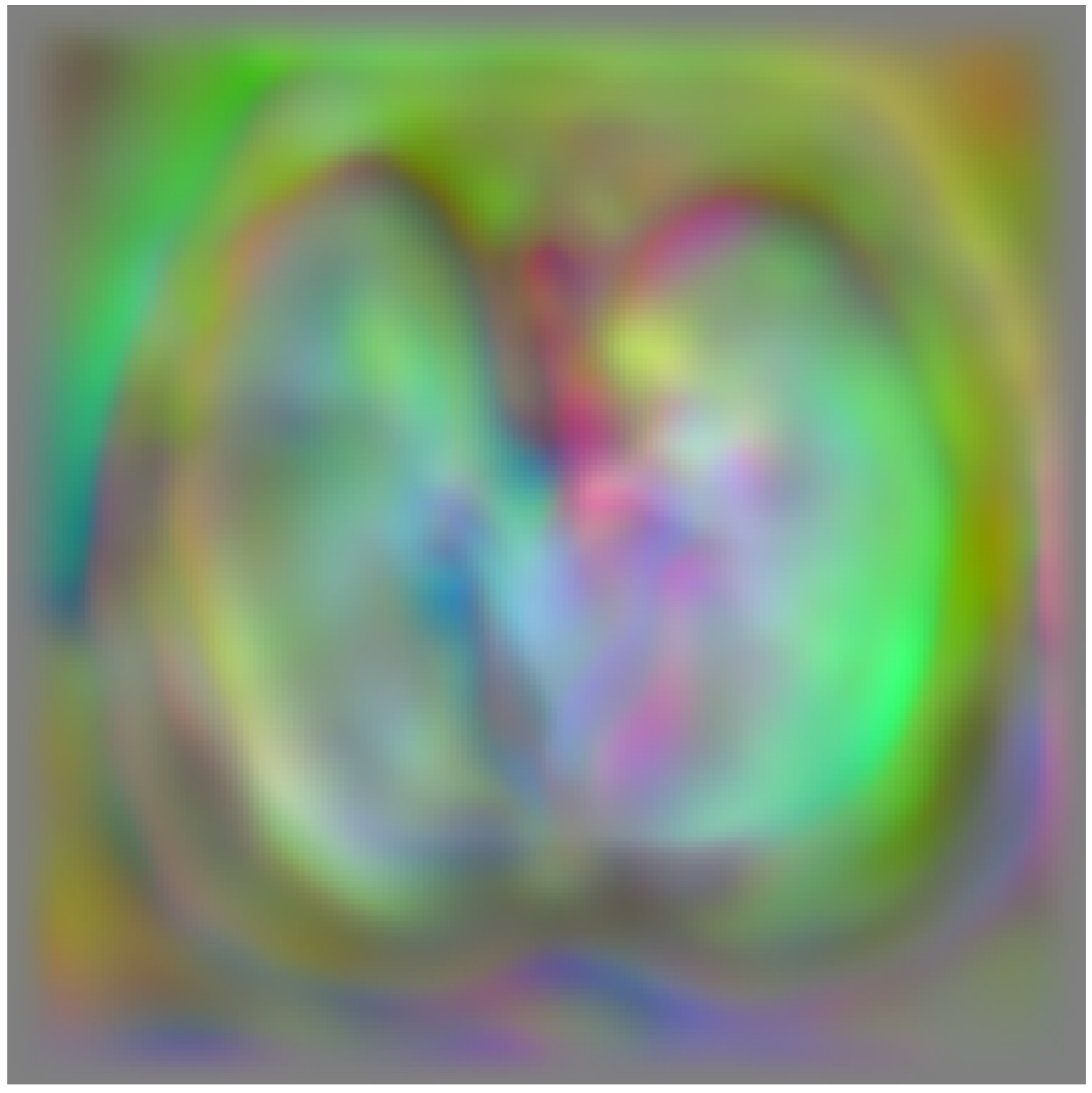}	
		&\includegraphics[width=0.144\textwidth]{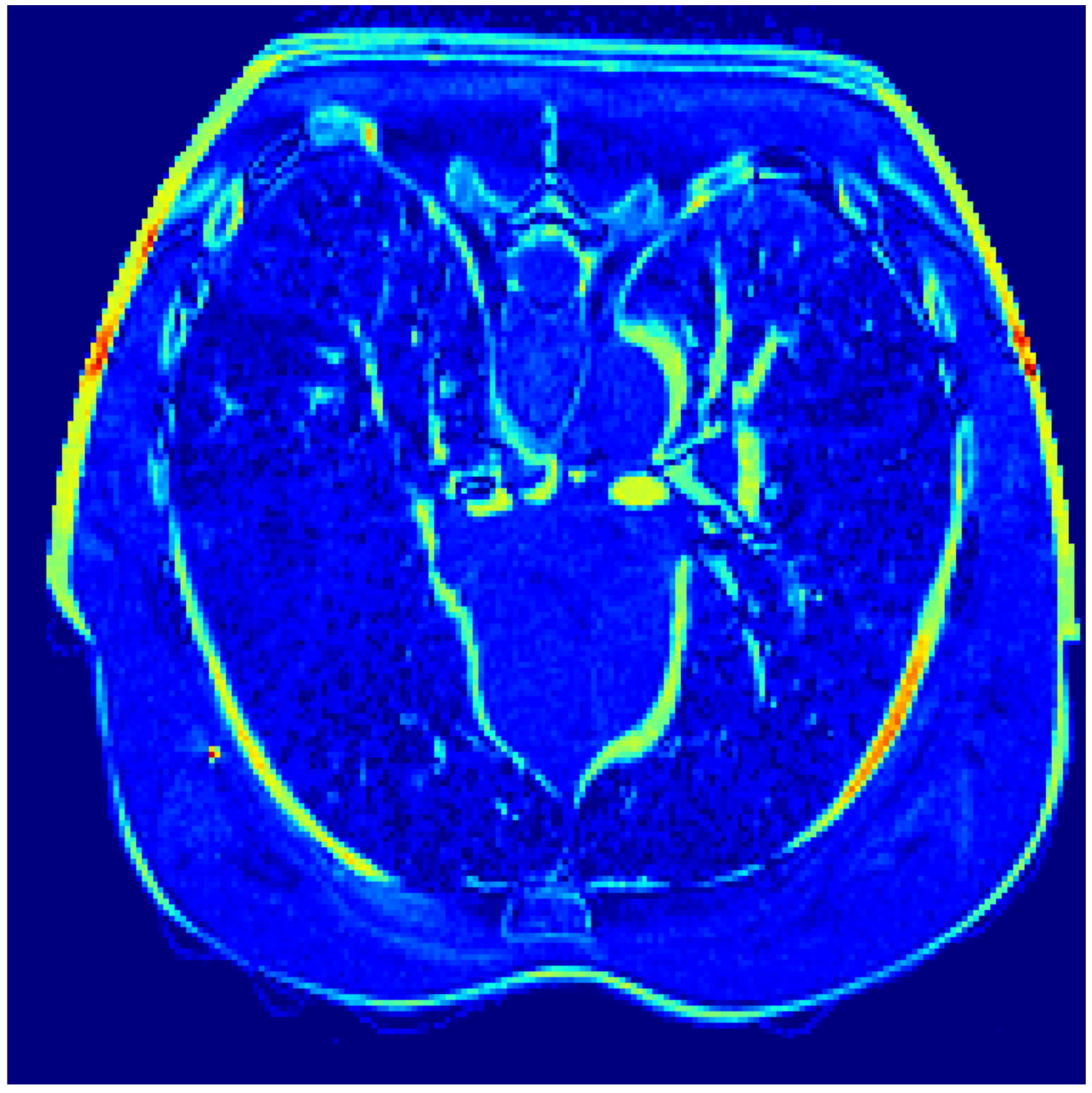}
		&\includegraphics[width=0.144\textwidth]{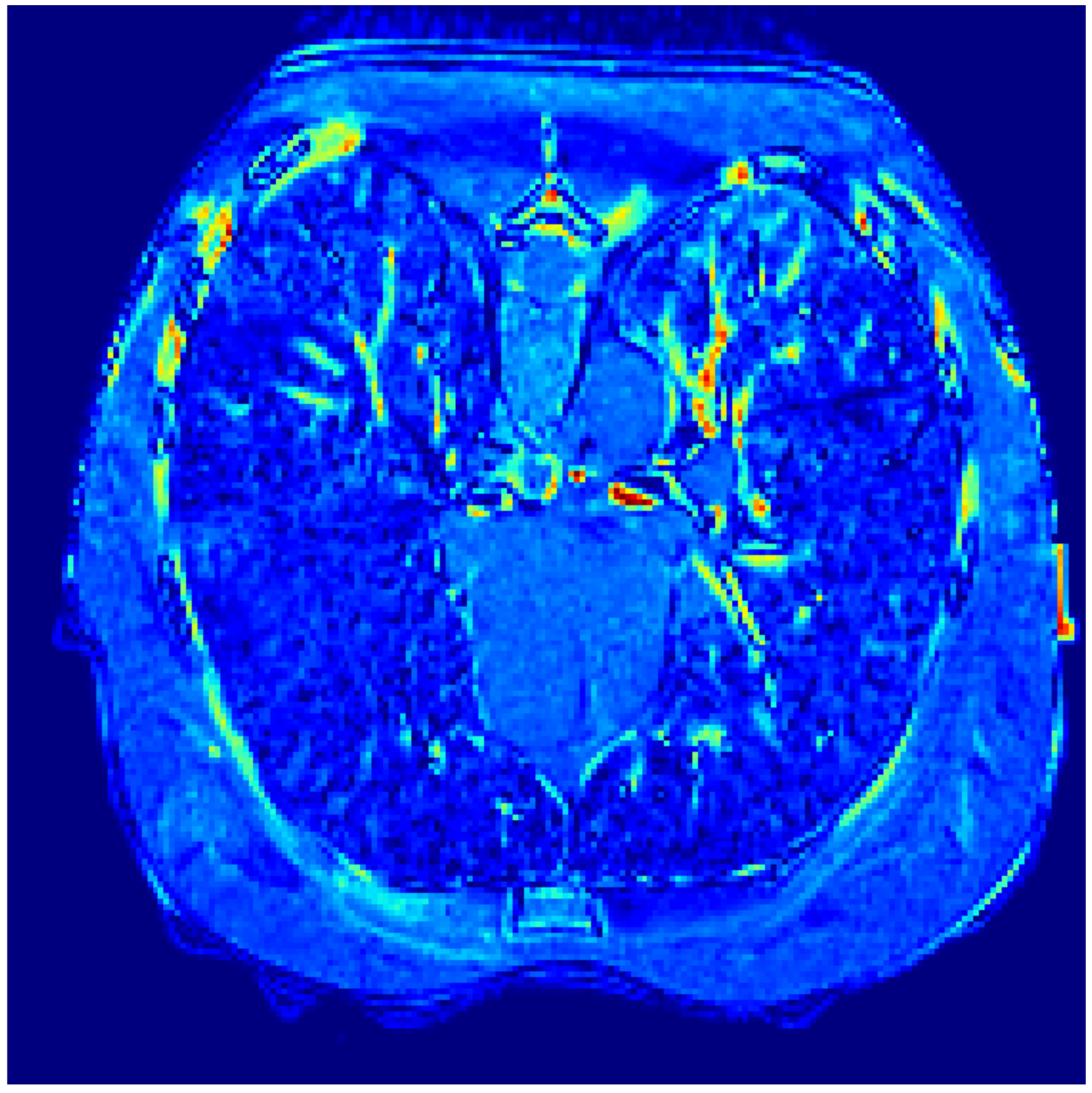} \\
		
		Source  &  Target  &  Warped  & Flow 	& Intensity Error B &  Intensity Error A \\
	
      \end{tabular}
      \caption{ Sample result of registering different images. Each row refers to an example registration case.  Example 2D slices of Intensity difference Before registration and Intensity difference After registration. The registration field is visualized by RGB images with each channel representing dimension. }
	\label{fig:examples-add} 
\end{figure*}

\subsubsection{Comparison results}
First, on the registration task of brain MR image-to-atlas, we quantitatively evaluate the accuracy of all these techniques in terms of running time and Dice score.

In Table.~\ref{tab:compare_dice} and Table.~\ref{tab:Flods}, only our method gives an obvious higher mean and lower variance of Dice score while decreasing the fold number to nearly zero, showing consistent satisfying performance on various datasets. In contrast to other registration algorithms, each registration method has a fixed appropriate network structure with training objectives for its specific alignment application, but our method can adaptively designs the optimal network structure and training objectives for different application scenarios. On the other hand, Table.~\ref{tab:compare_time} presents the runtime and model size among different methods whereas ours needs less inference time and fewer computational parameters.

Table.~\ref{tab:compare_added} demonstrates performance in terms of Dice score on challenging registration tasks, including brain T1 MR image-to-image, brain T2 MR image-to-image, multi-modal \textcolor{update}{(including T1-to-T2 and T2-to-T1)}, and knee T1 MR image-to-image registrations. While our method gives an obvious lower variance with a comparable Dice for all of these cases. 
Fig.~\ref{fig:boxplot1} depicts the stability of the methods in view of the box-plot of Dice score, where fewer outliers and lower variance indicate a more stable registration.
%
Note that, the optimization-based methods perform slightly better for the brain registration where the image pairs are much more similar to each other.
However, they are less satisfying in the knee T1-to-T1 registration.
When faced with such various challenging cases, learning-based models struggle to provide comprehensive solutions. While our method gives an obvious lower variance with a comparable mean of Dice for all these cases, showing stronger stability.

Explanations of unsatisfying performance of VM-diff~\cite{DalcaBGS19}.
Diffeomorphic voxelmorph fails entirely on \textcolor{update}{four} of the evaluation scenarios. It is surprising given the reasonable performance of the non-diffeomorphic models in those applications. 
However, we think the reason is that VM-diff set the Mean Square Error as the default similarity metric which is difficult and incapable to cover challenging registration scenarios. The failure of such magnitude may be caused by weak loss function design, rather than diffeomorphic implementations.

Our representative registration results are given in Fig.~\ref{fig:examples} and Fig.~\ref{fig:examples-add}. The first three registration cases in Fig.~\ref{fig:examples} contain image-to-atlas on T1 brain MR, image-to-image on T1 brain MR and T2 brain MR test pairs. 
The large deformations in scans make registration challenging and difficult. As a result, all the source images are well aligned to the target. 
The second three rows in Fig.~\ref{fig:examples-add} contain knee T1 MR data, multi-modal data, and lung CT inspiration-expiration images.  Although large deformations and intensity differences exist in scans, source images are well aligned to the target, demonstrating our outstanding performance.

\section{Discussions and Conclusions}\label{sec:conclusion}
We come with an automatic learning framework, dubbed as AutoReg, for medical deformable image registration. AutoReg removes the need for computer experts to well-design both architectures and training objectives for the specific type of medical data, also drastically alleviating computation burdens. 
We offer a compelling path toward an automated learning framework for medical image analysis. 
To our best knowledge, this is the first work to achieve AutoML for medical image registration.
Specifically, we construct a triple-level framework to jointly optimize the weights, architecture, and loss function of a deep network for DIR, so that we can automate the process of designing three major components: feature extraction, deformation model and objective function, given specific medical scenario.
Extensive experiments on different image contrasts, anatomical structures, and image modalities have shown that our method may automatically learn an optimal registration network for given volumes and achieve state-of-the-art performance. The auto-learned registration also runs extremely fast inheriting from common learning-based algorithms.

{\bf{Potential downsides of this method.}}
The AutoReg depends on labeled or annotated validation datasets, such as segmentation maps or landmarks, to perform optimization of hyperparameters in loss functions.
On the other hand, the proposed NAS strategy involves hand-crafted designs and has limited degrees of freedom in the searching space. Some of the important elements in neural architecture design such as the presence of skip connection, the depth of the network, and element-wise multiplication and addition are not included in the searching space.

{\bf{Future work.}}
We would extend this method to cover other architectural hyperparameters, like the network topology-level search space that controls the connections among cells, number of layers, resolution levels, and even total network capacity, attempting a more generalized framework for facilitating tuning of highly-parameterized registration models.

\section*{Acknowledgment}
This work was partially supported by the National Key R\&D Program of China (2020YFB1313503), the National Natural Science Foundation of China (Nos. 61922019 and 61672125), LiaoNing Revitalization Talents Program (XLYC1807088), and the Fundamental Research Funds for the Central Universities.
We thank Dr. Adrian V. Dalca at the Massachusetts Institute of Technology for his contributions to the open-source community for medical image registration.

\bibliographystyle{IEEEtran}
\bibliography{refs}

\vfill

\end{document}